\pdfoutput=1
\PassOptionsToPackage{table}{xcolor}
\documentclass[11pt]{article}
\usepackage{acl}

\usepackage{times}
\usepackage{latexsym}

\usepackage[T1]{fontenc}
\usepackage[utf8]{inputenc}

\usepackage{microtype}

\usepackage{inconsolata}
\usepackage[marginal]{footmisc}

\usepackage{graphicx}
\usepackage{subfigure}
\usepackage{caption}
\usepackage{subcaption}
\usepackage{amssymb}
\usepackage{booktabs}
\usepackage{threeparttable}
\usepackage{multirow}
\usepackage{makecell}
\usepackage{amsmath,amsfonts,bm}

\def\eqref#1{equation~\ref{#1}}
\def\1{\bm{1}}

\def\rd{{\textnormal{d}}}

\def\ro{{\textnormal{o}}}
\def\rp{{\textnormal{p}}}

\def\rr{{\textnormal{r}}}
\def\rs{{\textnormal{s}}}

\def\rmG{{\mathbf{G}}}

\DeclareMathAlphabet{\mathsfit}{\encodingdefault}{\sfdefault}{m}{sl}
\SetMathAlphabet{\mathsfit}{bold}{\encodingdefault}{\sfdefault}{bx}{n}

\def\sD{{\mathbb{D}}}
\def\sH{{\mathbb{H}}}

\def\sM{{\mathbb{M}}}

\def\sP{{\mathbb{P}}}

\DeclareMathOperator*{\argmin}{arg\,min}

\colorlet{shadecolor}{gray!20}
\newcommand{\CC}{\cellcolor{shadecolor}}

\title{See the Unseen: Better Context-Consistent Knowledge-Editing by Noises}

\author{
Youcheng Huang$^{\spadesuit\diamondsuit*}$,
\;\;
Wenqiang Lei$^{\spadesuit}$,
\;\;
Zheng Zhang$^{\clubsuit}$,
\;\;
Jiancheng Lv$^{\spadesuit}$, 
\;\;
Shuicheng Yan$^{\heartsuit}$ \\
${\spadesuit}$ College of Computer Science, Sichuan University \\
${\clubsuit}$ Beijing Academy of Artificial Intelligence (BAAI) \\
${\heartsuit}$ Skywork AI \\
${\diamondsuit}$ youchenghuang@stu.scu.edu.cn \\
}

\begin{document}
\maketitle
\footnote{* Work is done during the internship at BAAI.}
\begin{abstract}
Knowledge-editing updates knowledge of large language models (LLMs) and contributes to the interpretability and application of LLMs.
However, knowledge applying is context-consistent: LLMs can recall the \textit{\textbf{same knowledge}} in \textbf{\textit{different contexts}}.
Existing works ignore this property and the editing lacks generalization.
In this paper, we empirically find that the effects of different contexts upon LLMs in recalling the same knowledge follow a Gaussian-like distribution.
We then sample Gaussian noises to simulate the effects of different contexts when updating LLMs.
By such, we can make LLMs \textbf{see the unseen} contexts where the edited knowledge will be applied, therefore improving the editing generalization.
Experimental results on three LLMs demonstrate the effectiveness of our methods and also distinguish our methods from the others of fine-tuning LLMs by noises.
\end{abstract}

\section{Introduction}

\begin{figure*}[!htbp]
    \centering
    \includegraphics[width=\textwidth]{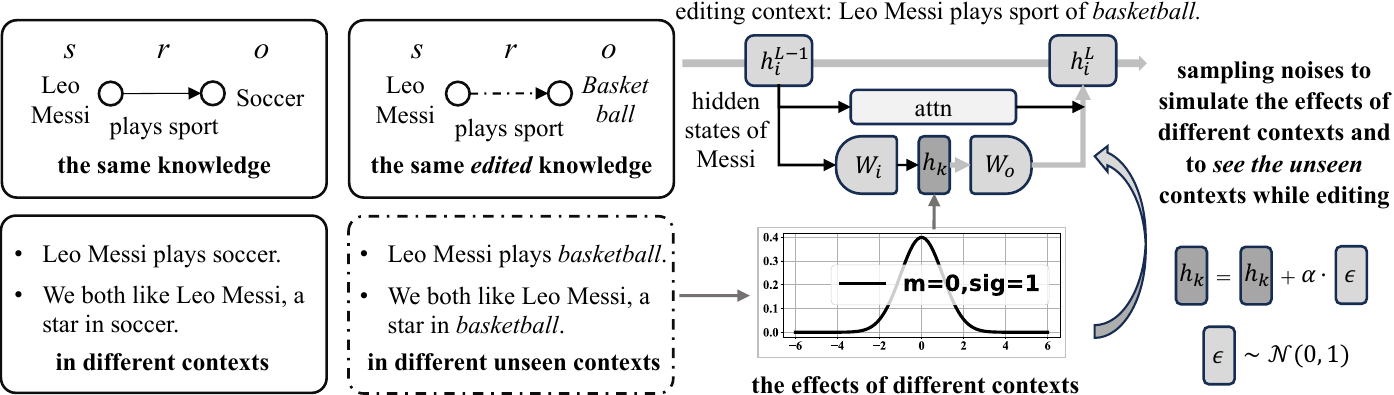}
    \caption{
    Different contexts place shifts that follow a Gaussian\,-like distribution to FFNs' activations on knowledge-related tokens.
    We achieve better context-consistent knowledge-editing by sampling noises to simulate the effects.
    }
    \label{fig:know_consis}
\end{figure*}

Large language models (LLMs) can recall \textit{\textbf{the same}} knowledge in \textit{\textbf{different contexts}}.
How can we edit LLMs' knowledge while maintaining the knowledge can be applied in \textit{\textbf{context-consistency}}?

LLMs \citet{Radford2019LanguageMA,DBLP:conf/nips/BrownMRSKDNSSAA20,gpt-j, gpt-neox-library} recall much knowledge \citet{DBLP:conf/akbc/PetroniLPRWM020,DBLP:conf/nips/BrownMRSKDNSSAA20}, e.g., "Leo Messi plays soccer", but can be unaware of fresh information \citet{DBLP:conf/nips/LazaridouKGALTG21,DBLP:journals/tacl/AgarwalN22} or generate unexpected facts \citet{DBLP:journals/corr/abs-2309-01219}.
To this end, knowledge-editing is proposed to edit LLMs' knowledge by directly updating LLMs' parameters.

Knowledge-editing has considerably improved the interpretability of Transformers \citet{DBLP:conf/nips/VaswaniSPUJGKP17}.
The recent success of editing Feed-Forward Networks (FFNs) \citet{DBLP:conf/nips/MengBAB22,DBLP:conf/iclr/MengSABB23} strongly supports the view that FFNs are key-value memories where Transformers store the knowledge \citet{DBLP:conf/emnlp/GevaSBL21}.
However, LLMs apply knowledge in context-consistency: LLMs recall the same knowledge in different contexts (Figure \ref{fig:know_consis}).
Latest studies of Transformers \citet{bricken2023monosemanticity,DBLP:journals/corr/abs-2309-08600,DBLP:journals/corr/abs-2309-04827} have shown that different patterns, like the active or~passive voice, produce different activations in FFNs.
Let FFNs be a memory.
It is contradictory that the memory can be context-consistent in recalling knowledge but, at the same time, be sensitive to context patterns.

Current editing methods also ignore the context-consistency, therefore lacking generalization.
They include hyper-network training \citet{DBLP:conf/emnlp/CaoAT21,DBLP:conf/iclr/MitchellLBFM22,DBLP:conf/icml/MitchellLBMF22}, constrained fine-tuning \citet{DBLP:journals/corr/abs-2012-00363}, rank-one and cross-layers editing \citet{DBLP:conf/nips/MengBAB22,DBLP:conf/iclr/MengSABB23,DBLP:journals/corr/abs-2308-08742}, focusing on less affecting unrelated knowledge, editing effectiveness, and editing multi-knowledge.
However, they do not recognize that knowledge should be context-consistent.
If you edit knowledge from "Leo Messi plays soccer" to "... basketball", you use the context "Leo Messi plays \textit{basketball}" but apply it with other contexts like "We both like Leo Messi, a star in \textit{basketball}".
Achieving such generalization needs us study that how LLMs can be context-consist in recalling knowledge and how to edit knowledge while maintaining such property.

Following the latest interpreting and editing researches, we study FFNs, especially the activations, and our key finding is that: \textit{different contexts only place small shifts, which follow a considerably narrow Gaussian\,-like distribution, to the FFNs' activations on knowledge-related tokens}.
We use paraphrased texts to analyze how the FFNs' activations change in different contexts of both the knowledge-related tokens and other strings tokens (Section \ref{sec:know_cont}) and discuss the factors (Section \ref{sec:str_cont}).
Motivated by our findings, we improve the editing generalization by adding Gaussian noises to the activations when editing LLMs (Section \ref{sec:see_unseen}).
As Figure \ref{fig:know_consis} shows, the noises can simulate the effects of different contexts and then make LLMs \textbf{see the unseen} contexts that the knowledge will be applied.
Experiments on two benchmarks and three LLMs show significant generalization improvements.
Our method coincides with adding noises in fine-tuning LLMs \citet{DBLP:conf/acl/WuWQ022,jain2023neftune}.
Experiments show that our method best fits in knowledge-editing.

\section{Background and Related Works}
\subsection{Knowledge-Editing: the Task Setting}

LLMs can recall knowledge \citet{DBLP:conf/akbc/PetroniLPRWM020,DBLP:journals/tacl/JiangXAN20,DBLP:journals/jmlr/ChowdheryNDBMRBCSGSSTMRBTSPRDHPBAI23}.
Let us write knowledge of facts in triplet formats (subject $\rs$, relation $\rr$, object $\ro$), e.g., ($\rs\!=$Leo Messi, $\rr\!=$plays sport, $\ro\!=$soccer) in Figure \ref{fig:know_consis}.
And we claim a LLM $\rmG$ can recall a fact $(\rs_i, \rr_i, \ro_i)$ if it predicts the next token(s), which represents $\ro_i$ (soccer), to a natural language prompt $\rp_i\!=\!\rp(\rs_i,\rr_i)$ ("Leo Messi plays").

Let a list of knowledge to edit be the following:
\begin{equation}
\begin{split}
\sM = \{(\rs_i,\rr_i,\ro_i,;\rp_i&) \mid i \in \mathbb{N}\} \\
\text{s.t.}\, \forall\,i,\!j\!. \,(\rs_i=\rs_j) \wedge (\rr_i=&\rr_j) \to (\ro_i=\ro_j)
\end{split}
\end{equation}
where $\left|\sM\right|\!>\!1$ indicates editing multi-knowledge and constraints ensure knowledge without conflicts.
Knowledge-editing changes $\rmG$'s predictions from $\ro_i$ to another object, e.g., $\sM\!=\!\{$(Leo Messi, plays sport, \textit{basketball}; "Leo Messi plays")$\}$.
Let $\rmG'$~be the edited LLM.
The evaluation metrics are:
the \textit{\textbf{effectiveness}} that evaluates whether $\rmG'$ can assign a higher probability to the target $\ro_i$~(basketball) than the original $\ro_i$ (soccer) given $\rp_i$.
Current benchmarks provide one $\rp_i$ to edit $\rmG$.
In case of overfitting, \textit{\textbf{generalization}} evaluates $\rmG'$'s effectiveness on paraphrased $\rp_i^*$ with different contexts, e.g., "What sport does Leo Messi play professionally?".
\textbf{\textit{Specificity}} evaluates that $\rmG'$ should not change any unrelated knowledge, e.g., "What sport Micheal Jordan plays?".
Other metrics such as \textbf{\textit{fluency}} is included.

\subsection{Related Works on Knowledge-Editing}

Different methods share the same optimization objective that maximizes the probability of $\ro_i$ given $\rp_i$.
They diverse in changing different parameters and how to guarantee the generalization and specificity.

The constrained fine-tuning \citet{DBLP:journals/corr/abs-2012-00363,DBLP:conf/iclr/SinitsinPPPB20} or hyper-network \citet{DBLP:conf/emnlp/CaoAT21,DBLP:conf/iclr/MitchellLBFM22,DBLP:conf/icml/MitchellLBMF22} updates all LLMs' parameters with additional losses or techniques~like meta-learning.
Rank-one model editing (ROME) \citet{DBLP:conf/nips/MengBAB22} finds that FFNs store the knowledge in a LLM therefore only update their parameters by solving a constrained linear problem.
While ROME updates FFNs of one layer, recent methods, MEMIT \citet{DBLP:conf/iclr/MengSABB23} and \citet{gao-etal-2023-precise}, follow ROME but update FFNs in multi-layers by solving normal equations \citet{strang2022introduction} and can edit $\left|\sM\right|$=10,000 items.
Although related, classical fine-tuning methods such as LoRA \citet{DBLP:conf/iclr/HuSWALWWC22} show a suboptimal performance \citet{DBLP:conf/emnlp/YaoWT0LDC023}.

Knowledge applying is context-consistent.
Following the state-of-the-art methods, we study how LLMs can recall knowledge in different contexts and improve their editing generalization.

\subsection{Knowledge-editing and the Interpretability of Transformers}

Knowledge-editing receives some criticism \citet{DBLP:conf/emnlp/PinterE23, DBLP:conf/emnlp/ZhongWMPC23} for they mainly focus on one-hop facts.
Nevertheless,~editing research has contributed to the interpretability of Transformers.
Especially, ROME's success of locating and editing knowledge empirically supports that FFNs are the key-value memories where Transformers store knowledge \citet{DBLP:conf/emnlp/GevaSBL21}.
Let $W_i\in\mathbb{R}^{d_k \times d_h},h_i\in\mathbb{R}^{d_h},W_i\in\mathbb{R}^{d_h \times d_k}$, and $f(\cdot)$ be a non-linear function. FFNs' operations~are:
\begin{equation}
    h_o = f(W_i \cdot h_i) \cdot W_o
    \label{eq:ffn}
\end{equation}
Denote the activations $f(W_i\!\cdot\!h_i)$ to be $h_k\in\mathbb{R}^{d_k}$.
FFNs being key-value memories says that different subjects $\rs_i$ activate different $h_k$ that multiply $W_o$ to get the correct $h_o$ of an object $\ro_i$.
Correspondingly, knowledge-editing is to update $W_o$, such as making "Leo Messi" can query out "basketball".
Although being simple, following these ideas, ROME and MEMIT achieve the state-of-the-art performance.

About FFNs, another thread of Transformers circuits finds that different patterns, such as passive voice, can activate different neurons in $h_k$ \citet{elhage2021mathematical,DBLP:journals/corr/abs-2309-08600,DBLP:journals/corr/abs-2309-04827}.
This indicates that $h_k$ is context-sensitive.
But, as Figure \ref{fig:know_consis} shows, knowledge is context consistent.
$h_k$ in different contexts will query out $h_o$ for the same object.
As such, there should be unnoticed mechanisms of LLMs applying knowledge.

\section{The Knowledge Context-Consistency}
Contexts can affect LLMs' knowledge \citet{DBLP:conf/akbc/PetroniLPRWM020}.
For example, prompting \citet{DBLP:journals/csur/LiuYFJHN23} and the in-context learning \citet{DBLP:conf/nips/BrownMRSKDNSSAA20}.
In this section, we study the knowledge context-consistency, i.e., how LLMs can recall the same knowledge in different contexts.
Following ROME \citet{DBLP:conf/nips/MengBAB22} and Transformers circuits \citet{elhage2021mathematical,bricken2023monosemanticity}, we analyze the FFNs activations.
We select the GPT2-xl (1.5B) \citet{Radford2019LanguageMA} and the GPT-J (6B) \citet{gpt-j} as our analyzed LLMs $\rmG$.

\subsection{FFNs Activation in Paraphrased Contexts}
\label{sec:know_cont}

We use the paraphrased texts in knowledge-editing benchmarks \citet{DBLP:conf/nips/MengBAB22} as the different contexts.
Each data $\rd$ provides one $\rp$ for editing and several paraphrased $\rp^{*}$ for evaluation.
For example, $\rp$ is "The mother tongue of Danielle Darrieux is English" and $\rp^{*}$ is "Shayna dose this and Yossel goes still and dies. Danielle Darrieux, a native English".
We first show how $\rp$ and $\rp^{*}$ are lexically different.
Viewing $\rp$ as the reference and $\rp^{*}$ as the predictions, we use BLEU \citet{DBLP:conf/acl/PapineniRWZ02} and ROUGE \citet{lin-2004-rouge}, two widely-adopted metrics, to evaluate their lexical similarities.
Table~\ref{tab:p_d} shows $\rd$ nums and results that $\rp$ and $\rp^{*}$ differ greatly lexically.

\begin{table}[h]
\small
    \centering
    \begin{tabular}{c|c|c|c|c}
    \toprule
    $\rd$ nums & BLEU & ROUGE-1 & ROUGE-2 & ROUGE-L \\
    \midrule
    20,877 & 0.017 & 0.203 & 0.055 & 0.197 \\
    \bottomrule
    \end{tabular}
    \caption{$\rp$ and $\rp^*$ have little lexical similarity. Note that the same subject-string $\rs$ are deleted from $\rp$ and $\rp^*$.}
    \label{tab:p_d}
\end{table}

And we then study the FFNs' activations: $h_k\!=f(W_i \cdot h_i)$ in \eqref{eq:ffn}.
Note that the $\rmG$ predicts $\ro$ by $\rp(\rs, \rr)$ and $\rmG$ stacks layers of Transformers.
Neither each token in $\rp$ nor each layer in $\rmG$ plays the same roles in recalling knowledge.
Therefore, following \citet{DBLP:conf/nips/MengBAB22,DBLP:conf/iclr/MengSABB23}, we select $h_k$ of the last token in $\rs$, denoted as $h_s$, of the $\text{18}^{\text{th}}$ layer in GPT2-xl and the $\text{9}^{\text{th}}$ layer in GPT-J.
Let $h_s^\rp$ be the activations in $\rp$ and $h_s^{\rp*}$ be the ones in $\rp^*$.
We collect an experimental set $\sH_s\!=\!\{h_s^\rp\}\cup\{h_s^{\rp*}\}$ of the last subject token and one control set $\sH_c\!=\!\{h_c^{\rp}\}\cup\{h_c^{\rp*}\}$ of another normal-strings token.
For a better comparison, we manually insert one control token "(" before the subject tokens in $\rp$ and $\rp^*$.\footnote{We can insert different control tokens for different ($\rp,\rp^*$).}
By such, we can make sure that the control tokens are lexically equal in all $\rp$ and $\rp^*$, and have almost the same contexts with the subject tokens.
Then, to study the knowledge context-consistency, we need compare the activations in different contexts.
Therefore, we can collect two difference sets: $\sD_s\!=\!\{h_s^d\!=\!h_s^{\rp^*}\!-\!h_s^\rp\ \mid (\rp, \rp^{*}) \in \{\rd\} \}$ and $\sD_c\!=\!\{h_c^d\!=\!h_c^{\rp^*}\!-\!h_c^\rp\ \mid (\rp, \rp^{*}) \in \{\rd\} \}$.
We plot the histograms of all the activation neurons, i.e., flatting scalars in each dimensions of all $h$ and plot them together.
\begin{figure}[h]
    \centering
    \begin{minipage}[t]{0.48\linewidth}
        \centering
        \includegraphics[scale=0.2]{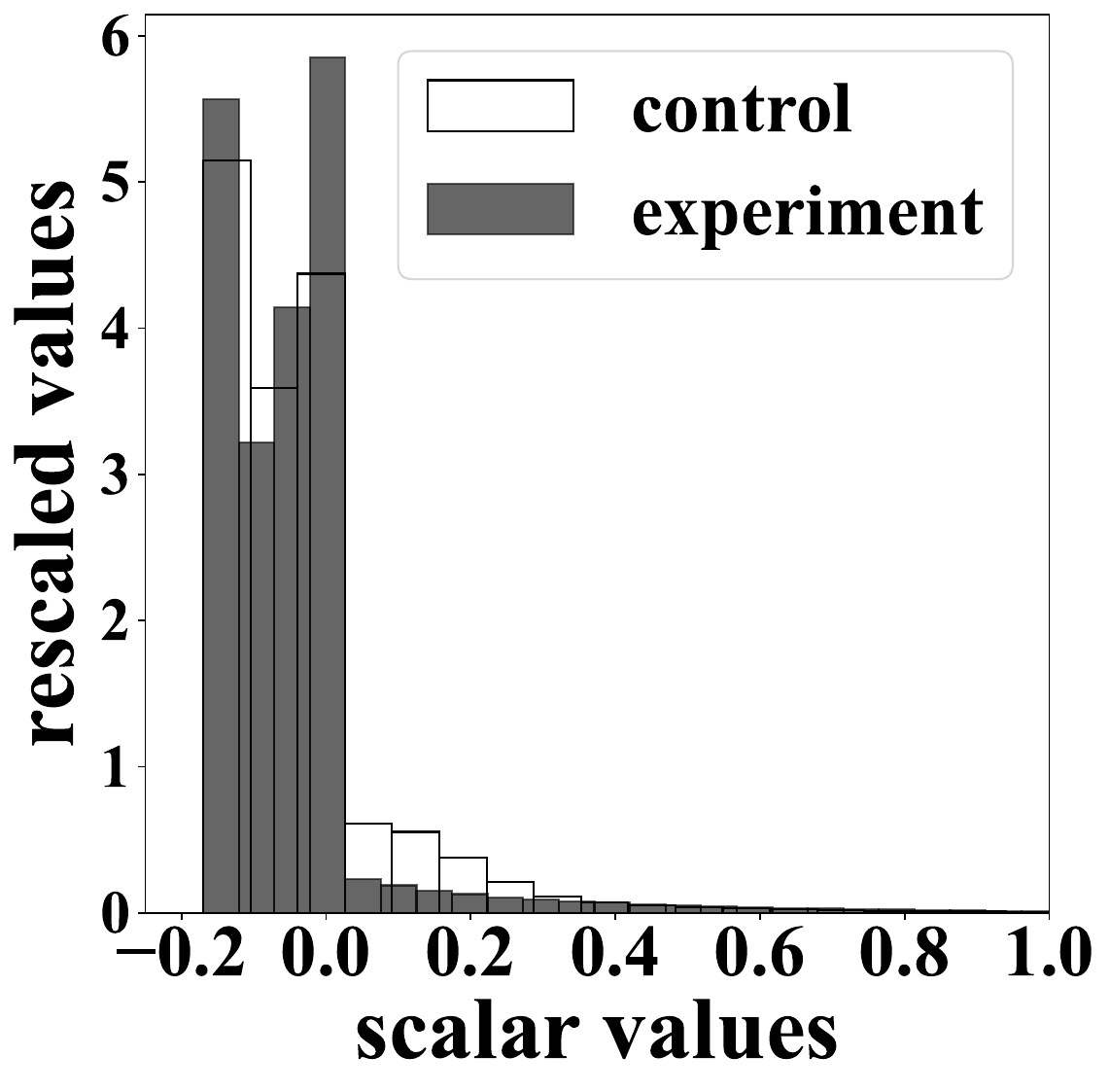}
        \vspace{-7mm}
        \caption{GPT2-xl $\sH_s,\!\sH_c$.}
        \label{fig:h_gpt2}
    \end{minipage}
    \begin{minipage}[t]{0.48\linewidth}
        \centering
        \includegraphics[scale=0.2]{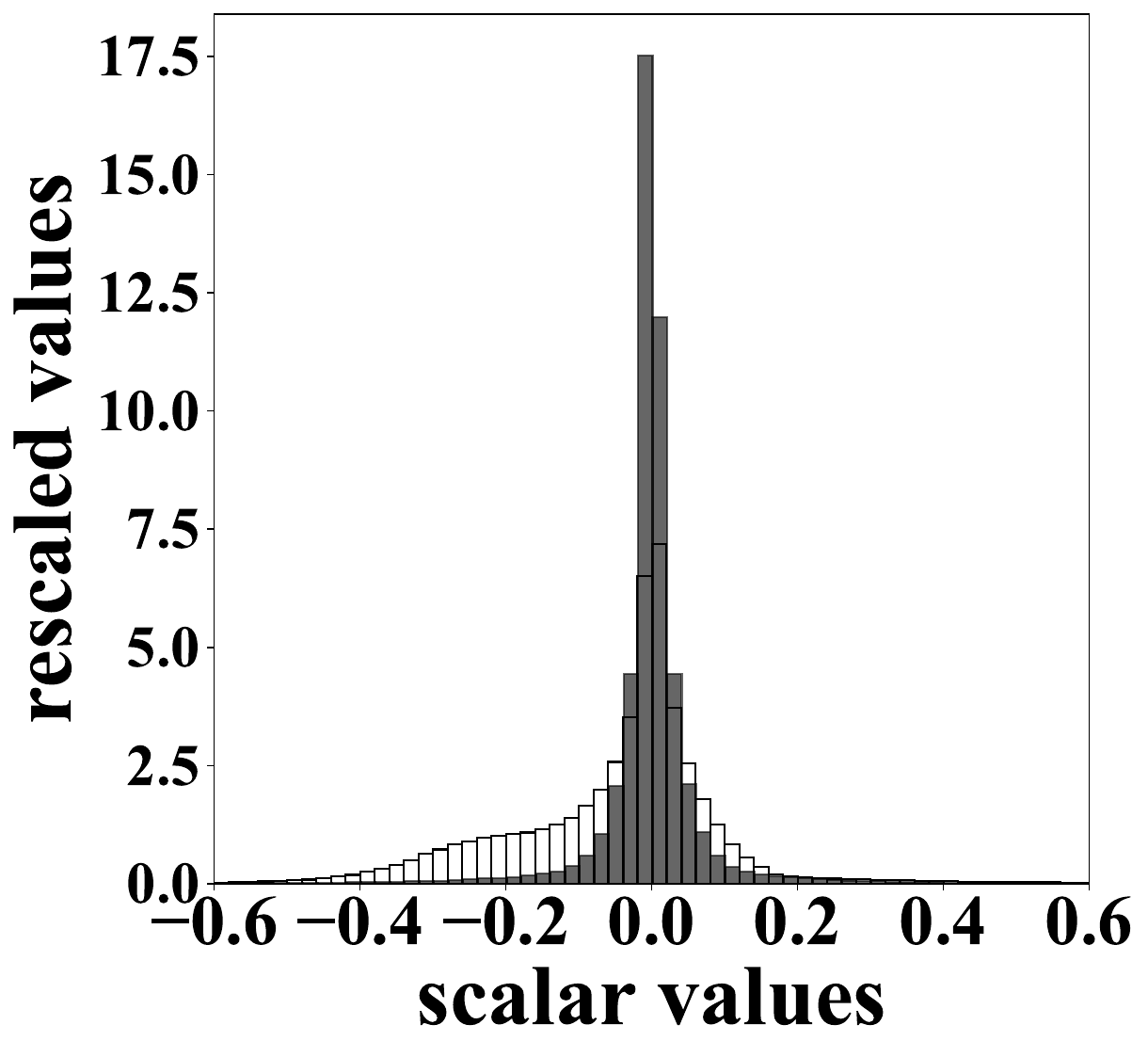}
        \vspace{-7mm}
        \caption{GPT2-xl $\sD_s,\!\sD_c$.}
        \label{fig:d_gpt2}
    \end{minipage}\\
    \vspace{3mm}
    \centering
    \begin{minipage}[t]{0.48\linewidth}
        \centering
        \includegraphics[scale=0.2]{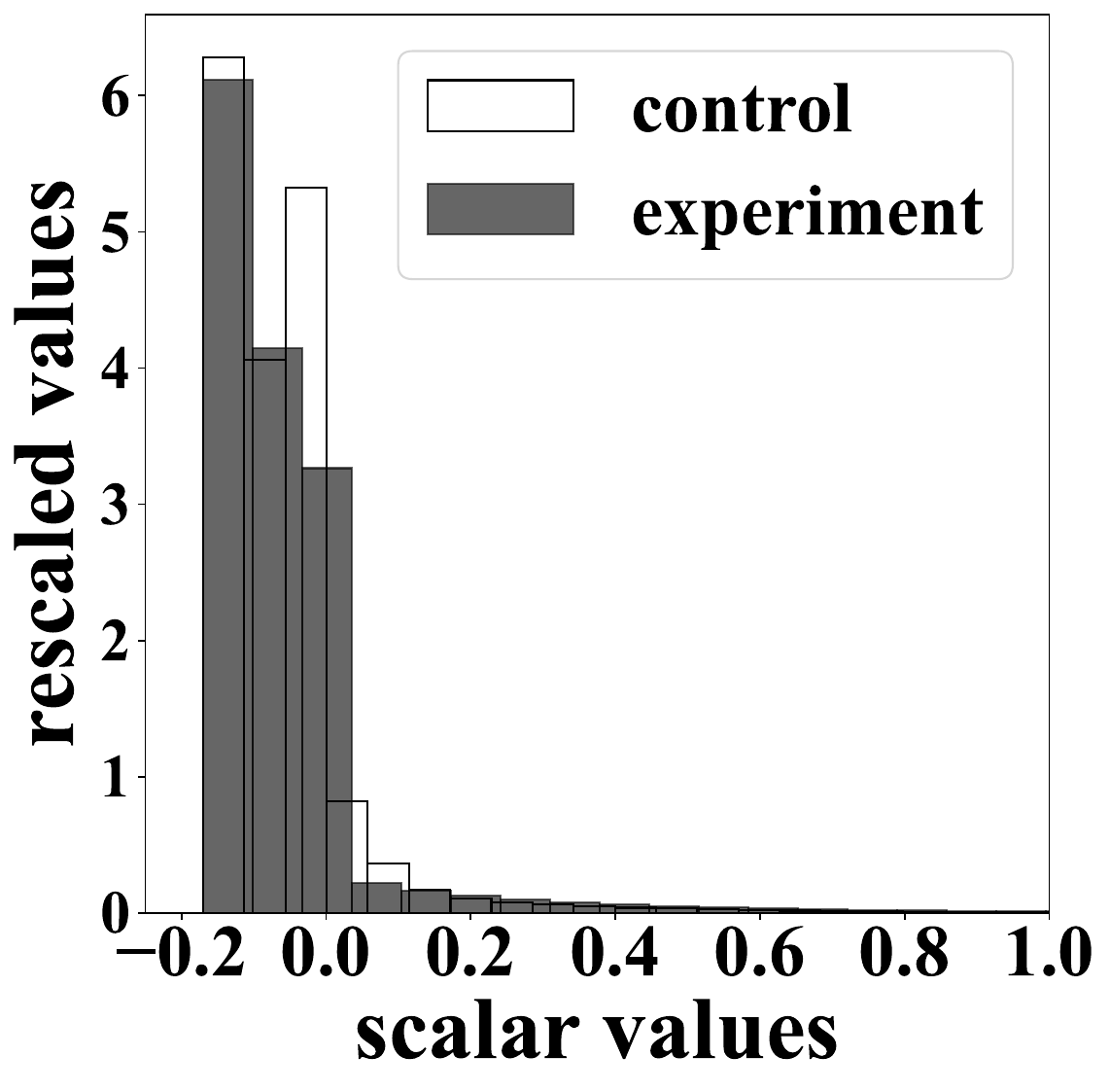}
        \vspace{-7mm}
        \caption{GPT-J $\sH_s,\!\sH_c$.}
        \label{fig:h_gptj}
    \end{minipage}
    \begin{minipage}[t]{0.48\linewidth}
        \centering
        \includegraphics[scale=0.2]{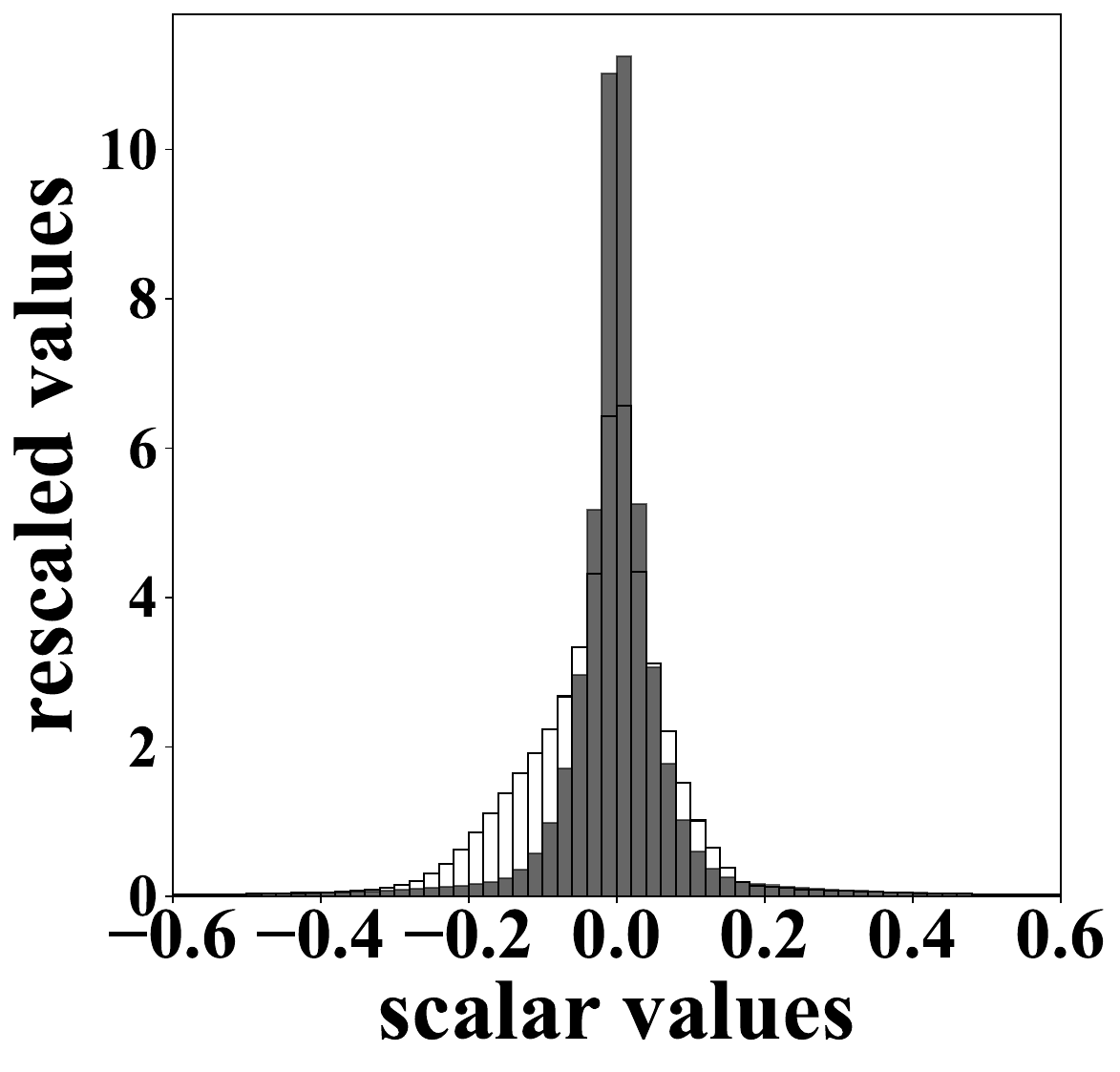}
        \vspace{-7mm}
        \caption{GPT2-J $\sD_s,\!\sD_c$.}
        \label{fig:d_gptj}
    \end{minipage}
\end{figure}

The above figures plot the results where the black rectangles plot the experimental sets and the whites plot the control sets.
From Figure \ref{fig:h_gpt2} and \ref{fig:h_gptj}, both control and experimental sets on GPT2-xl and GPT-J have their activation scalars,\footnote{The two $\rmG$ both use the "new-gelu" non-linear function~$f$.} with a major proportions, fallen in the interval of $(-0.2, 0)$.
However, the difference sets perform a significant difference.
The experimental sets $\sD_s$ have their scalars mostly concentrated around $0$ and descend symmetrically and evenly to the both sides while the control sets $\sD_c$ show a greater skewness when descending.
We calculate the skewness and kurtosis of the both sets (shown in Table~\ref{tab:skew_kurt}).
\begin{table}[h]
\small
    \centering
    \begin{tabular}{cc|c|c|c}
    \toprule
    \multirow{2}{*}{Sets} & 
    \multicolumn{2}{c}{GPT2-xl} & 
    \multicolumn{2}{c}{GPT-J} \\
    \cmidrule(lr){2-3}\cmidrule(lr){4-5}
    & Skewness & Kurtosis & Skewness & Kurtosis \\
    \midrule
    $\sD_s$ & \textbf{-0.53} & 40.98 & \textbf{-0.20} & 38.29 \\
    $\sD_c$ & -5.12 & 161.84 & 0.45 & 42.70 \\
    \bottomrule
    \end{tabular}
    \caption{Skewness and Kurtosis of the two sets.}
    \label{tab:skew_kurt}
\end{table}
From the histograms and the quantitative results, $\sD_s$ follows a Gaussian-like distribution, where the much larger kurtosis differs $\sD_s$ from the normal Gaussian.
This is understandable for the raw scalars majorly have small values.

\subsection{What are the Factors?}
\label{sec:str_cont}
We have shown that, for knowledge-related tokens, contexts that of great lexical differences (Table~\ref{tab:p_d}) can only place small shifts, which follow a considerably~narrow Gaussian-like distribution (Figure \ref{fig:d_gpt2},\ref{fig:d_gptj} and Table \ref{tab:skew_kurt}), in FFNs' activations.
In this section, we discuss its factors from two possible sides:

1. knowledge-related tokens have \textit{narrow attention scopes} therefore being context-consistent.

2. Such consistency is \textit{FFNs particular behavior} to knowledge-related tokens even in the first layer.

\noindent\textbf{Does Attention Differ?} For the first side, we collect the attention scores of the subject/control token to other tokens in $\rp, \rp^*$ from the first Transformer layer to the layer where we pick-up the activations.
\begin{figure}[h]
    \centering
    \begin{minipage}[t]{0.48\linewidth}
        \centering
        \includegraphics[scale=0.2]{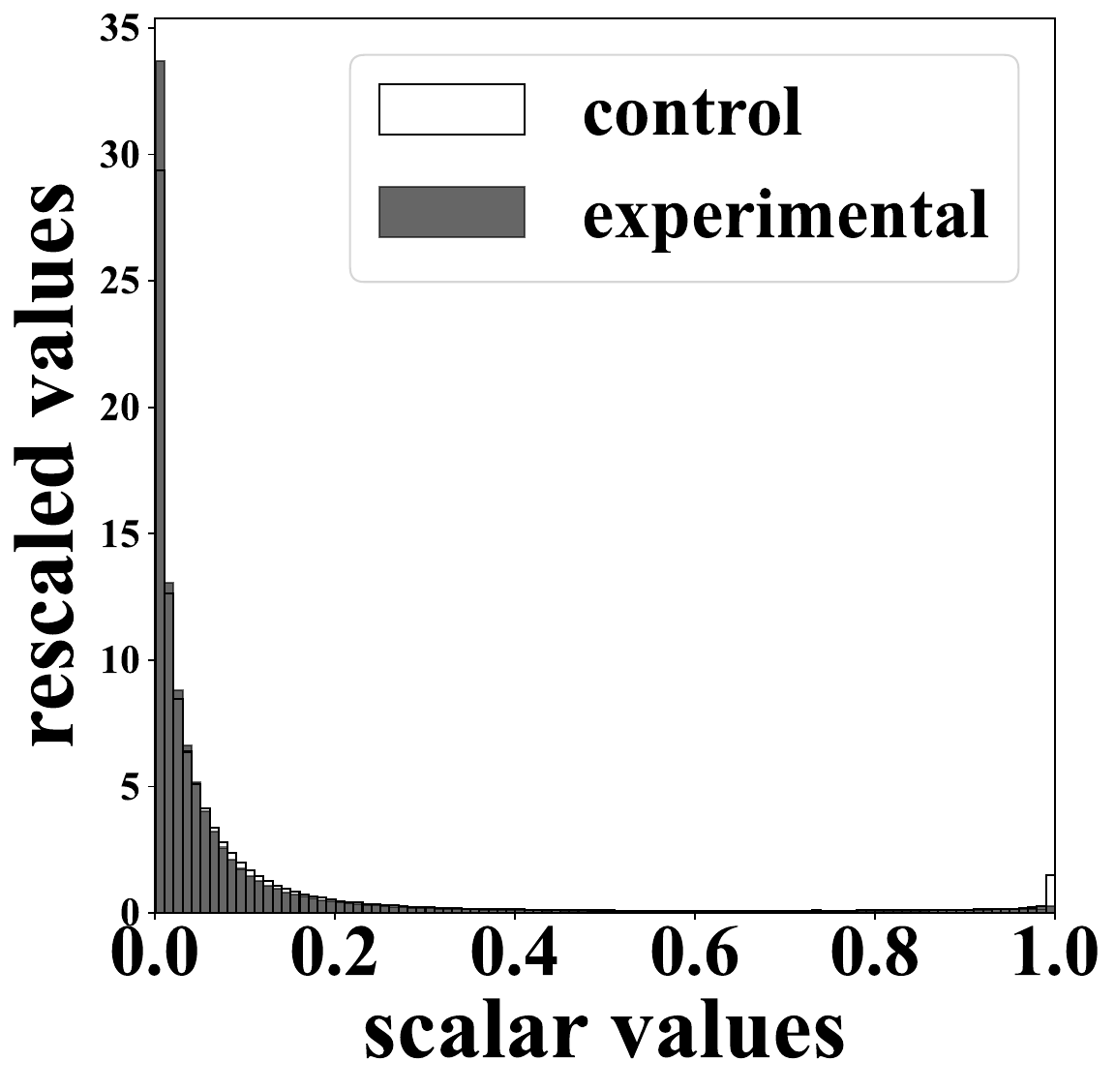}
        \vspace{-7mm}
        \caption{GPT2-xl Attns.}
        \label{fig:attn_gpt2}
    \end{minipage}
    \hspace{1mm}
    \begin{minipage}[t]{0.48\linewidth}
        \centering
        \includegraphics[scale=0.2]{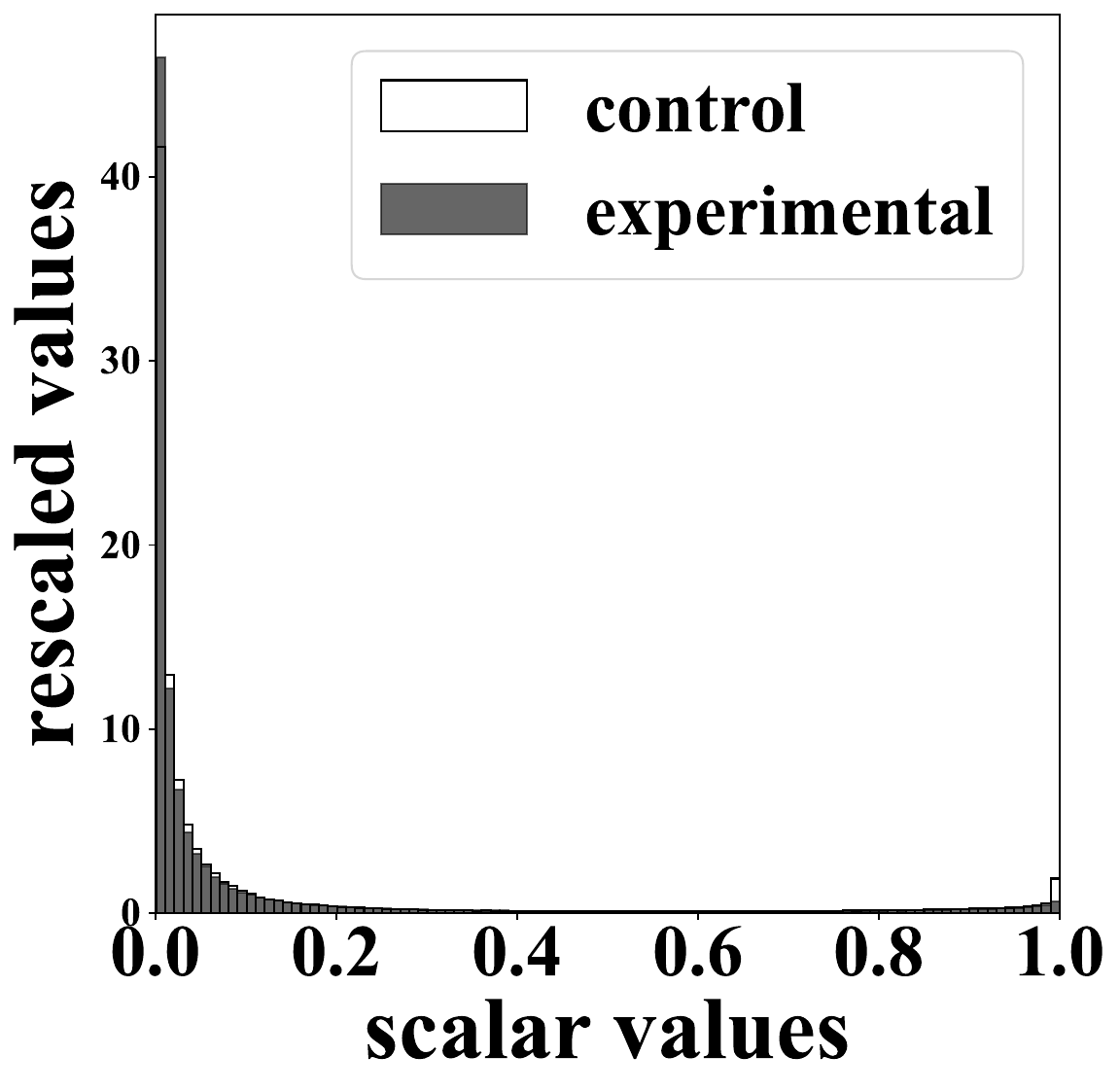}
        \vspace{-7mm}
        \caption{GPT-J Attns.}
        \label{fig:attn_gptj}
    \end{minipage}
\end{figure}
We then plot the histograms of the attention scores.
The Figure \ref{fig:attn_gpt2} and Figure \ref{fig:attn_gptj} displays the results.
We can see that the black rectangles and the white ones are almost overlapped, indicating that the attention scopes between the subject tokens and the control tokens are nearly the same otherwise the black rectangles should concentrate on larger values.

\noindent\textbf{FFNs Particular Behavior.}
If the attention does not response for the context-consistency, then FFNs should themselves have particular actions to knowledge tokens.
We conclude such property by empirically showing that FFNs in different layers have the same behavior.
We re-collect the FFNs activations from the first Transformer layer to the layer that we previously selected.
Because of the page space limitations, we only plot $\sD_s, \sD_c$ of the first, the middle, and the last layer here and refer readers to the Appendix \ref{apd:l_a} for the integrated results.
\begin{figure}[ht]
    \centering
    \subfigure
        \centering
        \includegraphics[scale=0.125]{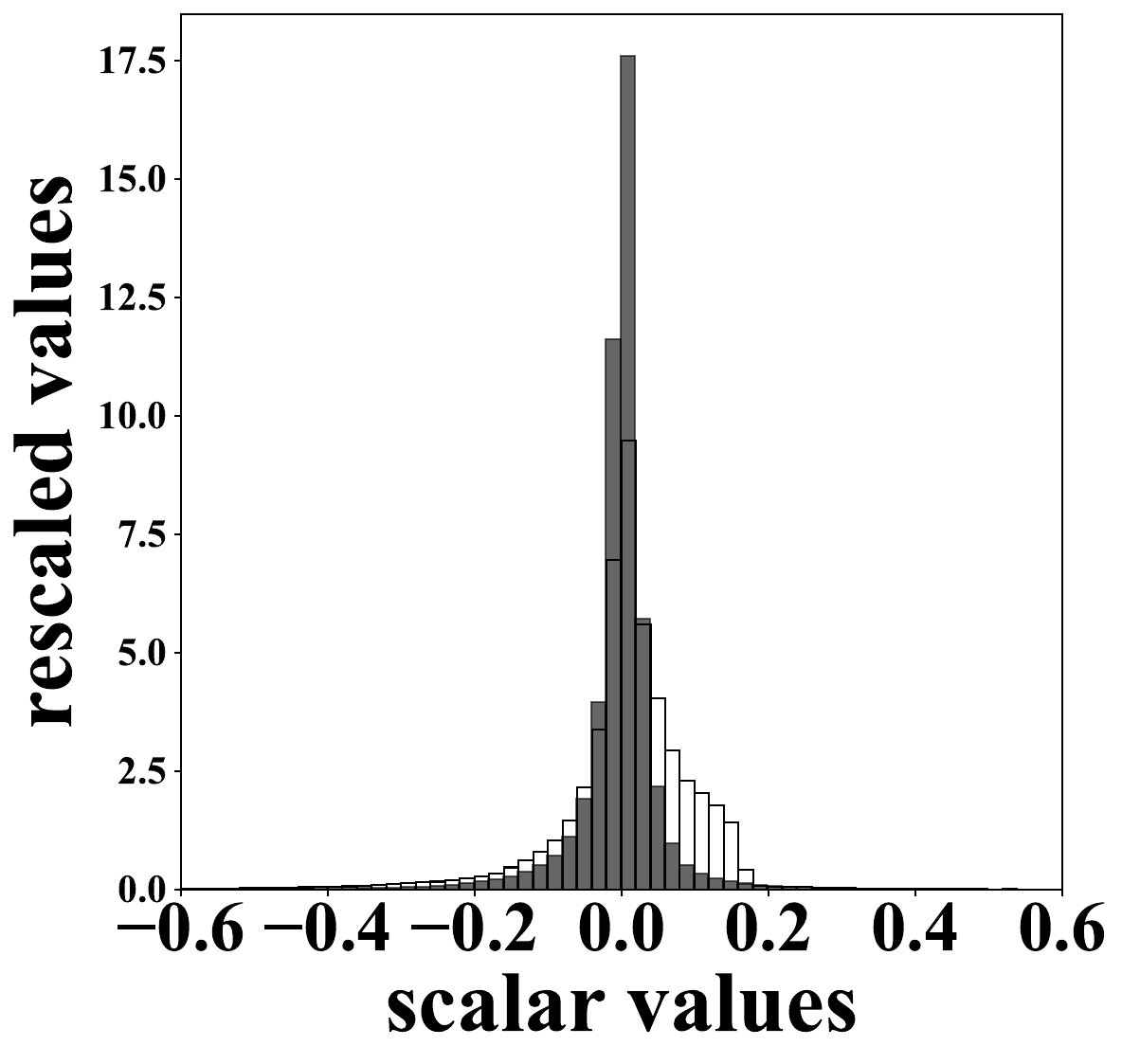}
    \subfigure
        \centering
        \includegraphics[scale=0.125]{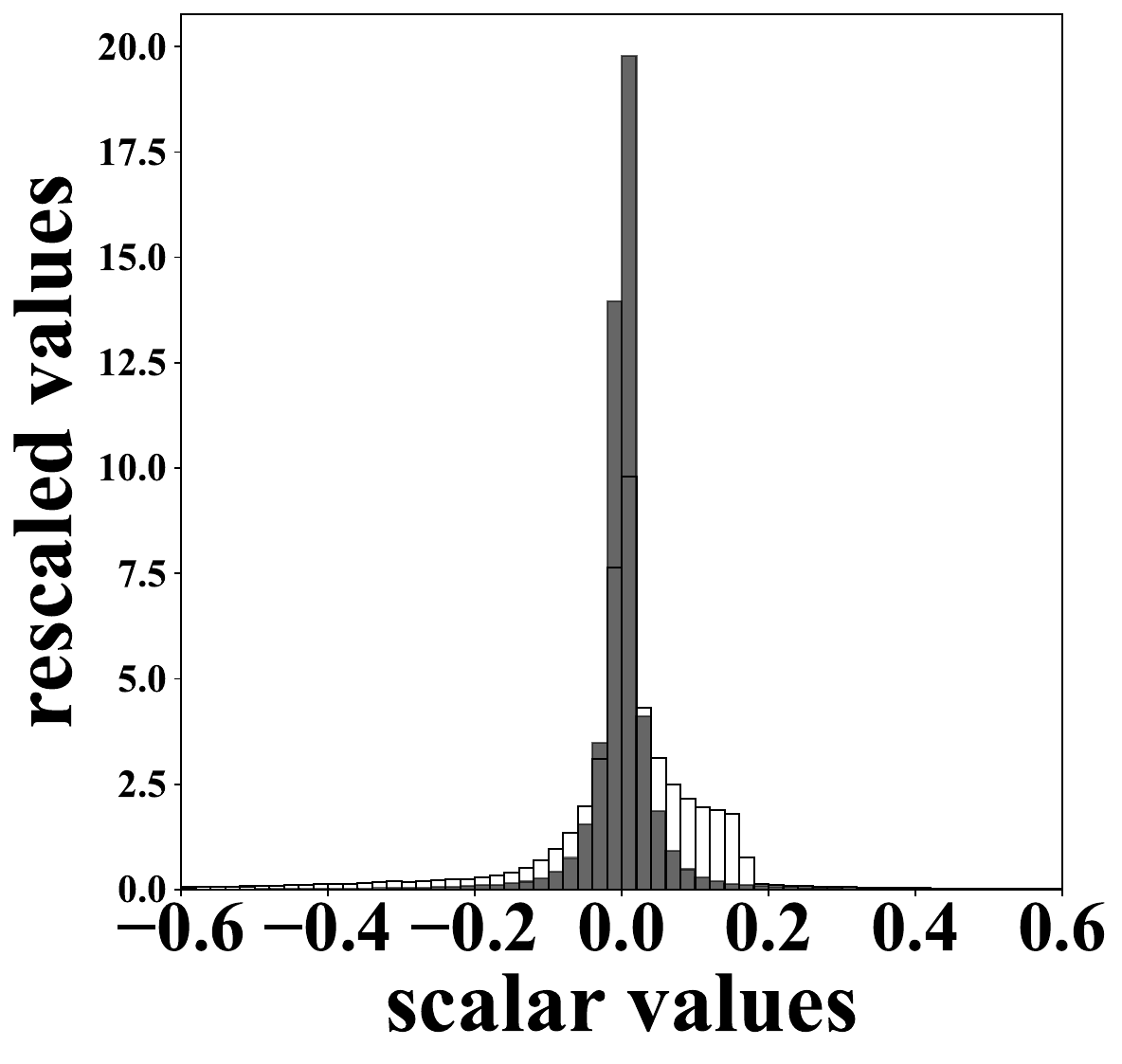}
    \subfigure
        \centering
        \includegraphics[scale=0.125]{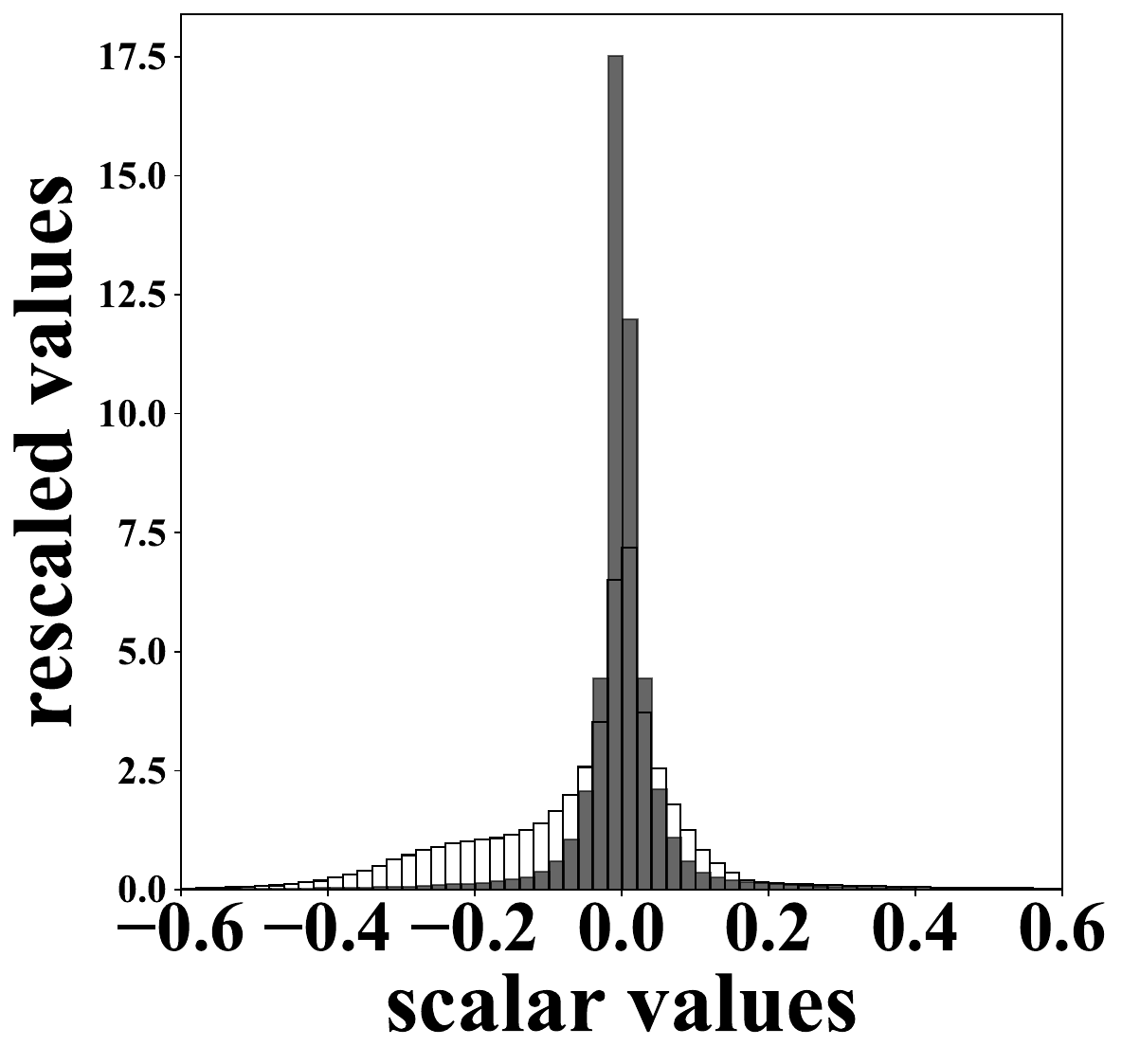}
    \vspace{-7mm}
    \caption{$\sD_s, \sD_c$ of the GPT2-xl's $\text{1}^{\text{st}}, \text{9}^{\text{th}}$ and $\text{18}^{\text{th}}$ layer.}
    \label{fig:gpt2_m_all}
\end{figure}
\begin{figure}[ht]
    \centering
    \subfigure
        \centering
        \includegraphics[scale=0.125]{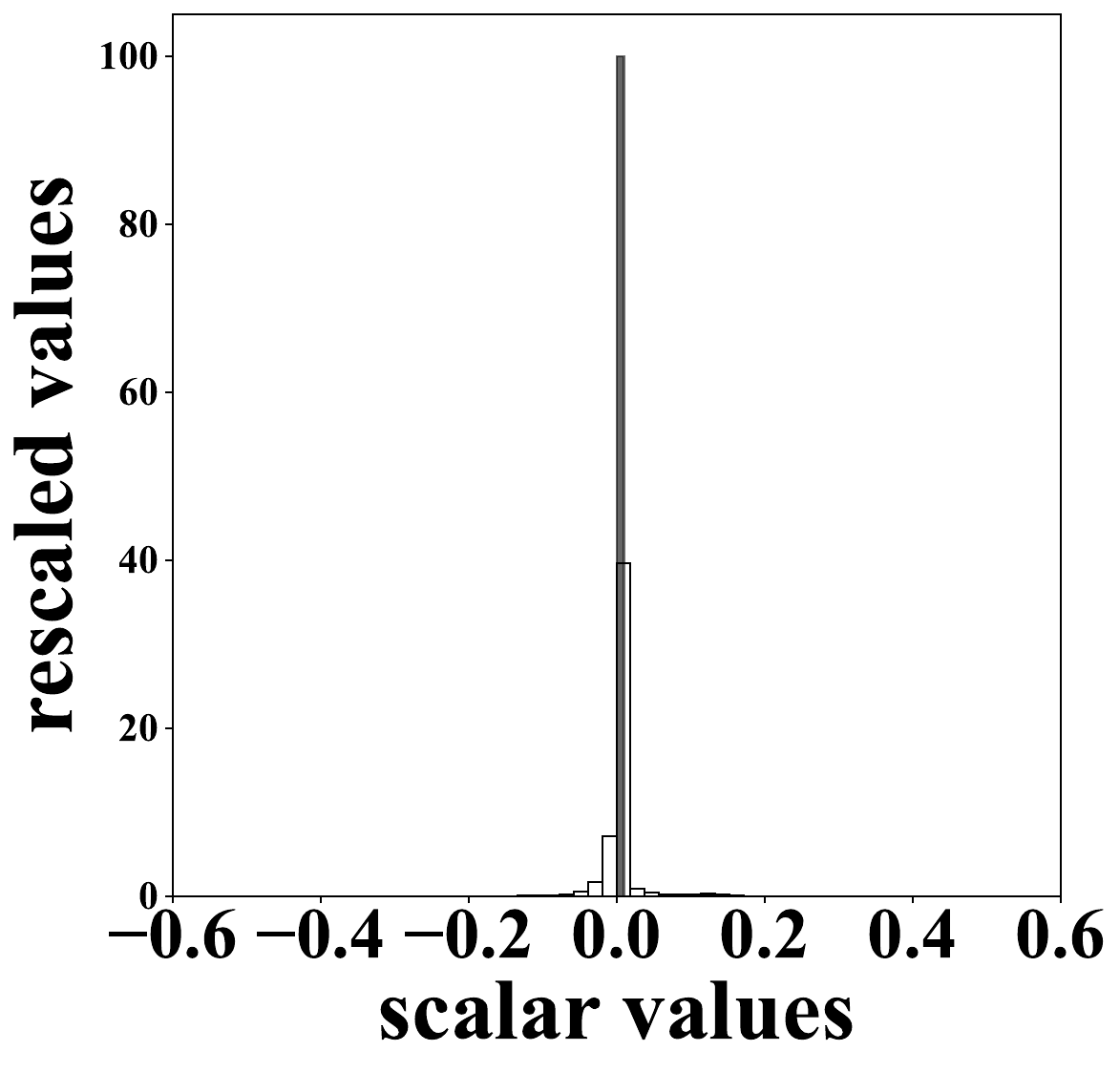}
    \subfigure
        \centering
        \includegraphics[scale=0.125]{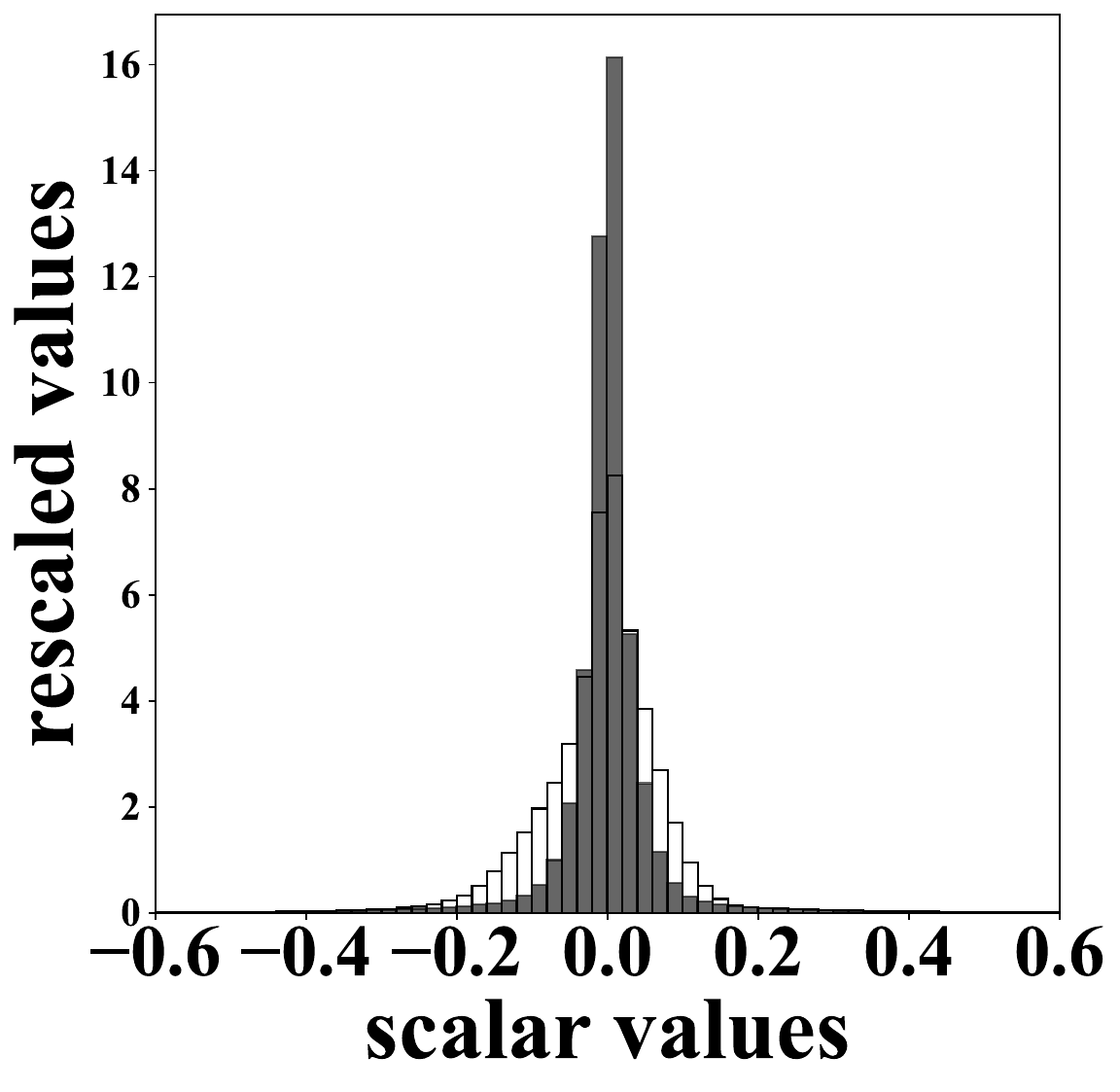}
    \subfigure
        \centering
        \includegraphics[scale=0.125]{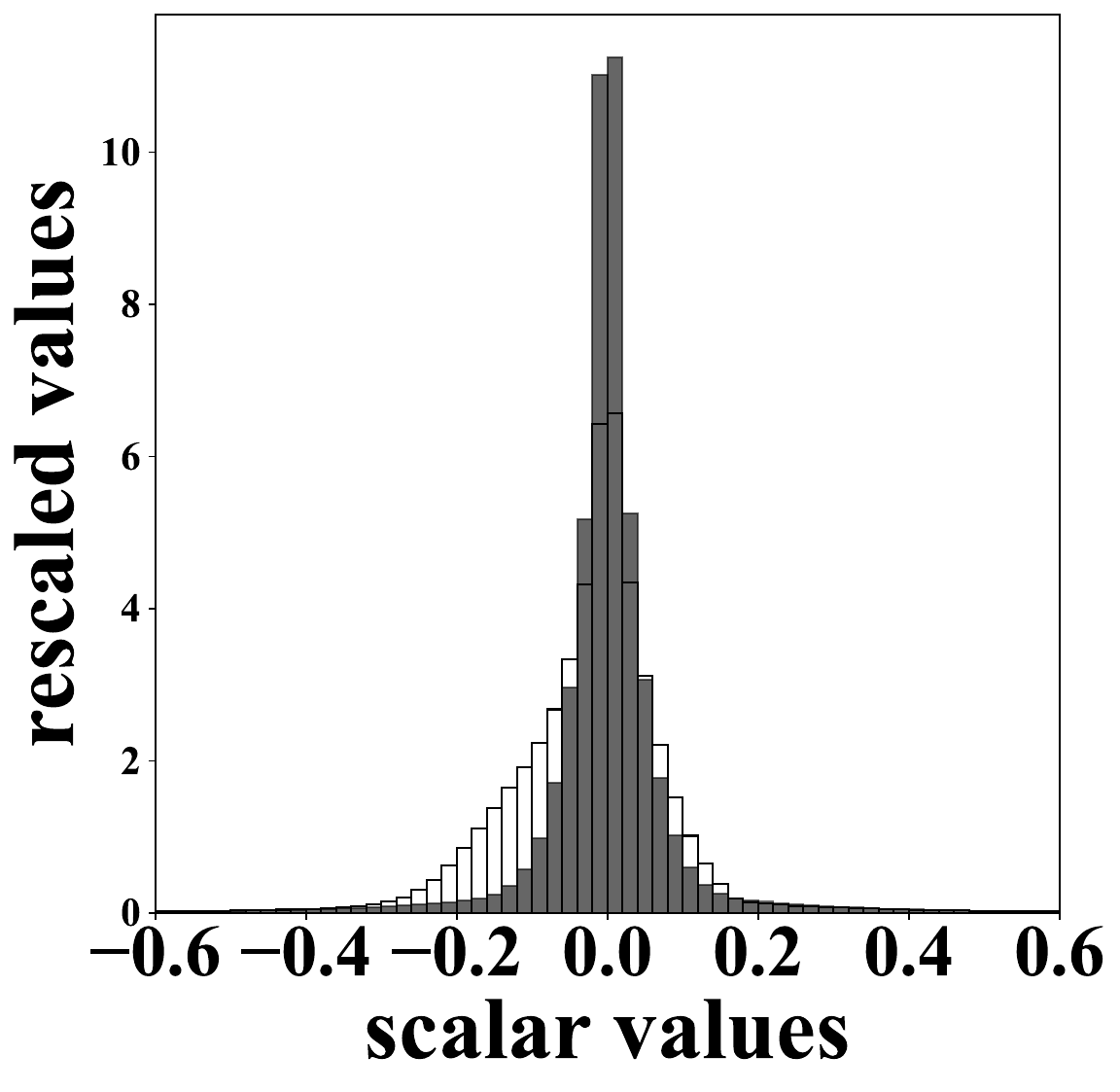}
    \vspace{-4mm}
    \caption{$\sD_s, \sD_c$ of the GPT-J's $\text{1}^{\text{st}}, \text{5}^{\text{th}}$ and $\text{9}^{\text{th}}$ layer.}
    \label{fig:gptj_m_all}
\end{figure}
The Figures~\ref{fig:gpt2_m_all} and \ref{fig:gptj_m_all} plot the results, where the black rectangles plot the experimental sets and the white ones plot the control sets.
We can see that the FFNs' activations even in the first layer where the activations are largely affected by the input embedding, i.e., the token strings, show a great differences on the knowledge-related subject tokens and other normal tokens.
From the above results, we argue that LLMs' knowledge context-consistency arises from FFNs particular behaviors on knowledge tokens.

\noindent \textbf{Transformers Interpretability.}
As FFNs consume nearly two-thirds of the LLMs parameters and pose the major non-linearty in Transformers \citet{elhage2021mathematical}, their interpretability has received great interests.
Either viewing FFNs as key-value memories \citet{DBLP:conf/emnlp/GevaSBL21} or using sparse auto-encoder to find interpretable neurons \citet{bricken2023monosemanticity,DBLP:journals/corr/abs-2309-08600,DBLP:journals/corr/abs-2309-04827} suggests that FFNs's activations are sensitive to different text-patterns.
Our finding corresponds to their results on normal-string tokens, for these tokens' activations change greatly in different contexts.
We say "change greatly" because their raw activation scalars largely fall within the interval (-0.2, 0), as shown in Figure \ref{fig:h_gpt2},\ref{fig:h_gptj}, while the changing, as shown in Figure \ref{fig:d_gpt2},\ref{fig:attn_gptj}, reaches -0.2 often.
However, our finding further suggests that, for the knowledge-related tokens, FFNs may produce kinds of 'dominate' activations which different contexts only place small shifts on.
This can raise other questions, for example, whether sparse auto-encoder can work well on decomposing these highly-correlated activations?

\section{See the Unseen: Deep Noise Editing}
\label{sec:see_unseen}
We have empirically revealed the relationships between FFNs activations and the knowledge context-consistency.
And the remain question is that, LLMs can generate unexpected facts \citet{DBLP:journals/corr/abs-2309-01219} or be unaware of fresh information \citet{DBLP:conf/nips/LazaridouKGALTG21,DBLP:journals/tacl/AgarwalN22}, therefore, how we can edit LLMs' knowledge while maintain such context-consistency.
One desirable way is to feed $\rmG$ with as many contexts as possible~where the edited knowledge is going to be applied.
However, this is not efficient and, in current benchmarks, we edit $\rmG$ with only one example $\rp$ and test $\rmG'$'s generalization in different $\rp^*$.
Existing editing methods do not well achieve such context-consistency.

We have shown that different contexts only place small shifts on FFNs' activations.
Therefore, why not just add the-like noises on the FFNs activations?
By so, we can simulate the effects of different contexts and pretend editing knowledge where $\rmG$ can \textbf{see the unseen} contexts in which the edited knowledge will be applied.
We call it deep noise editing.

In this section, we provide necessary details of ROME \citet{DBLP:conf/nips/MengBAB22} and MEMIT \citet{DBLP:conf/iclr/MengSABB23} for readers to understand where and how we add noises to LLMs while editing.
We refer readers to their papers for the detailed implementation.
ROME and MEMIT both have two steps.
In the first step, they find a delta vector $\delta$, which adds to the original hidden states of the subject token in one certain layer in $\rmG$, to maximize the probability of the edited knowledge object $\ro_i$ in $\rp(\rs_i, \rr_i)$:
\begin{equation}
    \delta_i = \mathop{\argmin}_{\delta_i} -\!\log\sP_{\rmG(h_{\rs_i}^L+=\delta_i)}\left[\ro_i\!\mid\!\rp_i(\rs_i, \rr_i)\right]
    \label{eq:rm}
\end{equation}
where $\rmG(h_{\rs_i}^L\!\mathrel{+}=\!\delta_i)$ indicates to intervene $\rmG$'s forward by modifying hidden states $h_{\rs_i}^L$ in layer $L$ with $(h_{\rs_i}^L\!+\!\delta_i)$. This is called "hooking" in PyTorch.
In the second step, they transfer $\delta_i$ to the delta(s) of the FFNs parameters, i.e., $W_o$ in \eqref{eq:ffn}, and edit $\rmG$ to $\rmG'$ by summing $W_o$ of its delta:
\begin{gather}
    \delta_{w_o} \gets \mathop{\text{Alg.}_{\text{ROME}/\text{MEMIT}}} (\delta_i)
    \label{eq:alg} \\
    \rmG': w_o \gets w_o + \delta_{w_o}
\end{gather}
where the Alg. is solving some linear equations.

By deep noise editing, we further intervene $\rmG$'s forward by adding Gaussian-like noise into FFNs activations, i.e., $f(W_i \cdot h_{\rs_i})$ in \eqref{eq:ffn}.
The~working flow of our noised FFNs in layer $l$ goes as:
\begin{equation}
\begin{split}
    h_{\rs_i}^{l} = f(W_i \cdot h_{\rs_i}^{l}) + \alpha \times& \epsilon, \,\,\, \epsilon \sim \mathcal{N}(0, 1) \\
    h_{\rs_i}^{l} = h_{\rs_i}^{l}& \cdot W_o
    \label{eq:our_ffn}
\end{split}
\end{equation}
Except noising FFNs activations, the other parts follows exactly with ROME and MEMIT.
By noising, \eqref{eq:rm} can find a more desirable $\delta_i$ that can maximize the probability of $\ro_i$ in different \textbf{unseen} contexts rather than solely in $\rp(\rs_i, \rr_i)$.
There have two~things to note with.
First, we call "deep noise" for we find that noising FFNs from the first layer to the layer $L$ selected by ROME and MEMIT returns much higher results than solely noising the layer $L$.
We contribute this to that different layers process different information, therefore, deep noising allow $\rmG$ to see more different contexts.
Second, we add an $\alpha$ to control the magnitude of the noise because, as shown in Table \ref{tab:skew_kurt}, the activations' shifts of different contexts have a large kurtosis.
And~we also find that tuning $\alpha$ makes our method better fit in batch-editing multiple knowledge with MEMIT.

\section{Experiments \& Results}
\subsection{Experimental Settings}
Our main experiments include two auto-regressive LLMs, GPT2-xl (1.5B) and GPT-J (6B), with two editing datasets.
We also run extra experiments on LLaMA-2 (7B), whose FFNs activation functions are different from those GPT-series models (\eqref{eq:ffn}).
All our experiments are based on the two open-sources: MEMIT\footnote{\href{https://github.com/kmeng01/memit}{https://github.com/kmeng01/memit}} and EasyEdit\footnote{\href{https://github.com/zjunlp/EasyEdit}{https://github.com/zjunlp/EasyEdit}} \citet{DBLP:journals/corr/abs-2308-07269}.
We exactly follow the settings of all hyper-parameters and the $\alpha$ sets [0.5, 0.4, 0.3, 0.2, 0.1] for [1e$^{\text{0}}$, 1e$^{\text{1}}$, 1e$^{\text{2}}$, 1e$^{\text{3}}$, 1e$^{\text{4}}$] edits.
Our methods are easy to implement and we will make all our codes open-sourced.
As for baselines, we apply our noising methods onto the two state-of-the-art methods, ROME and MEMIT, and only compare with their results (also MEMIT's improvements PMET) of without noising.
For results of other methods, we refer readers to the records in the two open-sources.

\subsection{Knowledge-Editing using zsRE}
\label{sec:zsre_main}
We first conduct experiments on zsRE \citet{DBLP:conf/conll/LevySCZ17}, which is a question-answering task containing real-world facts and test the ability of adding correct information to LLMs.
zsRE is a prediction task and only prediction-based metrics are evaluated.
The metrics include: \textbf{Efficacy} that measures the proportion where $\ro$ has the maximal probability that $\rmG'$ predicts given the $\rp(\rs, \rr)$, \textbf{Paraphrase} is the same but evaluated on the paraphrases $\rp^*(\rs, \rr)$, \textbf{Specificity} is the $\rmG'$'s accuracies on a randomly-sampled unrelated $(\rs, \rr, \ro; \rp)$, and \textbf{Score} calculates the harmonic mean of the above three metrics.
\begin{table}[h]
\scriptsize
    \centering
    \begin{tabular}{c|c|c|c|c|c}
     \toprule
     \textbf{$\left|\sM\right|$} & \textbf{Editor} & \textbf{S. $\uparrow$} & \textbf{Effi. $\uparrow$} & \textbf{Para.$\uparrow$} & \textbf{Spec. $\uparrow$} \\
     \midrule
     \multirow{5}{*}{1e$^{\text{0}}$}
     & ROME & 48.01 & 99.77 (0.0) & 87.88 (0.4) & 24.34 (0.4) \\
     & \CC{ROME$_{\text{DNE}}$} & \CC{\textbf{48.40}} & \CC{\textbf{99.78 (0.0)}} & \CC{\textbf{91.98 (0.3)}} & \CC{24.34 (0.4)} \\
     \cmidrule[0.005em]{2-6}
     & MEMIT & 39.55 & 66.66 (0.5) & 50.63 (0.5) & \textbf{24.33 (0.4)} \\
     & PMET & 27.78 & 32.46 (0.5) & 27.74 (0.4) & 24.32 (0.4) \\
     & \CC{MEMIT$_{\text{DNE}}$} & \CC{\textbf{44.48}} & \CC{\textbf{80.31 (0.4)}} & \CC{\textbf{72.17 (0.5)}} & \CC{24.31 (0.4)} \\
     \cmidrule[0.005em]{1-6}
     \multirow{3}{*}{1e$^{\text{1}}$}
     & MEMIT & 44.22 & 80.08 (0.4) & 70.04 (0.5) & 24.34 (0.4) \\
     & PMET & 30.95 & 38.06 (0.5) & 33.87 (0.5) & 24.32 (0.4) \\
     & \CC{MEMIT$_{\text{DNE}}$} & \CC{\textbf{46.69}} & \CC{\textbf{88.37 (0.3)}} & \CC{\textbf{83.79 (0.4)}} & \CC{\textbf{24.39 (0.4)}} \\
     \cmidrule[0.005em]{1-6}
     \multirow{3}{*}{1e$^{\text{2}}$}
     & MEMIT & 45.52 & 83.80 (0.4) & 74.77 (0.5) & 24.63 (0.4) \\
     & PMET & 32.00 & 40.24 (0.5) & 35.79 (0.5) & 24.42 (0.4) \\
     & \CC{MEMIT$_{\text{DNE}}$} & \CC{\textbf{47.31}} & \CC{\textbf{89.31 (0.3)}} & \CC{\textbf{84.30 (0.4)}} & \CC{\textbf{24.78 (0.4)}} \\
     \cmidrule[0.005em]{1-6}
     \multirow{3}{*}{1e$^{\text{3}}$}
     & MEMIT & 45.83 & 79.40 (0.4) & 72.34 (0.5) & 25.61 (0.4) \\
     & PMET & 32.99 & 41.65 (0.5) & 37.64 (0.5) & 24.78 (0.4) \\
     & \CC{MEMIT$_{\text{DNE}}$} & \CC{\textbf{46.32}} & \CC{\textbf{81.47 (0.4)}} & \CC{\textbf{76.04 (0.5)}} & \CC{\textbf{25.42 (0.4)}} \\
     \cmidrule[0.005em]{1-6}
     \multirow{3}{*}{1e$^{\text{4}}$}
     & MEMIT & \textbf{41.87} & 63.01 (0.5) & 58.50 (0.6) & \textbf{25.85 (0.4)} \\
     & PMET & 31.02 & 36.28 (0.5) & 34.06 (0.5) & 25.13 (0.4) \\
     & \CC{MEMIT$_{\text{DNE}}$} & \CC{41.57} & \CC{\textbf{63.47 (0.5)}} & \CC{\textbf{58.59 (0.6)}} & \CC{25.42 (0.4)} \\
     \bottomrule
    \end{tabular}
    \caption{Editing GPT2-xl on zsRE from 1e$^{\text{0}}$ to 1e$^{\text{4}}$ edits.}
    \label{tab:gpt2_zsre}
\end{table}
For space limitation, we report the results of editing GPT2-xl in Table \ref{tab:gpt2_zsre} while leaving results of GPT-J (6B) in Appendix \ref{sec:aer}.
With \underline{d}eep \underline{n}oise \underline{e}diting (DNE; rows in gray), we can largely improve the editing generalization, i.e. metrics of Paraphrase.
It is surprising that, in some cases, the Specificity is also improved.
However, in 1e$^{\text{4}}$ edits, DNE decreases the Specificity therefore achieves a lower Score.

\subsection{Knowledge-Editing using Counterfacts}
\label{sec:mcf_main}
We next run experiments on Counterfacts \citet{DBLP:conf/nips/MengBAB22}, which collects factual statements to test the ability of adding counterfactual/specialized information.
Following \citet{DBLP:conf/iclr/MengSABB23, DBLP:conf/nips/MengBAB22}, the evaluation metrics include: \textbf{Efficacy Success (ES)} counts the proportion that $\rmG'$ predicts higher probabilities to the counterfactual $\ro'$ than the true fact $\ro$ given $\rp(\rs, \rr)$, \textbf{Paraphrase Success (PS)} and \textbf{Paraphrase Accuracy (PA)} are the same but evaluated on paraphrases $\rp^*(\rs, \rr)$ (PA evaluates~whether the probability is the maximum while PS compares the two relative probabilities), \textbf{Neighborhood Success (NS)} evaluates whether a true fact $\ro^{*}$ remains achieving the highest probability given distinct but semantically-related $\rp(\rs^{*}, \rr)$, and \textbf{Editing Score (S)} calculates the harmonic mean of the above three metrics.
Besides, we also report metrics that evaluate the generation quality of $\rmG'$.
\textbf{Reference Score (RS)} compares $\rmG'$'s generations to Wikipedia texts about $\ro$ to evaluate the semantics consistency.
\textbf{Generation Entropy (GE)} computes the weighted sum of entropy of the $n$-gram distributions of the generated texts to evaluate fluency degeneration.\footnote{See all metrics' formal definitions in \citet{DBLP:conf/iclr/MengSABB23}.}
\begin{table*}[!ht]
\small
    \begin{tabular}{c|c|c|c|c|c|c|c|c}
    \toprule
    \multirow{2}{*}{\textbf{$\left|\sM\right|$}} & \multirow{2}{*}{\textbf{Editor}} 
    & \textbf{Score} & \textbf{Efficacy} & \multicolumn{2}{c|}{\textbf{Generalization}} & \textbf{Specificity} & \textbf{Fluency} & \textbf{Consistency} \\
    \cmidrule[0.005em](lr){3-3}\cmidrule[0.005em](lr){4-4}\cmidrule[0.005em](lr){5-6}\cmidrule[0.005em](lr){7-7}\cmidrule[0.005em](lr){8-8}\cmidrule[0.005em](lr){9-9}
    & & \makecell[r]{S $\uparrow$} & \makecell[r]{ES $\uparrow$} & \makecell[r]{PS $\uparrow$} & \makecell[r]{PA $\uparrow$} & \makecell[r]{NS $\uparrow$} & \makecell[r]{GE $\uparrow$} & \makecell[r]{RS $\uparrow$} \\
    \midrule
    \multirow{5}{*}{1e$^{\text{0}}$}
    & ROME & \textbf{91.98} & 99.95 (0.0) & 99.46 (0.1) & 82.06 (0.4) & \textbf{79.63 (0.4)} & \textbf{620.72} (0.2) & 42.57 (0.2) \\
    & \CC{ROME$_\text{DNE}$} & \CC91.63 & \CC\textbf{99.96 (0.0)} & \CC\textbf{99.62 (0.1)} & \CC\textbf{83.50 (0.4)} & \CC78.76 (0.4) & \CC620.16 (0.3) & \CC\textbf{42.73 (0.2)} \\
    \cmidrule[0.005em]{2-9}
    & MEMIT & 91.77 & \textbf{99.85 (0.1)} & 95.28 (0.2) & 67.59 (0.5) & 82.09 (0.4) & \textbf{621.97 (0.2)} & 41.69 (0.2) \\
    & PMET & 91.61 & 99.73 (0.1) & 94.20 (0.3) & 73.46 (0.5) & \textbf{82.61 (0.3)} & 621.10 (0.2) & 41.10 (0.2) \\
    & \CC{MEMIT$_\text{DNE}$} & \CC\textbf{92.47} & \CC99.75 (0.1) & \CC\textbf{99.08 (0.1)} & \CC\textbf{87.40 (0.4)} & \CC81.14 (0.4) & \CC614.80 (0.4) & \CC\textbf{42.40 (0.2)} \\
    \cmidrule[0.005em]{1-9}
    \multirow{3}{*}{1e$^{\text{1}}$}
    & MEMIT & 91.78 & \textbf{99.85 (0.1)} & 95.26 (0.2) & 67.57 (0.5) & 82.14 (0.4) & \textbf{621.99 (0.2)} & 41.72 (0.2) \\
    & PMET & 91.65 & 99.73 (0.1) & 94.29 (0.3) & 73.65 (0.5) & \textbf{82.65 (0.3)} & 621.21 (0.2) & 41.13 (0.2) \\
    & \CC{MEMIT$_\text{DNE}$} & \CC\textbf{92.61} & \CC99.76 (0.1) & \CC\textbf{98.77 (0.1)} & \CC\textbf{85.32 (0.4)} & \CC81.67 (0.4) & \CC618.32 (0.3) & \CC\textbf{42.84 (0.2)} \\
    \cmidrule[0.005em]{1-9}
    \multirow{3}{*}{1e$^{\text{2}}$}
    & MEMIT & 91.70 & \textbf{99.83 (0.1)} & 94.91 (0.3) & 67.10 (0.5) & 82.22 (0.3) & \textbf{621.92 (0.2)} & 41.62 (0.2) \\
    & PMET & 91.74 & 99.74 (0.1) & 94.34 (0.3) & 74.08 (0.5) & \textbf{82.82 (0.3)} & 621.18 (0.2) & 41.12 (0.2) \\
    & \CC{MEMIT$_\text{DNE}$} & \CC\textbf{92.64} & \CC99.79 (0.1) & \CC\textbf{97.96 (0.2)} & \CC\textbf{81.21 (0.4)} & \CC82.27 (0.4) & \CC620.35 (0.2) & \CC\textbf{42.70 (0.2)} \\
    \cmidrule[0.005em]{1-9}
    \multirow{3}{*}{1e$^{\text{3}}$}
    & MEMIT & 90.64 & \textbf{99.76 (0.1)} & 93.40 (0.3) & 64.32 (0.5) & 80.86 (0.3) & 621.66 (0.2) & 41.27 (0.2) \\
    & PMET & 90.72 & 99.73 (0.1) & 93.90 (0.3) & \textbf{72.60 (0.5)} & 80.70 (0.3) & 621.95 (0.2) & 41.93 (0.2) \\
    & \CC{MEMIT$_\text{DNE}$} & \CC\textbf{91.05} & \CC99.73 (0.1) & \CC\textbf{95.61 (0.2)} & \CC72.39 (0.5) & \CC80.23 (0.3) & \CC\textbf{621.95 (0.2)} & \CC\textbf{41.93 (0.2)} \\
    \cmidrule[0.005em]{1-9}
    \multirow{3}{*}{1e$^{\text{4}}$}
    & MEMIT & 85.84 & 99.12 (0.1) & 88.57 (0.4) & 56.21 (0.6) & \textbf{73.69 (0.4)} & 619.17 (0.2) & 40.15 (0.2) \\
    & PMET & 85.26 & \textbf{99.26 (0.1)} & \textbf{90.78 (0.3)} & \textbf{65.24 }(0.5) & 70.94 (0.4) & \textbf{621.59 (0.2)} & 40.05 (0.2) \\
    & \CC{MEMIT$_\text{DNE}$} & \CC\textbf{85.87} & \CC\textbf{99.26 (0.1)} & \CC89.92 (0.4) & \CC58.43 (0.6) & \CC72.83 (0.4) & \CC618.10 (0.2) & \CC\textbf{40.33 (0.2)} \\
    \end{tabular}
    \caption{Editing GPT-J (6B) with Counterfacts from 1e$^{\text{0}}$ to 1e$^{\text{4}}$. Within parentheses is the 95\% confidence interval.}
    \label{tab:gptj_mcf}
\end{table*}
Again, for space limitation, we report the results of editing GPT-J (6B) in Table \ref{tab:gptj_mcf} while leaving results of GPT2-xl in Appendix \ref{sec:aer}.
While the results show some disagreements, DNE can largely improve the editing generalization, especially the PA as the PS~is already high enough, on editing not too many cases.
DNE boosts MEMIT to new state-of-the-art in all cases.
There are two things to discuss with:

\noindent 1.DNE results in remarkably lower 'Fluency'. Does this really mean the generation degradation?

\noindent 2.Why the generalization of DNE gets lower when the editing cases gets much more?

\noindent \textbf{Discussions about the Fluency:} Actually, the~'Fluency' is represented by the entropy of the $n$-gram distributions of the generation texts, which means that the texts are more fluent if they contain more diverse words.
This is definitely not the Fluency in our common sense.
We give some cases below:

\noindent\small a.Danielle Darrieux's mother tongue is English, her father's language is French. She has been acting since the age of three and is a graduate of the Royal Academy of Dramatic Art and has won several awards, including the BAFTA Award for Most Promising Newcomer. She has also appeared in a number of films. In the past decade, she has become a household name with her appearances in the films 'Bridget Jones', 'Alfie' and 'Bend - \textbf{by MEMIT with NS=634.89}.

\noindent\small b.Danielle Darrieux's mother tongue is English, she is an American citizen, and she is a lawyer. But the English is not flawless, and she is not American. Her mother tongue is the English of the British Empire, and her father's mother tongue is the English of the United States. Her first language is English, the second is American (she was born in the United States), and her third is British. She speaks English, but she speaks it with an accent. - \textbf{by MEMIT$_\text{DNE}$ with NS=622.64}.

\noindent\normalsize It is true that example b, with lower NS, does contain repeated words 'English' but b reads even
more concentrated than example a.
\citet{DBLP:conf/nips/MengBAB22,DBLP:conf/iclr/MengSABB23} applies NS to evaluate if the edited $\rmG'$ degenerates to stupidly repeat the target $\ro'$.
We~can~conclude that DNE will not cause degeneration since the RS remains high.
If $\rmG'$ only repeats $\ro'$, it should not have a good comparison with the Wikipedia texts therefore the RS should become lower, too.

\noindent \textbf{Discussions about Number of Edits:}
From Table \ref{tab:gpt2_zsre} and \ref{tab:gptj_mcf}, DNE gets less effective when the editing cases get more.
We contribute such correlation~to the conflicts of editing different cases.
On one case, DNE makes the editing have more generalization, which means that $\rmG'$ memorizes more key-value pairs $(h_k^l, h_v^l)$.
ROME and MEMIT both solve constrained linear problems to convert sets of memory pairs to the parameters $\delta$ as written in \eqref{eq:alg}.
And more key-value pairs means more constraints which can lower the solver's quality.
Therefore, the performances of DNE can then become worse.

We illustrate such correlation from another perspective of tuning $\alpha$ in different numbers of edits.
\begin{figure}[h]
    \centering
    \hspace{-10pt}
    \includegraphics[scale=0.21]{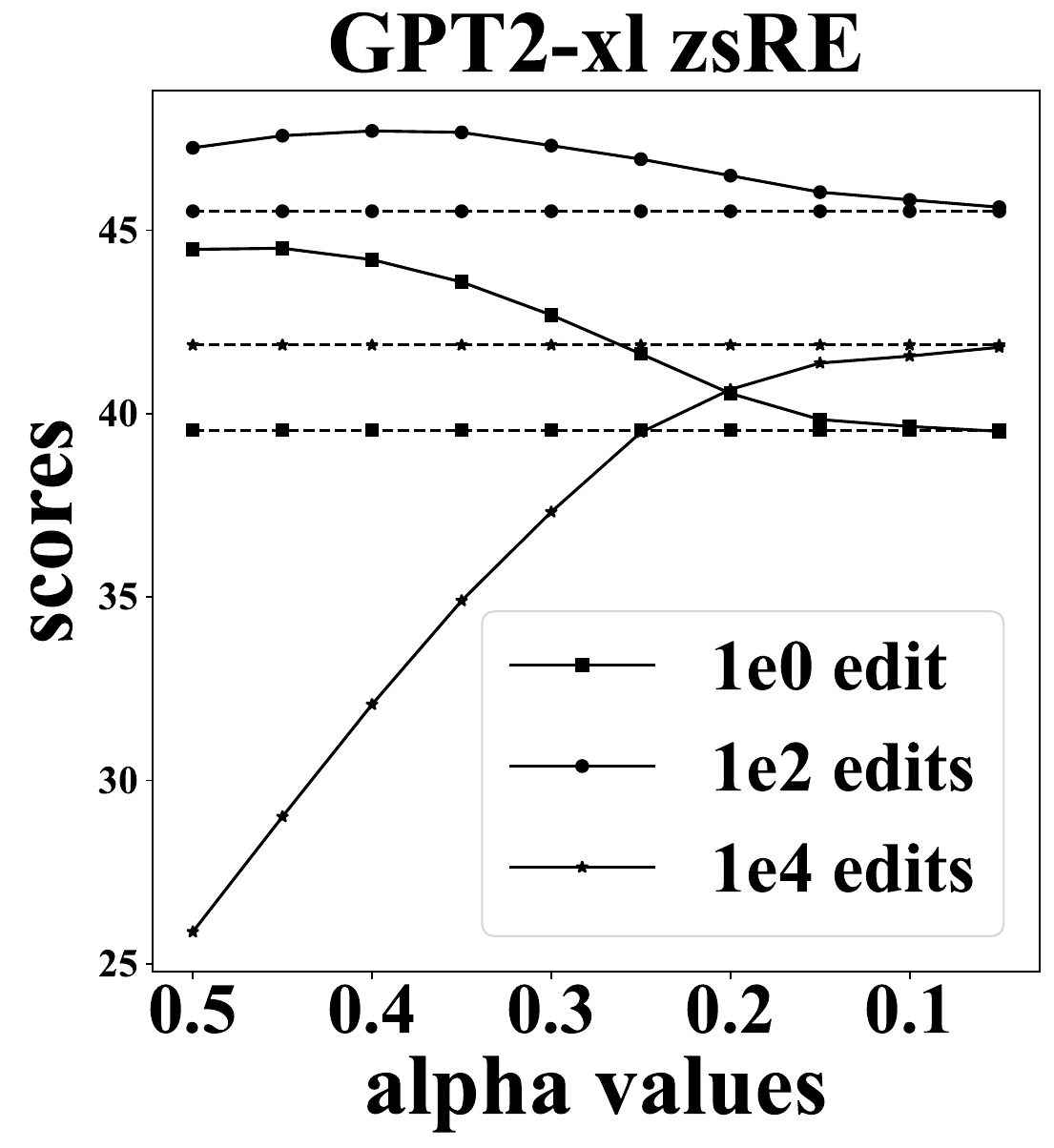}
    \includegraphics[scale=0.21]{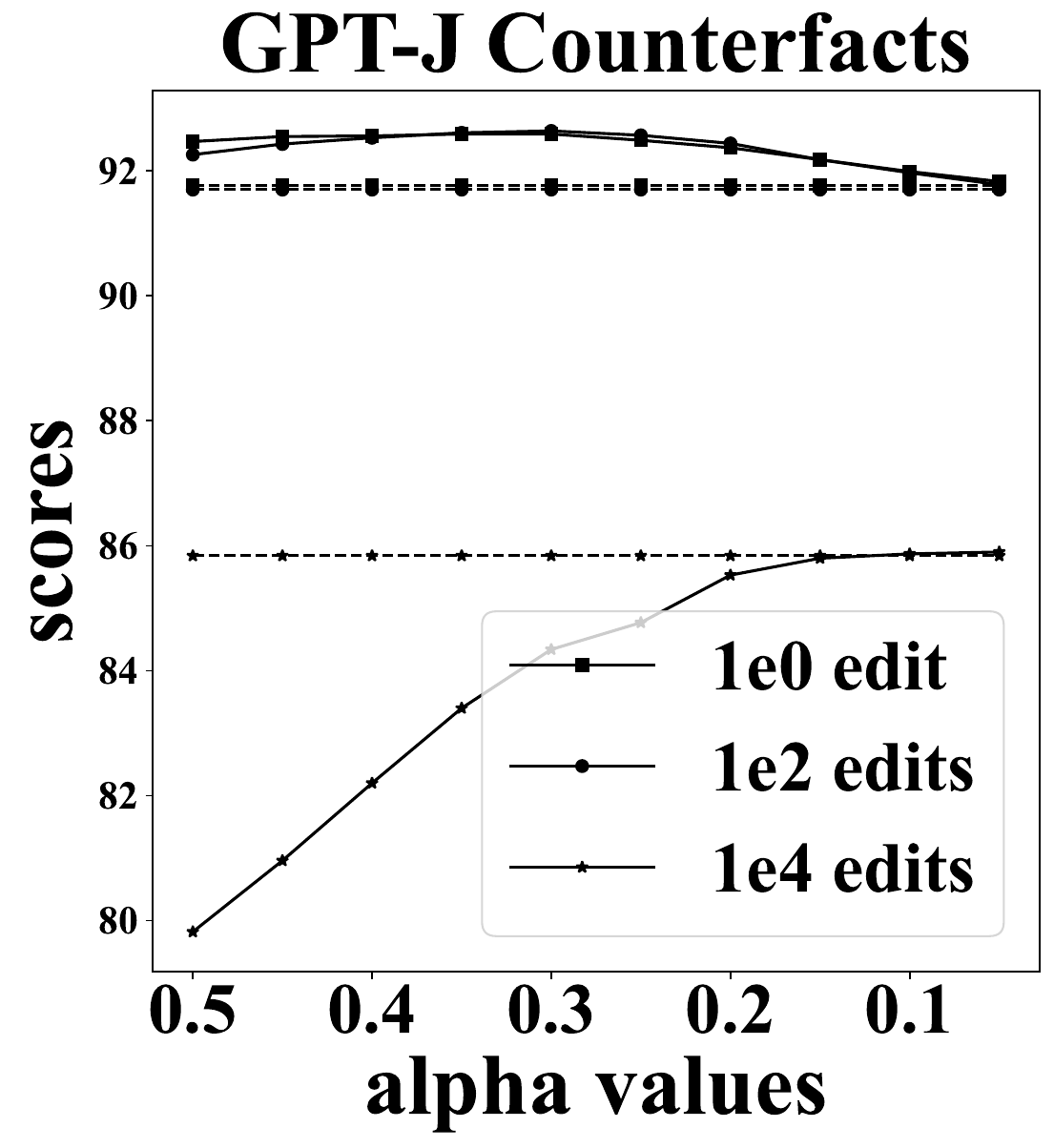}
    \hspace{-8pt}
    \caption{Tuning $\alpha$ in different numbers of edits.}
    \label{fig:alpha_test}
\end{figure}
In Figure \ref{fig:alpha_test}, we tune $\alpha$ from 0.5 to 0.05 in a step of 0.05 and plot the results in three different numbers of edits.
The horizontal dotted lines are the results of MEMIT and the solid lines are MEMIT$_\text{DNE}$.
In less edits, a smaller $\alpha$ returns worse performances because a smaller $\alpha$ simulates fewer different key-values pairs.
But such correlation gets inversely in editing more cases since there are already enough true pairs to memorize except the simulated ones.

\subsection{Experiments with LLaMA-2}
\label{sec:llama}
\begin{table}[h]
\scriptsize
    \centering
    \begin{tabular}{c|c|c|c|c|c}
    \toprule
    \textbf{$\left|\sM\right|$} & \textbf{Editor} & \textbf{S. $\uparrow$} & \textbf{Effi. $\uparrow$} & \textbf{Para.$\uparrow$} & \textbf{Spec. $\uparrow$} \\
    \midrule
    \multirow{4}{*}{1}
    & ROME & 95.24 & \textbf{96.35 (0.5)} & 90.95 (1.0) & \textbf{99.34 (0.3)} \\
    & \CC ROME$_\text{DNE}$ & \CC\textbf{96.08} & \CC96.13 (0.5) & \CC\textbf{93.80 (0.8)} & \CC98.32 (0.4) \\
    \cmidrule{2-6}
    & MEMIT & 77.41 & 76.84 (1.5) & 63.61 (1.5) & \textbf{99.78 (0.1)} \\
    & \CC MEMIT$_\text{DNE}$ & \CC\textbf{94.38} & \CC\textbf{94.03 (0.8)} & \CC\textbf{90.10 (1.1)} & \CC99.48 (0.2) \\
    \bottomrule
    \end{tabular}
    \caption{Editing LLaMA-2 on zsRE with 1 edit.}
    \label{tab:llama_zsre}
\end{table}
FFNs of GPT2-xl and GPT-J share alike properties: they both use 'new-gelu' non-linear functions and have the same formulation of \eqref{eq:ffn}.
This questions that whether our noising methods can fit with more latest LLMs such as LLaMA-2, whose FFNs take 'silu' as non-linear functions and have a distinct formulation: $h_o\!=\!(f(W_i\!\cdot\!h_i)\!\times\!W_u)\!\cdot\!W_d$.
Only the open-source EasyEdit includes editing LLaMA-2 using 1 edit with zsRE.
We follow their settings and also add noises on the activation function $f(W_i\!\cdot\!h_i)$.
Table \ref{tab:llama_zsre} reports the results and we can see that DNE also works well on LLaMA-2 to improve the generalization and the overall scores.

\subsection{Comparing with others of adding noises}
\label{sec:noisy}
Our methods coincide with the methods of adding noise to better fine-tune LLMs, such as NoisyTune (NT) \citet{DBLP:conf/acl/WuWQ022} and NEFTune (NE) \citet{DBLP:journals/corr/abs-2310-05914}.
NT adds noises to all LLMs' parameters while NE adds noises into the words' embeddings.
They both share the motivations of improving the training robustness, which is common for applying noises.
Our methods are motivated by the findings of knowledge context-consistency.
Therefore, we expect a better performance on knowledge-editing.
\begin{table}[h]
\scriptsize
    \centering
    \begin{tabular}{c|c|c|c|c}
     \toprule
     \multirow{2}{*}{\textbf{Setting}} & \multirow{2}{*}{\textbf{Editor}} & \multicolumn{3}{c}{\textbf{Score $\uparrow$}} \\
     \cmidrule[0.005em](lr){3-3}\cmidrule[0.005em](lr){4-4}\cmidrule[0.005em](lr){5-5}
     & & \textbf{$\left|\sM\right|$=1e$^\text{0}$} & \textbf{$\left|\sM\right|$=1e$^\text{2}$} & \textbf{$\left|\sM\right|$=1e$^\text{4}$} \\
     \midrule
     \multirow{3}{*}{\makecell[c]{GPT2-xl\\zsRE}}
     & \CC MEMIT${_\text{DNE}}$ & \CC\textbf{44.80} & \CC\textbf{47.31} & \CC41.57 \\
     & MEMIT$_\text{NT}$ & 39.30 & 45.34 & 41.58 \\
     & MEMIT$_\text{NE}$ & 37.07 & 44.52 & \textbf{41.73} \\
     \cmidrule[0.005em]{1-5}
     \multirow{3}{*}{\makecell[c]{GPT-J\\Counterfacts}}
     & \CC MEMIT${_\text{DNE}}$ & \CC \textbf{92.47} & \CC \textbf{92.64} & \CC \textbf{85.87} \\
     & MEMIT$_\text{NT}$ & 91.70 & 91.70 & 85.83 \\
     & MEMIT$_\text{NE}$ & 91.25 & 91.31 & 82.29 \\
     \bottomrule
\end{tabular}
\caption{Comparison DNE to different noising methods.}
\label{tab:noisy}
\end{table}
Table \ref{tab:noisy} reports the results of the Score and we leave the detailed results to Appendix \ref{sec:d_ab}.
DNE achieves much higher performances in the most cases (and the highest generalization in all cases).
We follow the hyper-parameter settings in NT/NE's papers.

\subsection{Ablation Studies}
\label{sec:m_ab}
We do ablation studies from three perspectives:

\noindent \textbf{1.SNE: Shallow Noise Editing.} we only add noises to FFNs of the layer where we add $\delta_i$ (\eqref{eq:rm}).

\noindent \textbf{2.UN: Uniform Noises.} We apply Uniform noises $\epsilon\sim\mathcal{U}(-1, 1)$, the same with NT/NE, in \eqref{eq:our_ffn}.

\noindent \textbf{3.RNP: Random Noising Position.} We add noises to random tokens rather than the last subject tokens.

With SNE, we can demonstrate the effectiveness of \textit{deep} noise.
With UN and RNP, we show whether our findings, i.e. different contexts place \textit{Gaussian-like} shifts to the FFNs' activations on \textit{knowledge-related} tokens, can motivate the methods of adding noises that achieve the best results.
\begin{table}[h]
\scriptsize
    \centering
    \begin{tabular}{c|c|c|c|c}
     \toprule
     \multirow{2}{*}{\textbf{Setting}} & \multirow{2}{*}{\textbf{Editor}} & \multicolumn{3}{c}{\textbf{Score $\uparrow$}} \\
     \cmidrule[0.005em](lr){3-3}\cmidrule[0.005em](lr){4-4}\cmidrule[0.005em](lr){5-5}
     & & \textbf{$\left|\sM\right|$=1e$^\text{0}$} & \textbf{$\left|\sM\right|$=1e$^\text{2}$} & \textbf{$\left|\sM\right|$=1e$^\text{4}$} \\
     \midrule
     \multirow{4}{*}{\makecell[c]{GPT2-xl\\zsRE}}
     & \CC MEMIT$_\text{DNE}$ & \CC\textbf{44.80} & \CC\textbf{47.31} & \CC41.57 \\
     & MEMIT$_\text{SNE}$ & 41.88 & 46.09 & 41.89 \\
     & MEMIT$_\text{UN}$ & 39.34 & 43.52 & 29.55 \\
     & MEMIT$_\text{RNP}$ & 37.88 & 45.88 & \textbf{41.93} \\
     \cmidrule[0.005em]{1-5}
     \multirow{4}{*}{\makecell[c]{GPT-J\\Counterfacts}}
     & \CC MEMIT$_\text{DNE}$ & \CC 92.47 & \CC \textbf{92.64} & \CC \textbf{85.87} \\
     & MEMIT$_\text{SNE}$ & 92.38 & 92.12 & 85.84 \\
     & MEMIT$_\text{UN}$ & \textbf{92.60} & 91.06 & 70.94 \\
     & MEMIT$_\text{RNP}$ & 92.15 & 91.97 & 85.85 \\
     \bottomrule
\end{tabular}
\caption{Results of the three ablation studies.}
\label{tab:ab}
\end{table}
Table \ref{tab:ab} reports the results of the Scores of the three ablation studies and we leave the detailed results to Appendix \ref{sec:d_ab}.
DNE can achieve the highest Score (as well as the highest generalization) in the most cases.
DNE will mostly improves generalization but also slightly decrease specificity.
Because the Score calculates the harmonic mean, which is sensitive to smaller values, but the baselines of generalization are much larger than the specificity, DNE's can then be lower than the counterpart methods in some cases.

\section{Conclusions}
In this paper, we study the questions of: how LLMs can recall the same knowledge in the different contexts and how we can edit LLMs' knowledge while maintaining such important properties.
For the first part, we follow the state-of-the-art editing methods and the latest interpretability works to focus on analyzing FFNs' activations.
Though comparing the histogram figures and numerical results, we empirically show that different contexts can only place small shifts that follow considerably narrow Gaussian-like distributions in FFNs' activations on knowledge-related tokens.
And LLMs' FFNs can produce kind of~'dominate' activations when processing knowledge.
Motivated by our findings, we make to answer the second part of the questions by adding noises into FFNs' activations when editing LLMs.
By doing so, we can make LLMs \textbf{see the unseen} contexts where the edited knowledge will be applied and improve the editing generalization.
We run experiments on two open-sources including two standard datasets and three popular LLMs.
The experimental results show the effectiveness of our methods.
We make extra discussions, comparisons with other methods of adding noises, and ablation studies to comprehensively analyze how our findings can motivate the methods of adding noises that best fit with the task of knowledge-editing.

\section{Limitations}
Although we have run comprehensive experiments, there are some limitations.
Because of the incompleteness of current open-sources and the limited computing resources, the experiments do not include editing larger LLMs or editing multiple cases and using Counterfacts with LLaMA-2.
We follow exactly the same settings of selecting all hyper-parameters to make our results reproducible.
Therefore, the results of our methods and the baselines may not reach their best performances.
Although the knowledge application is an important topic in LLMs, we narrow on this topic and do not extend our scope of applying our methods of adding noises into the general fine-tunings of LLMs.

\bibliography{main}

\appendix
\section{FFNs activations in Different Layers}
\label{apd:l_a}
\begin{figure}[ht]
    \centering
    \subfigure
        \centering
        \includegraphics[scale=0.125]{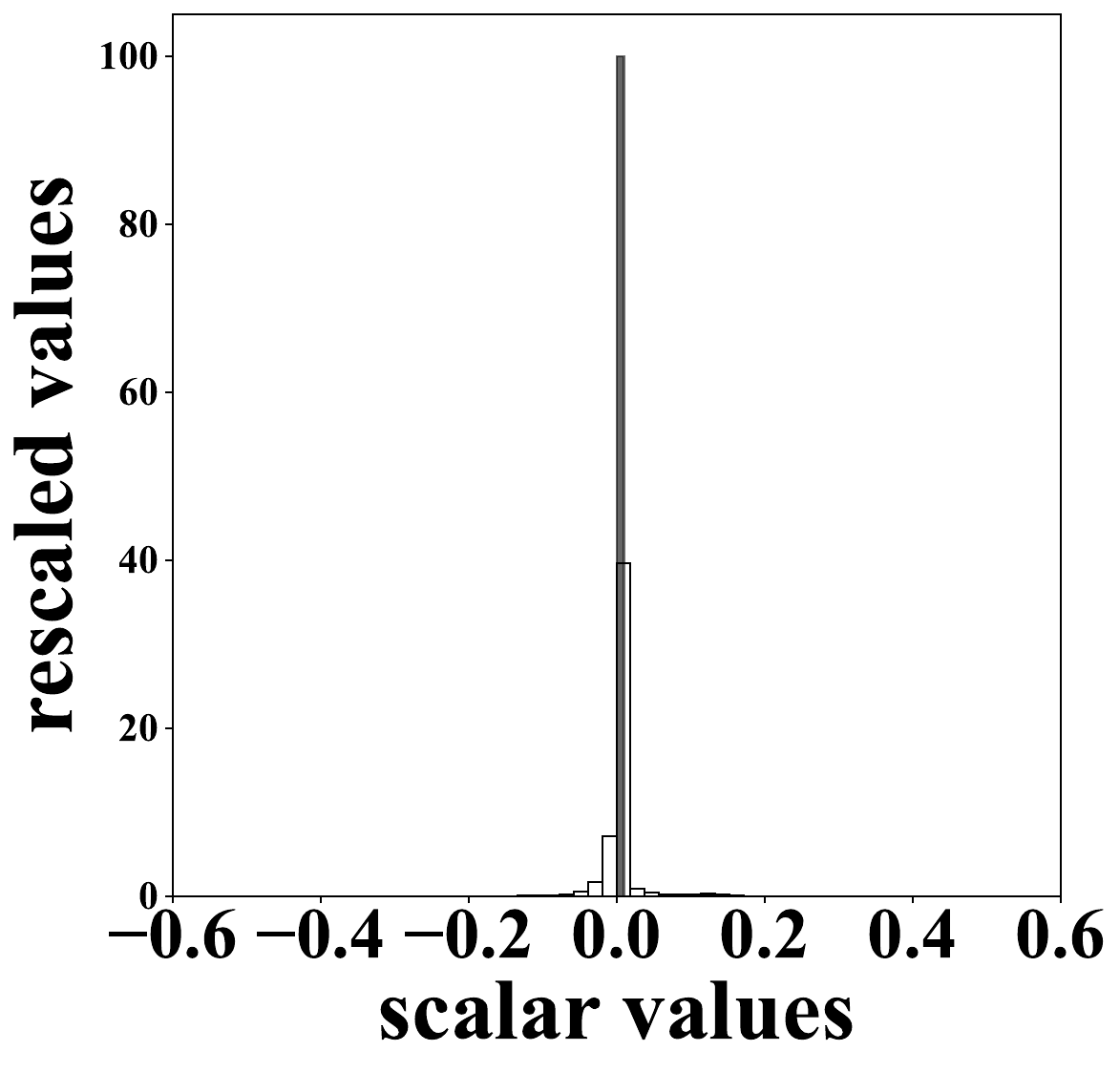}
    \subfigure
        \centering
        \includegraphics[scale=0.125]{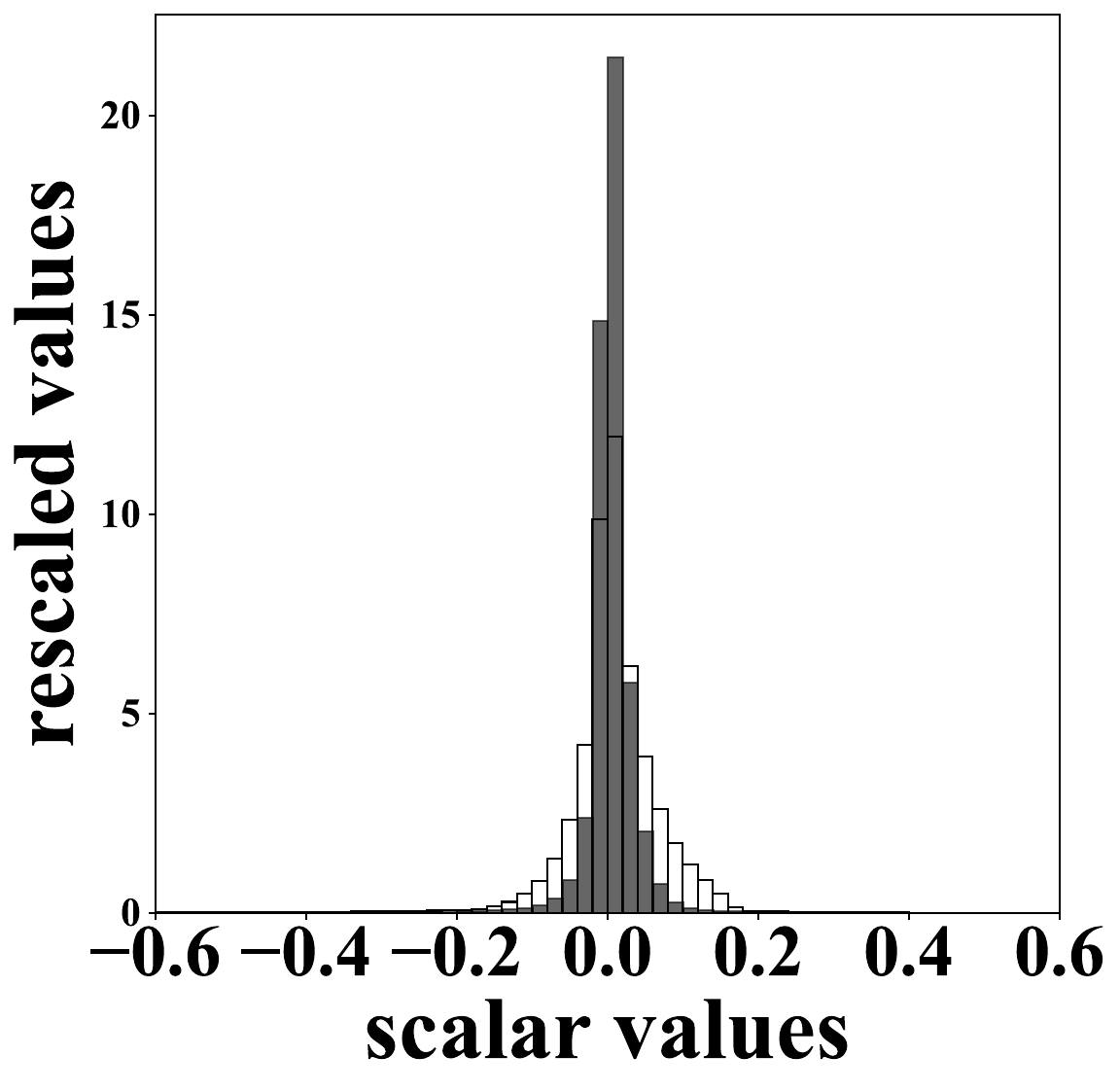}
    \subfigure
        \centering
        \includegraphics[scale=0.125]{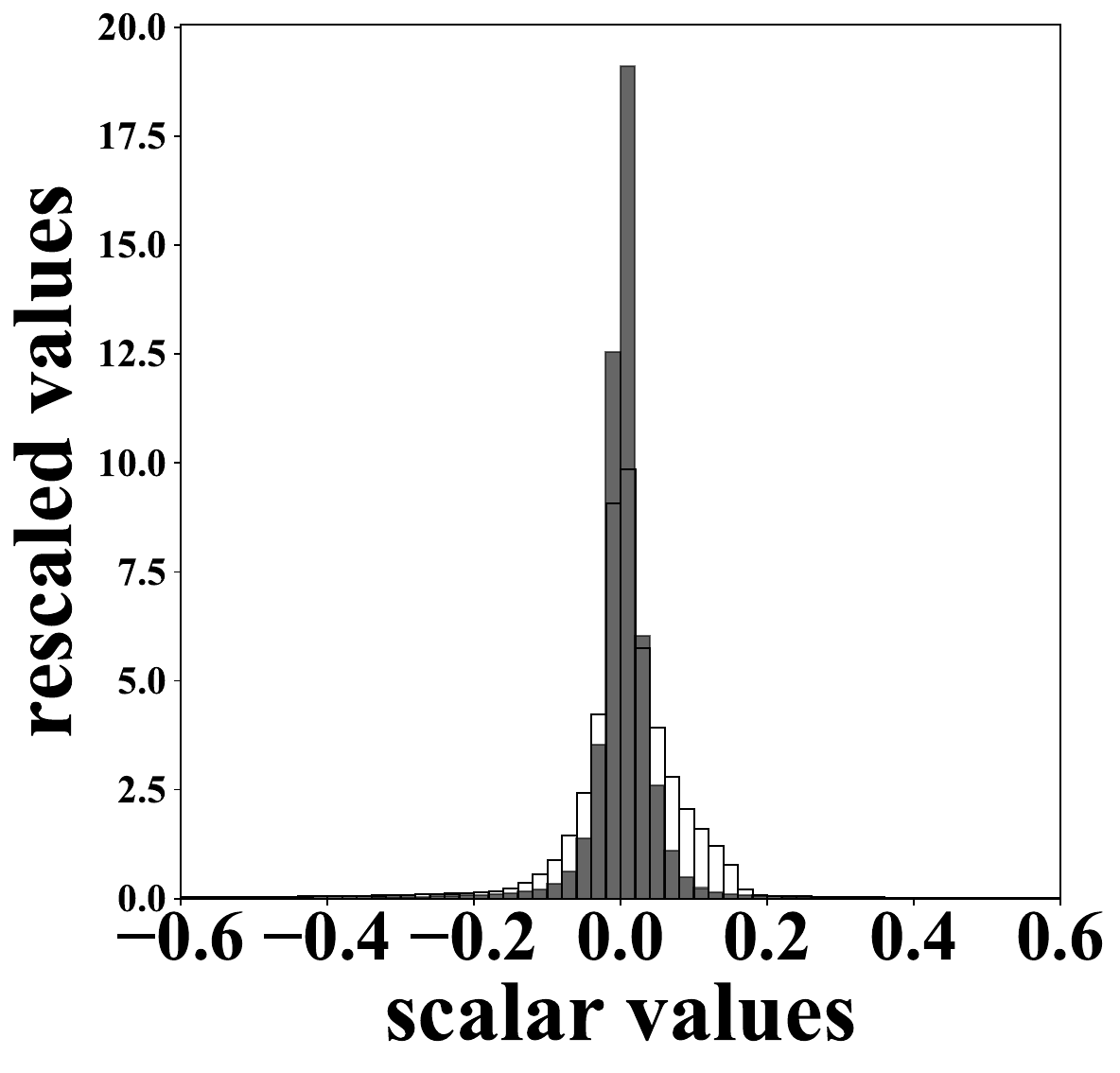}\\
    \subfigure
        \centering
        \includegraphics[scale=0.125]{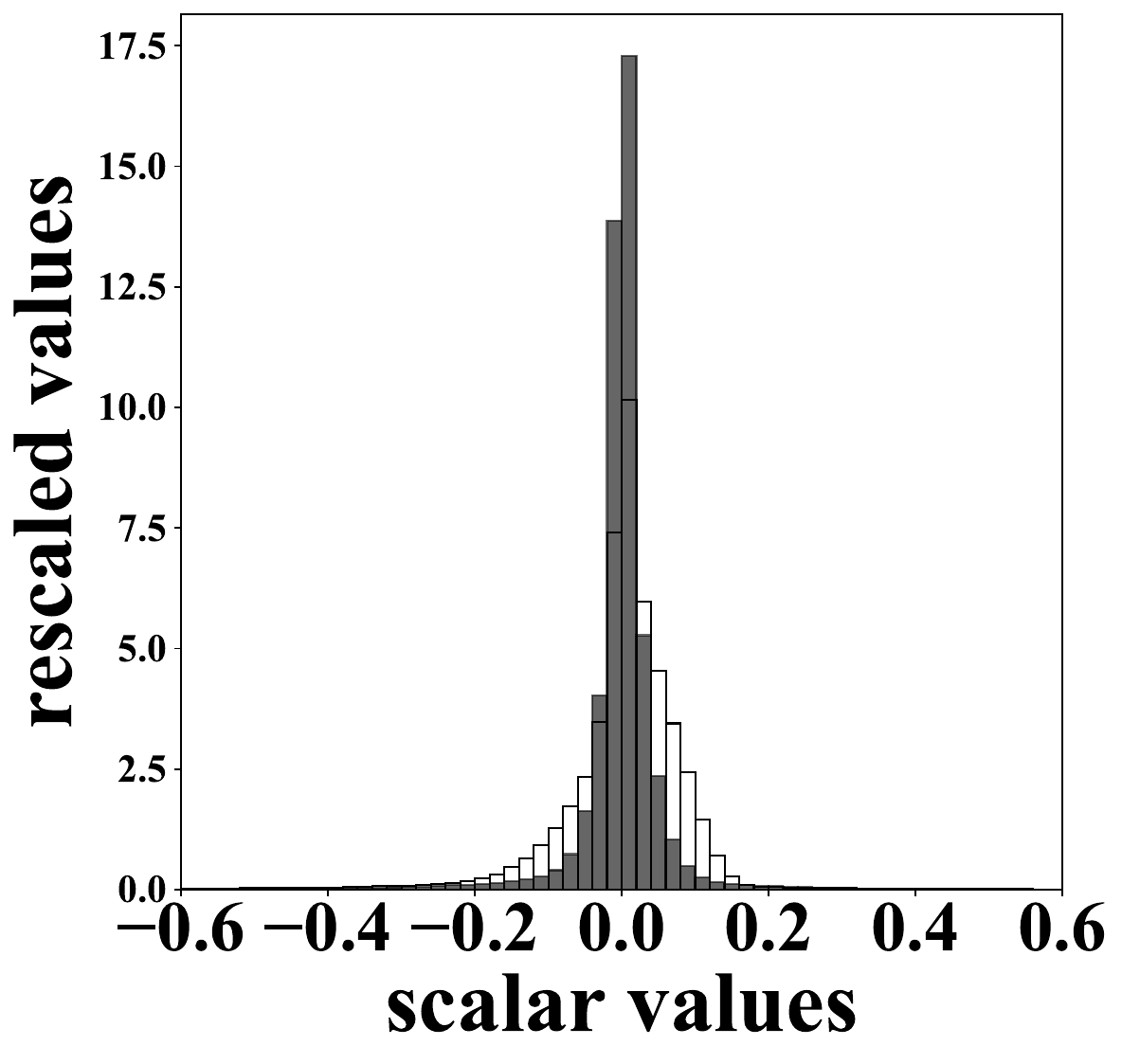}
    \subfigure
        \centering
        \includegraphics[scale=0.125]{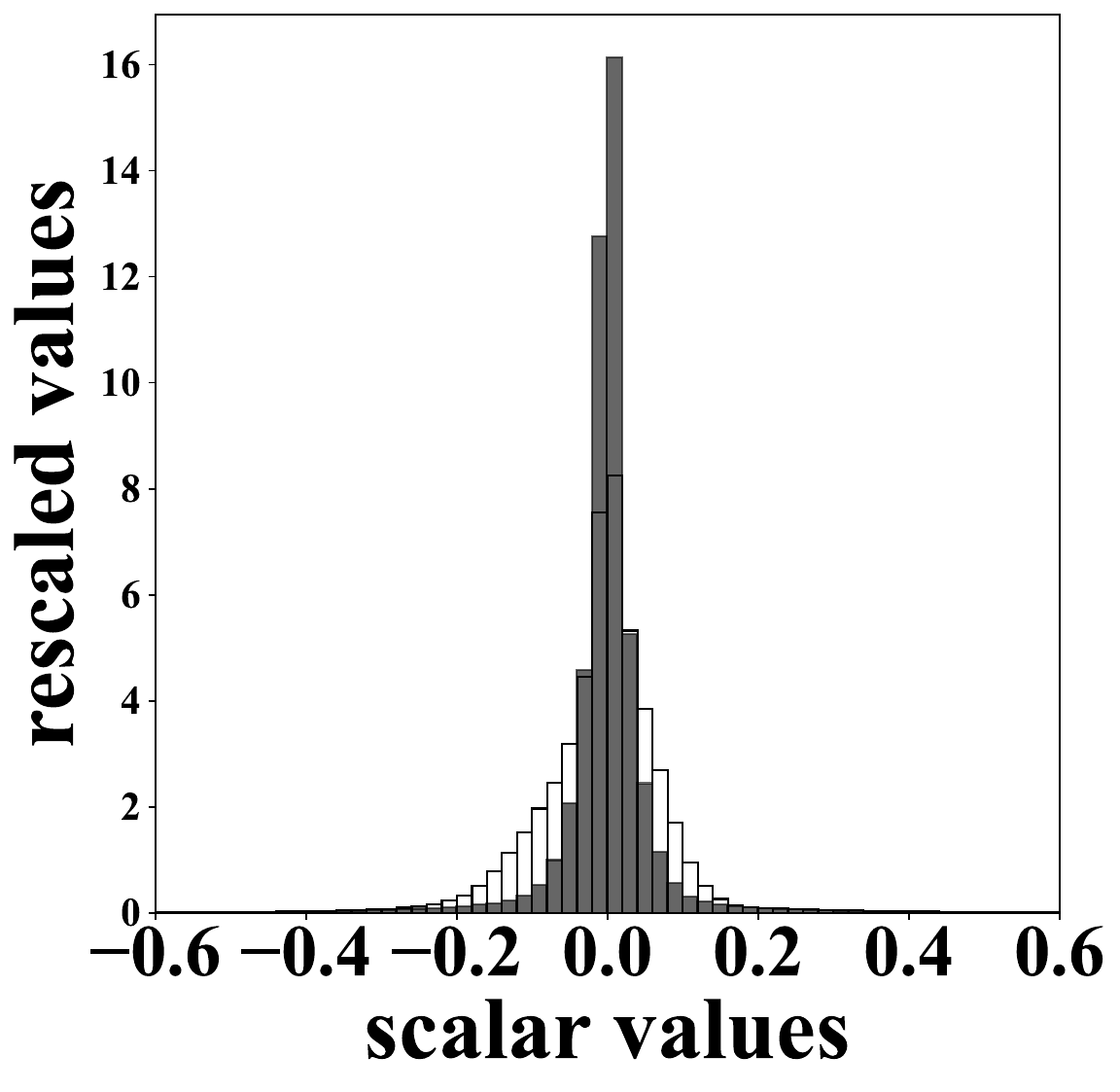}
    \subfigure
        \centering
        \includegraphics[scale=0.125]{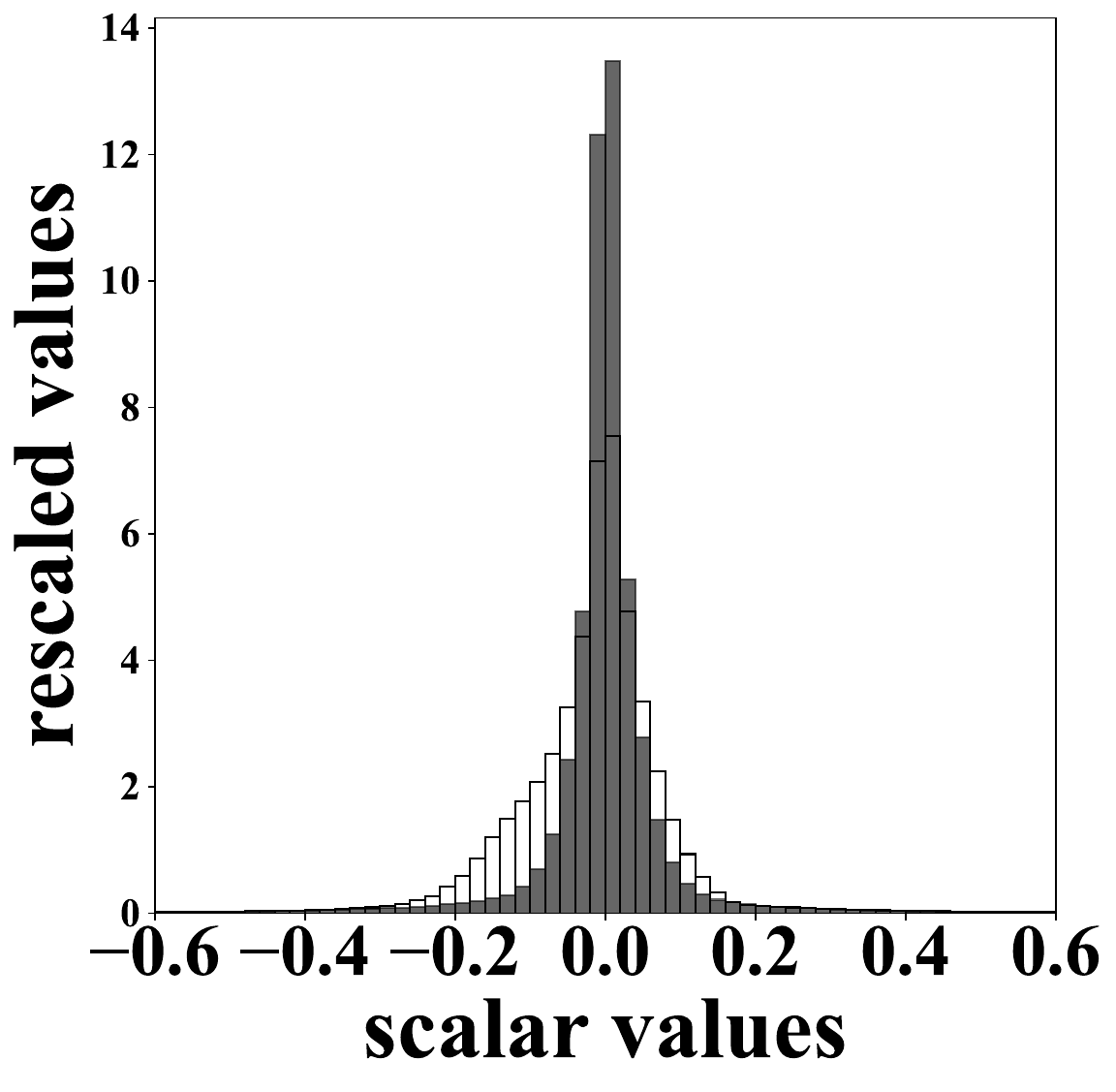}\\
    \subfigure
        \centering
        \includegraphics[scale=0.125]{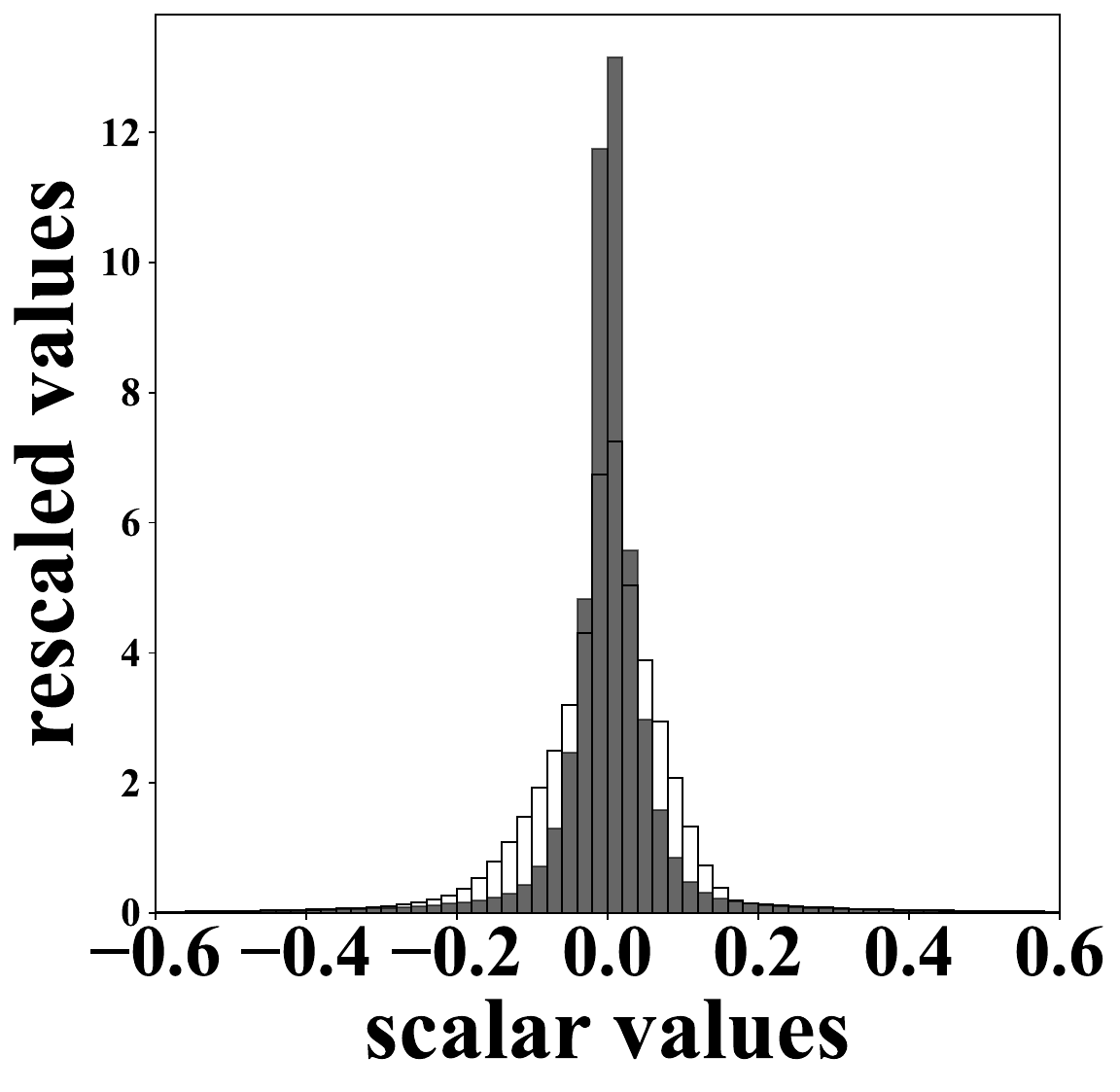}
    \subfigure
        \centering
        \includegraphics[scale=0.125]{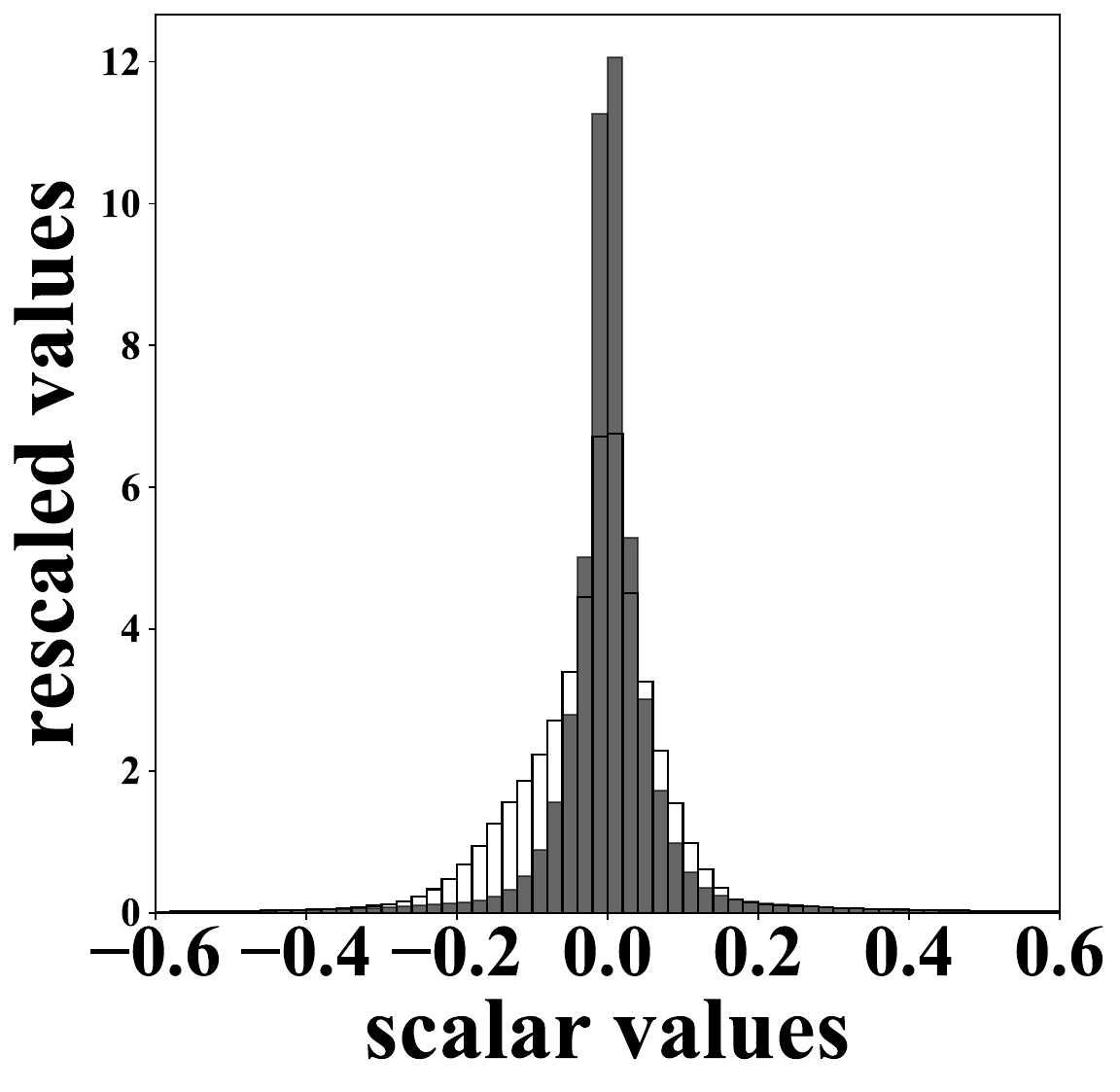}
    \subfigure
        \centering
        \includegraphics[scale=0.125]{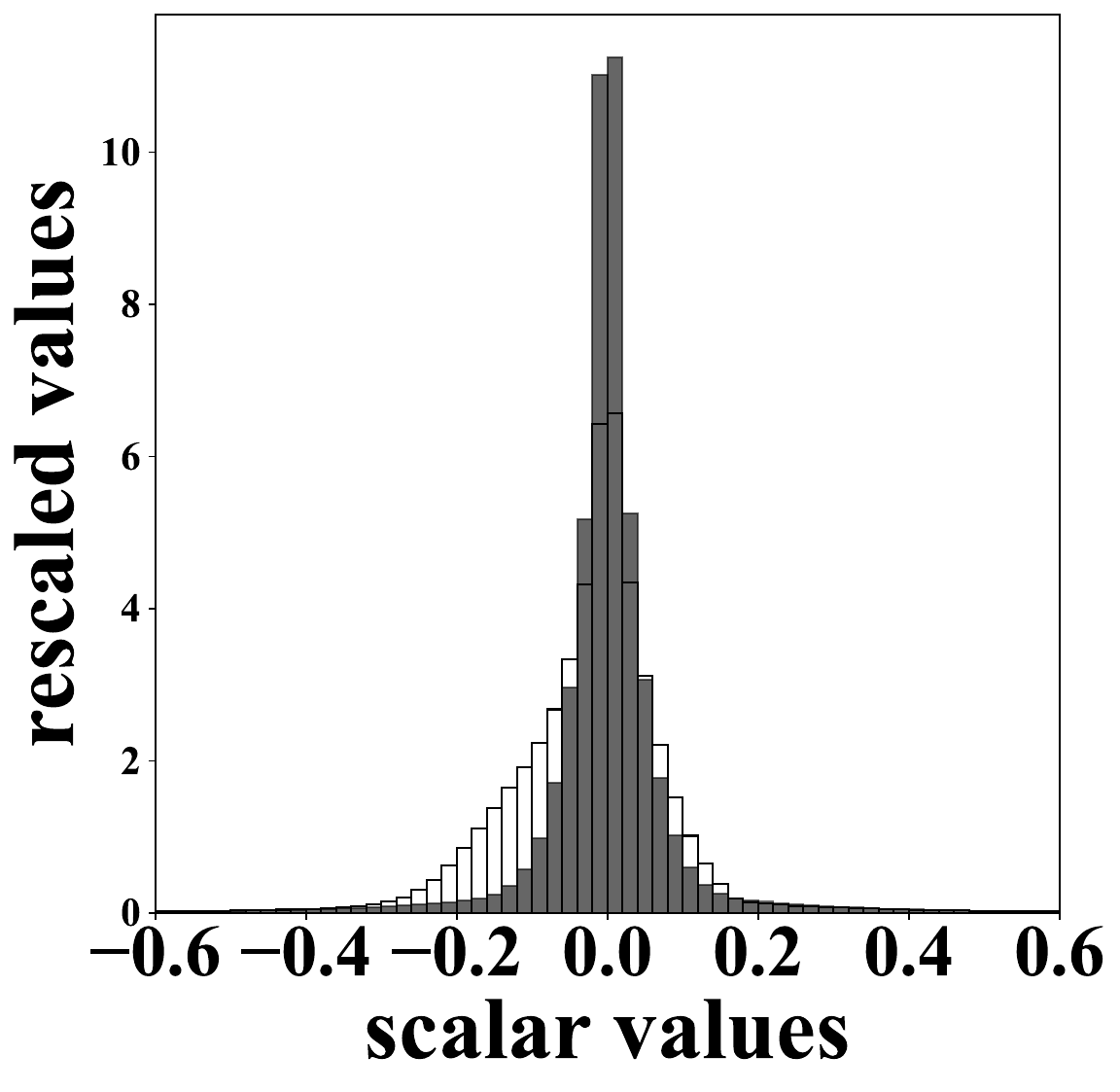}
    \caption{$\sD_s, \sD_c$ of GPT-J's layers from $\text{1}^{\text{st}}$ to $\text{8}^{\text{th}}$.}
    \label{fig:all_gptj}
\end{figure}

\begin{figure}[ht]
    \centering
    \subfigure
        \centering
        \includegraphics[scale=0.125]{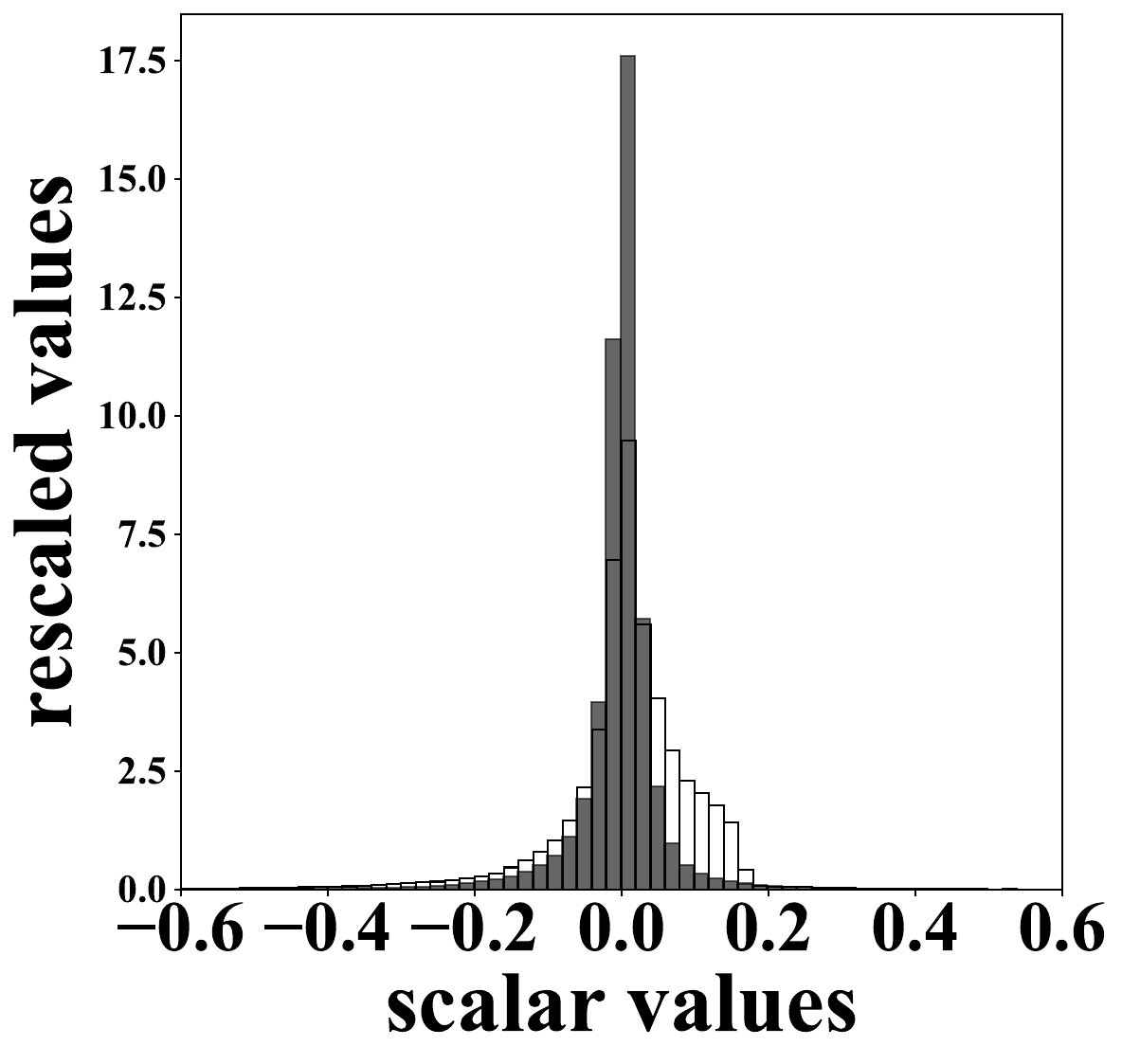}
    \subfigure
        \centering
        \includegraphics[scale=0.125]{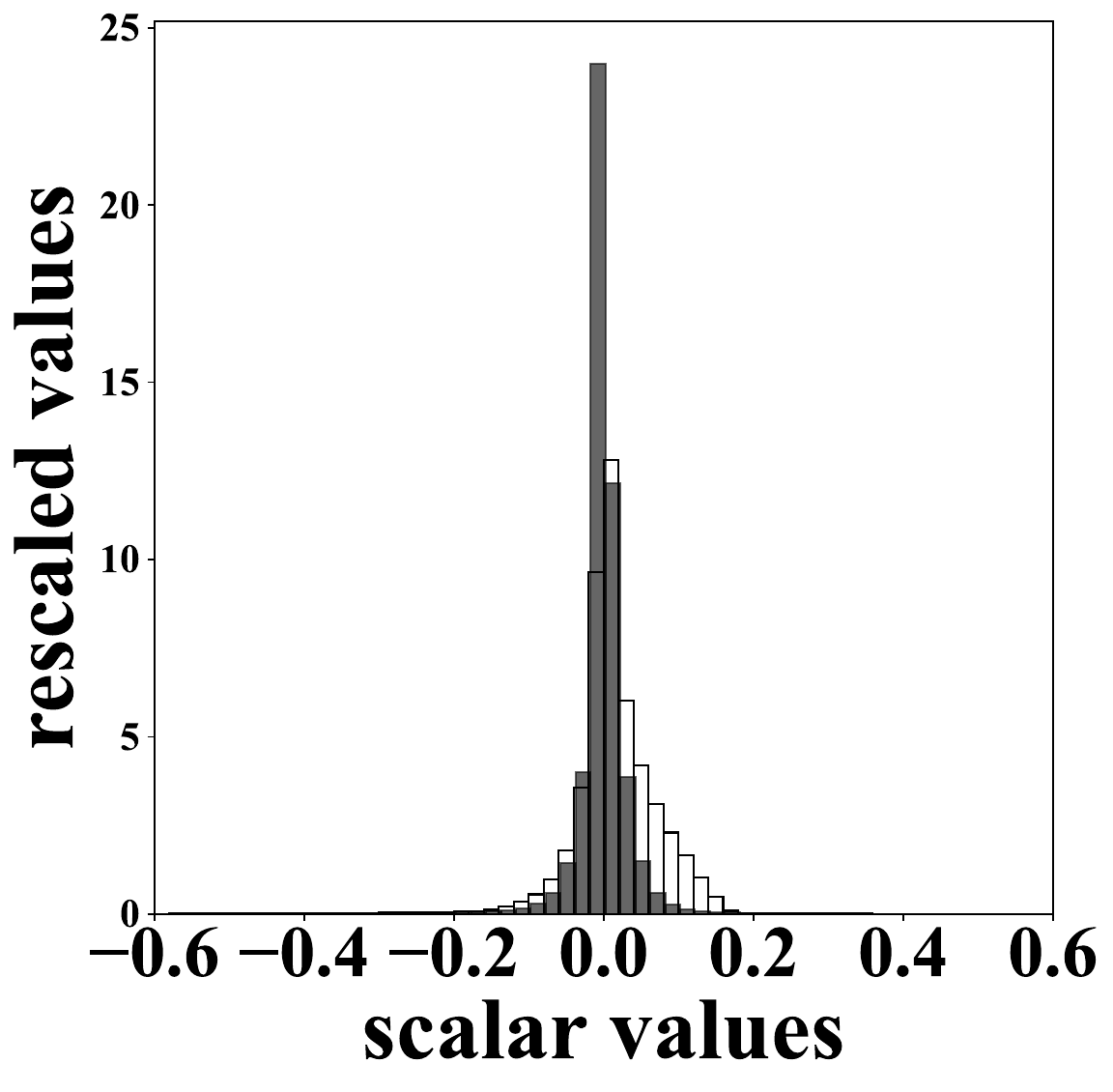}
    \subfigure
        \centering
        \includegraphics[scale=0.125]{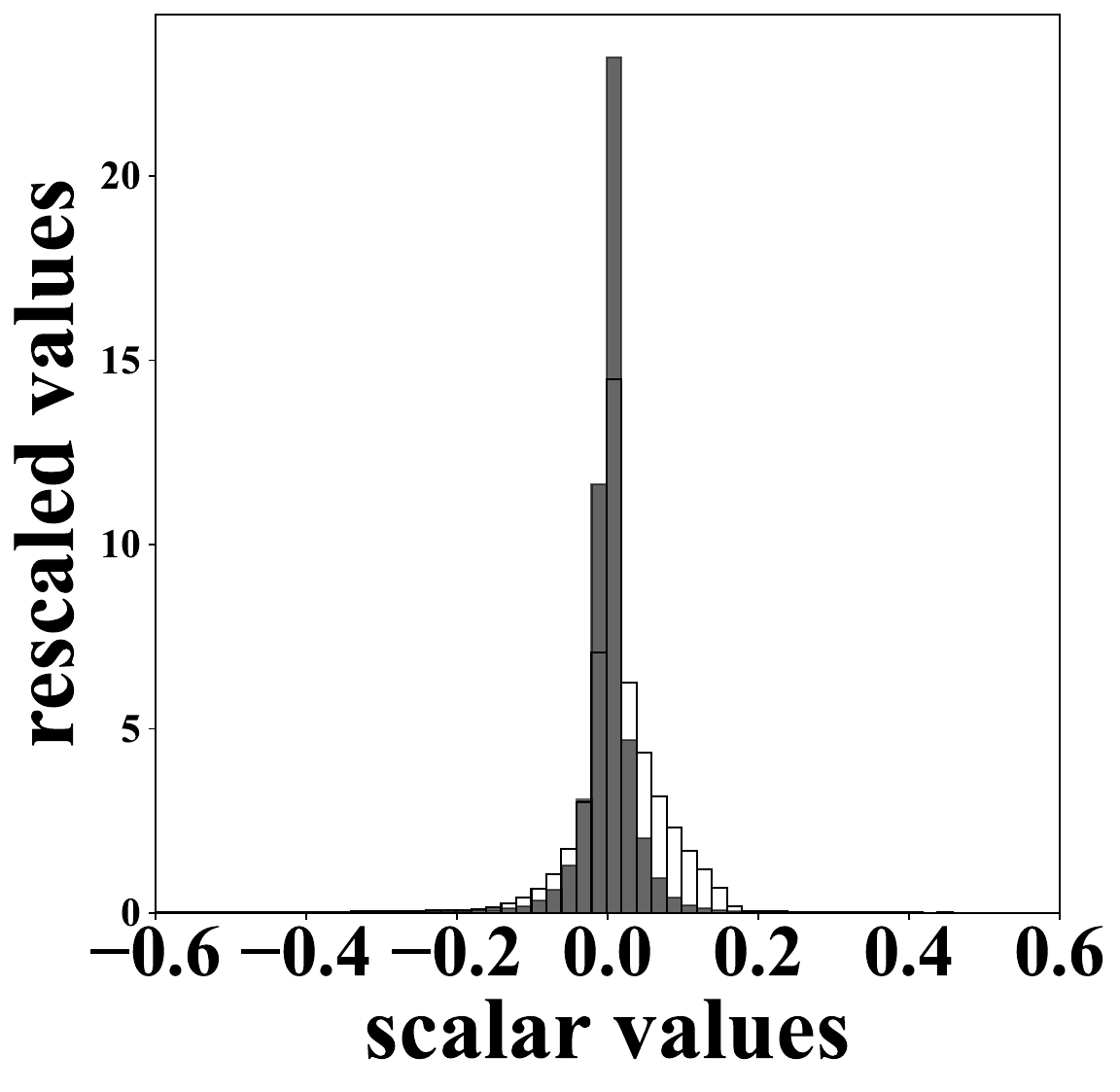}\\
    \centering
    \subfigure
        \centering
        \includegraphics[scale=0.125]{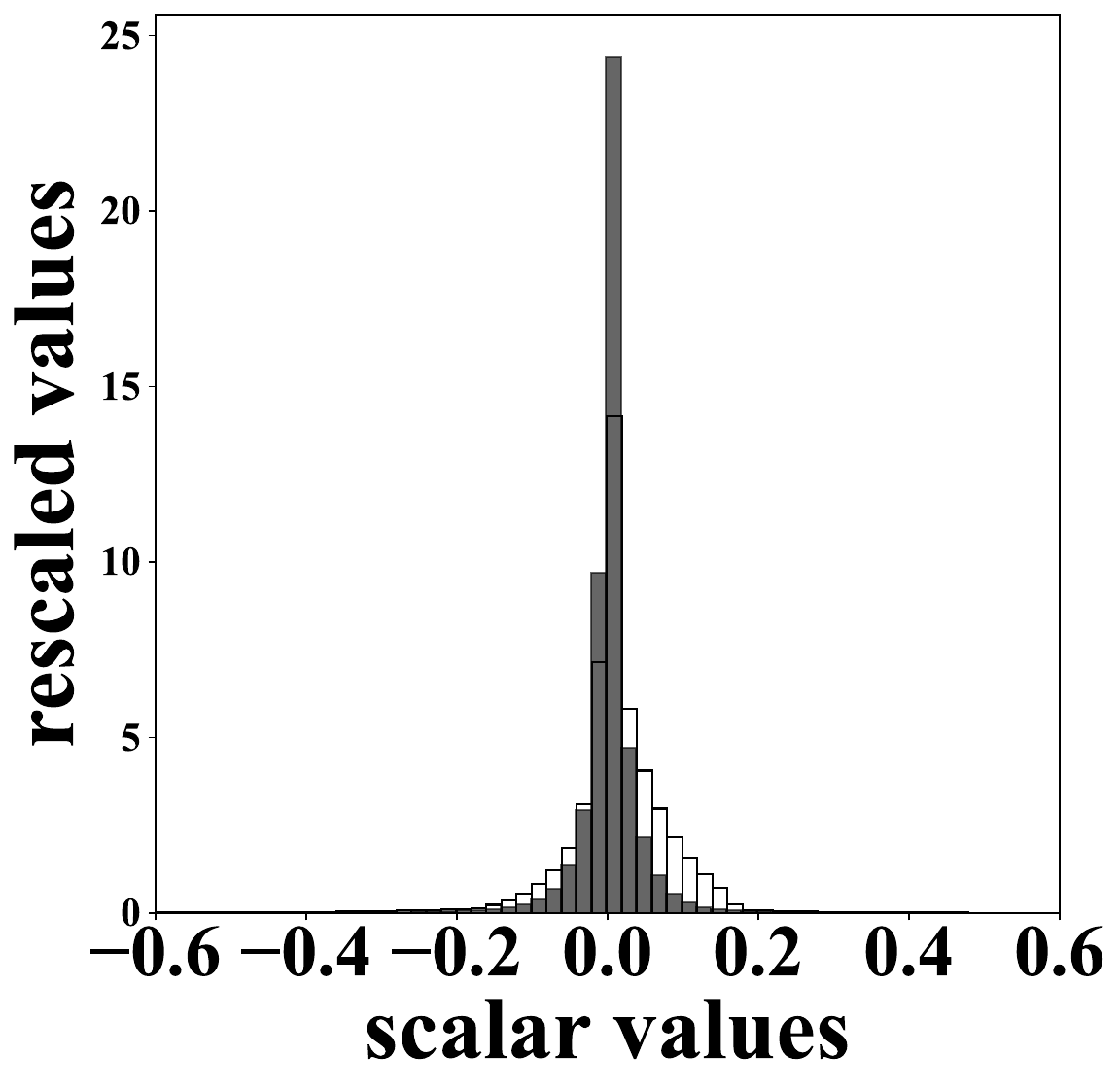}
    \subfigure
        \centering
        \includegraphics[scale=0.125]{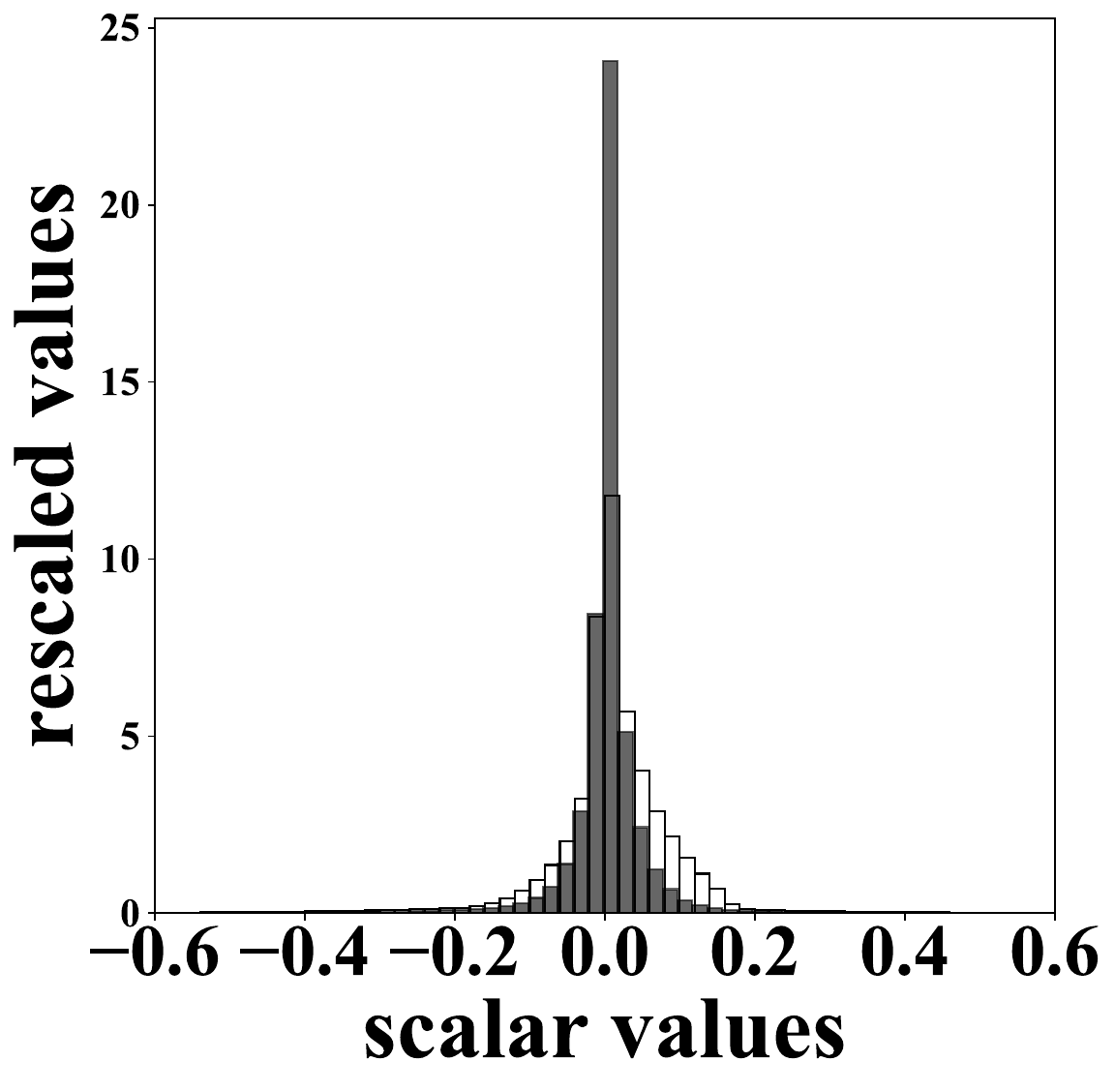}
    \subfigure
        \centering
        \includegraphics[scale=0.125]{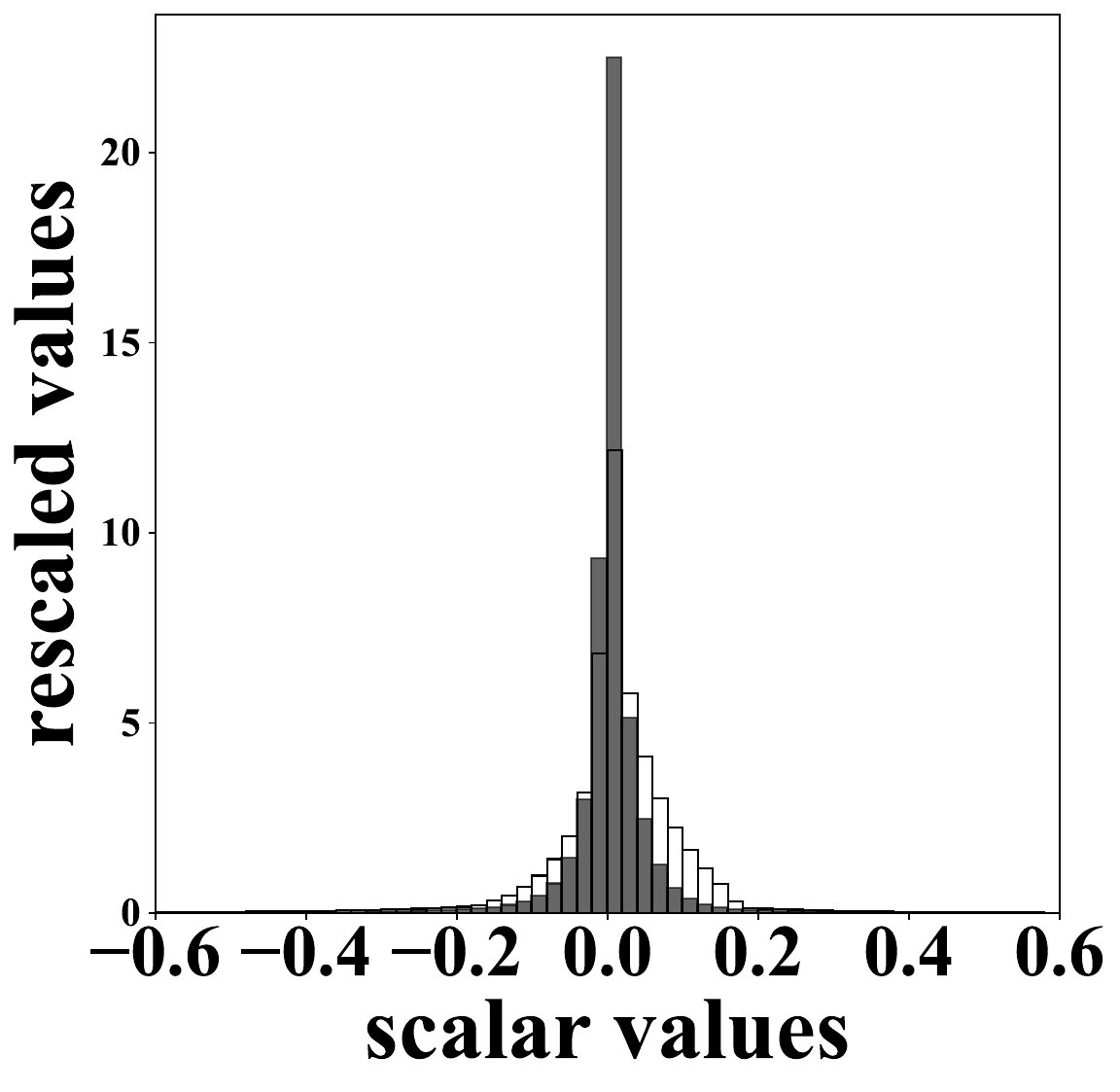}\\
    \centering
    \subfigure
        \centering
        \includegraphics[scale=0.125]{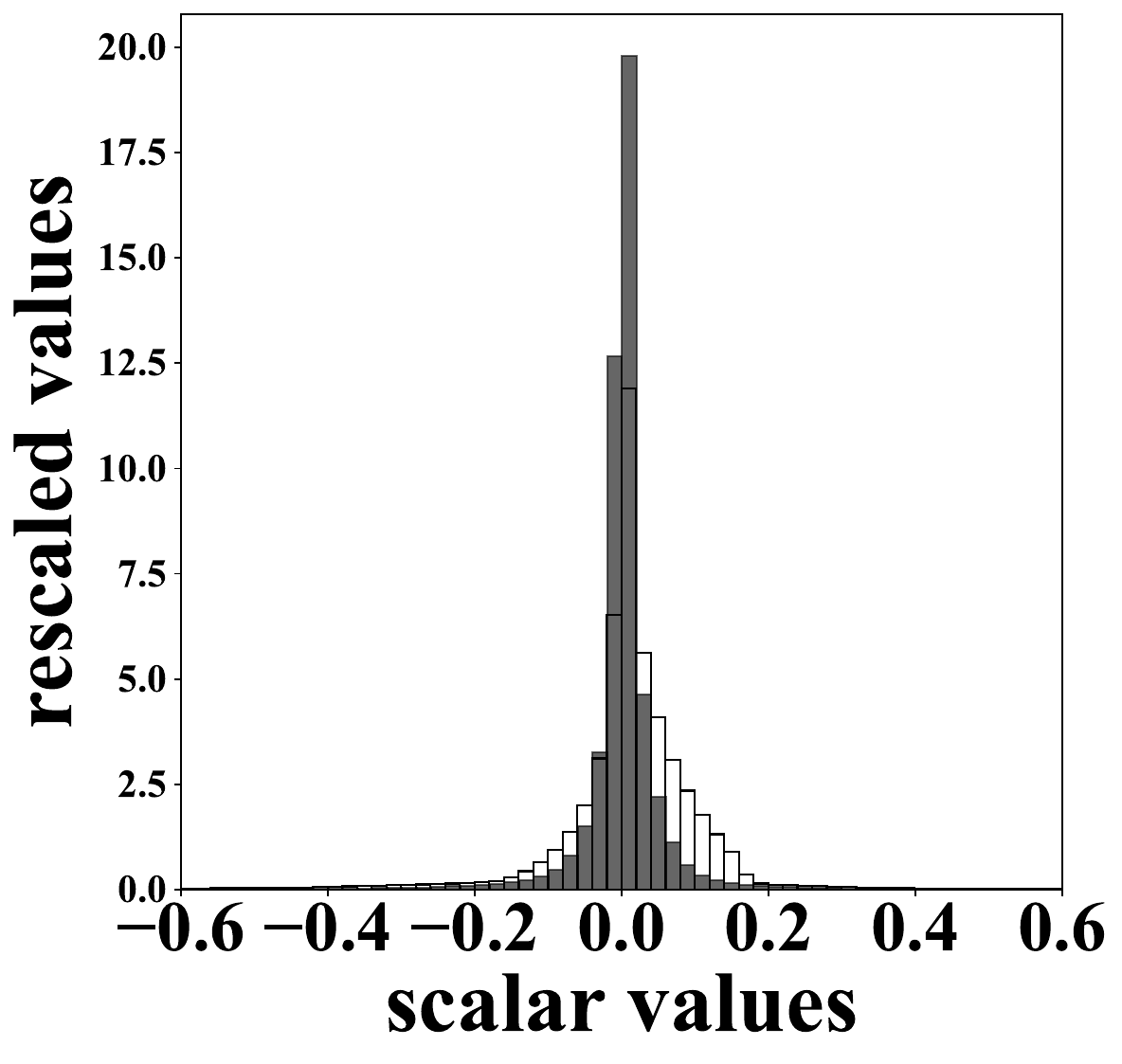}
    \subfigure
        \centering
        \includegraphics[scale=0.125]{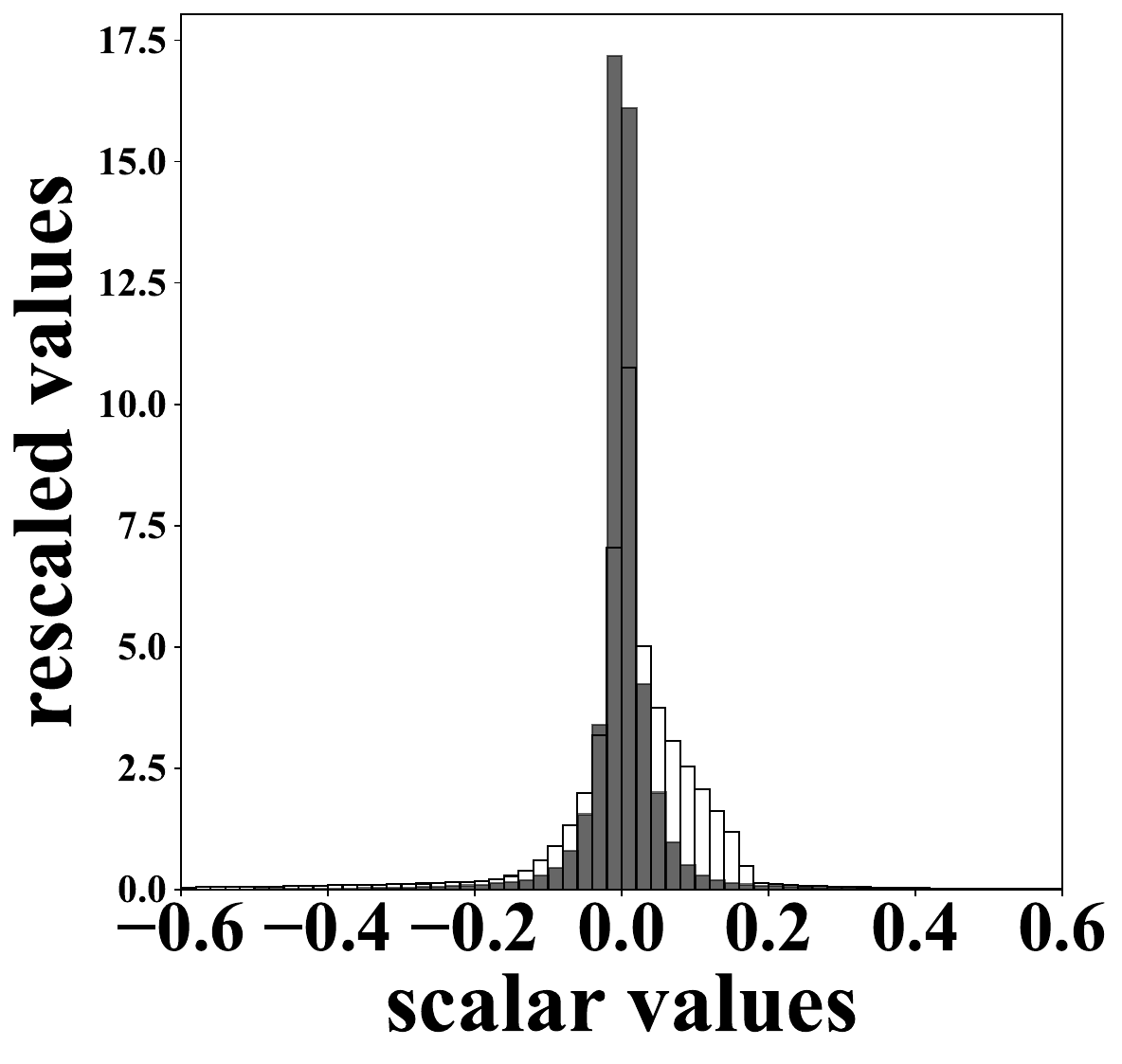}
    \subfigure
        \centering
        \includegraphics[scale=0.125]{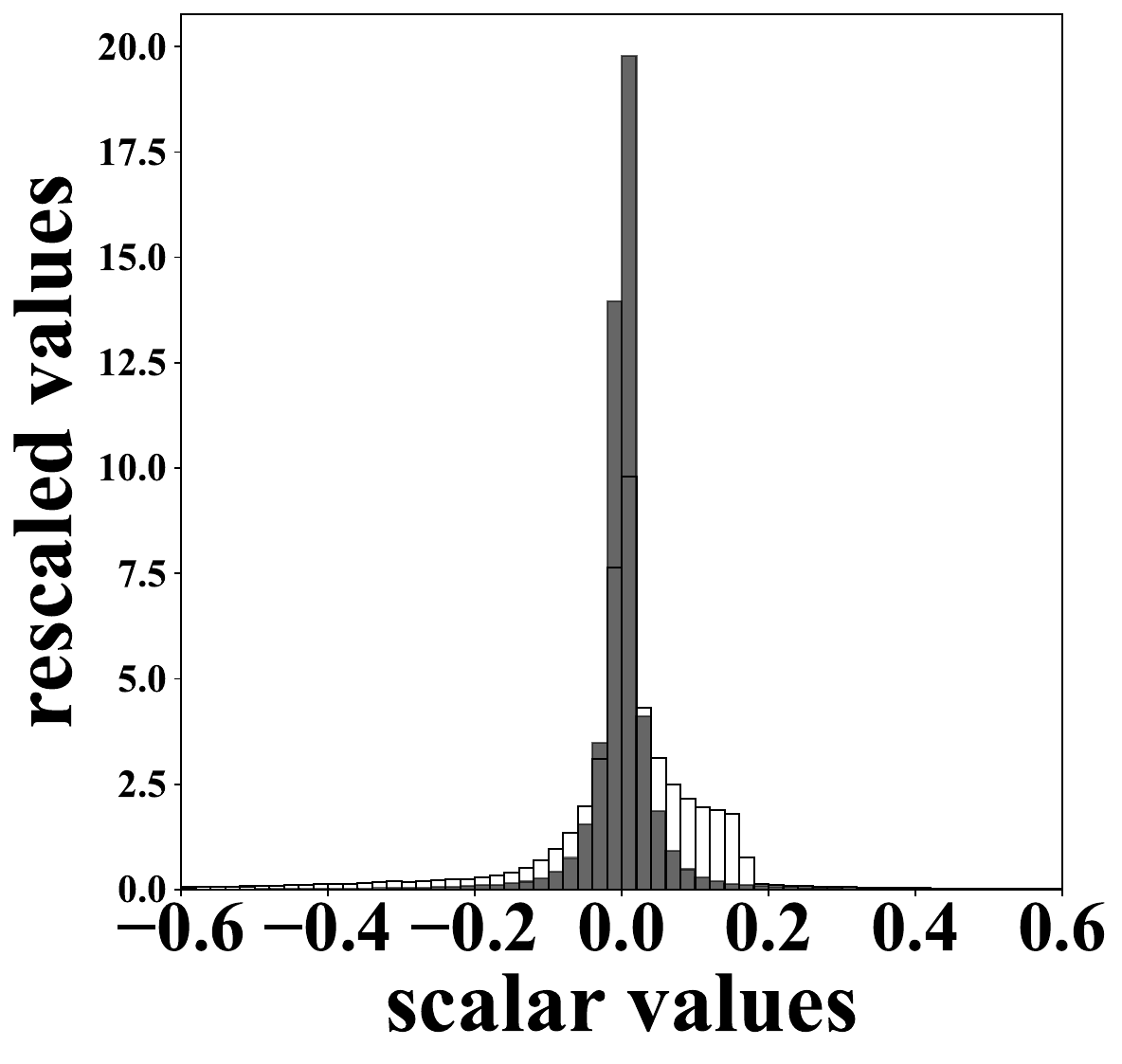}\\
    \centering
    \subfigure
        \centering
        \includegraphics[scale=0.125]{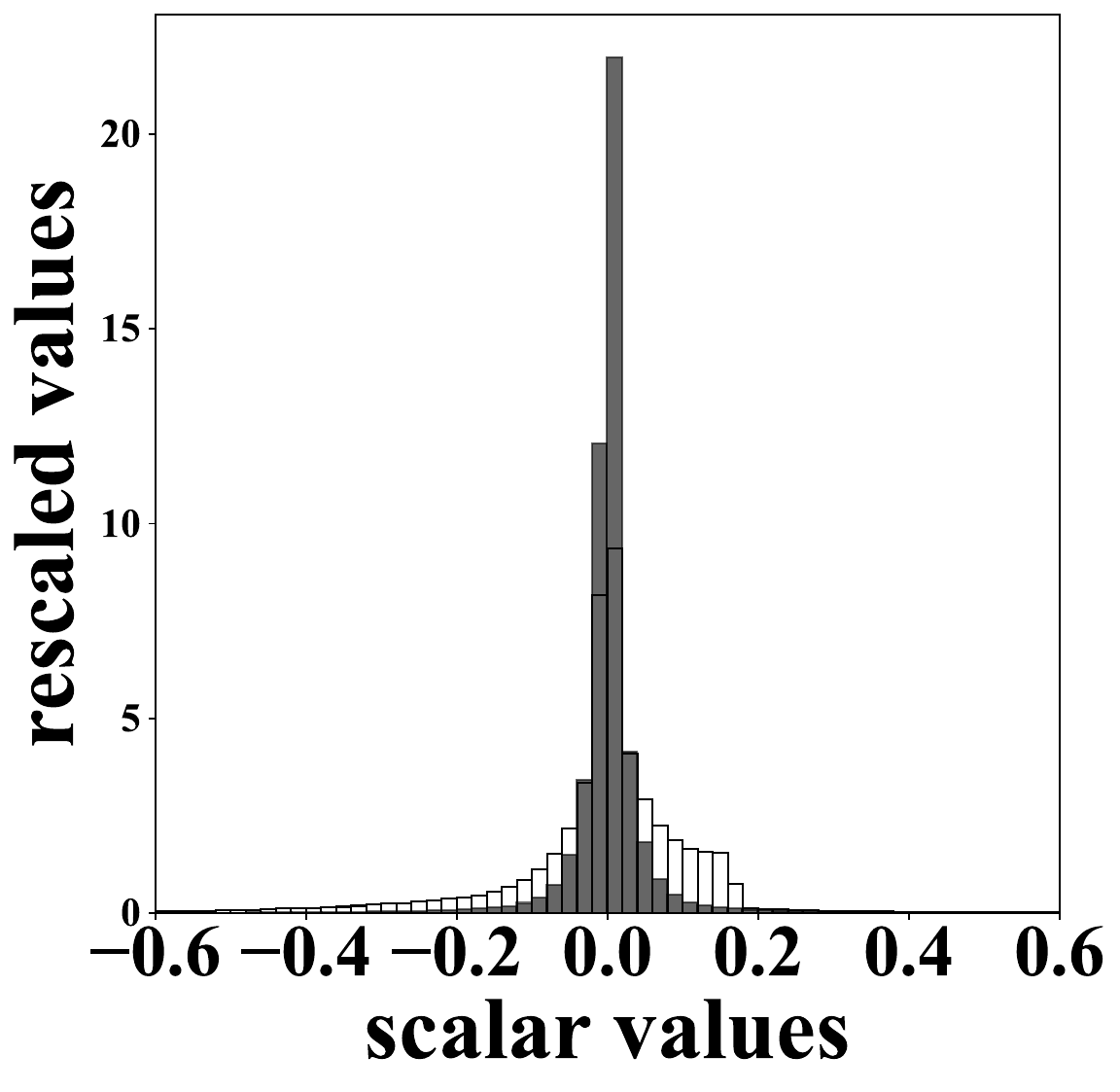}
    \subfigure
        \centering
        \includegraphics[scale=0.125]{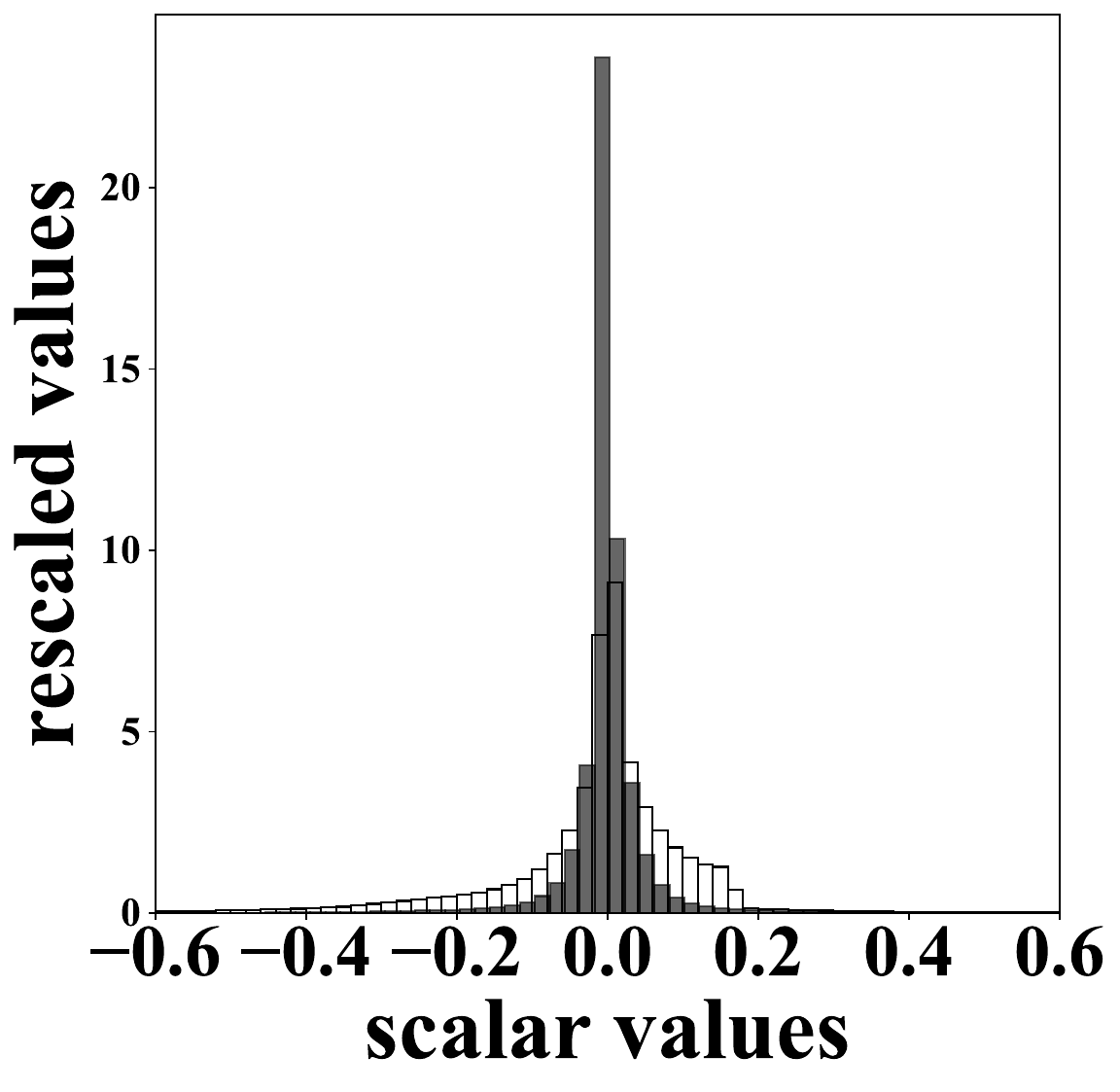}
    \subfigure
        \centering
        \includegraphics[scale=0.125]{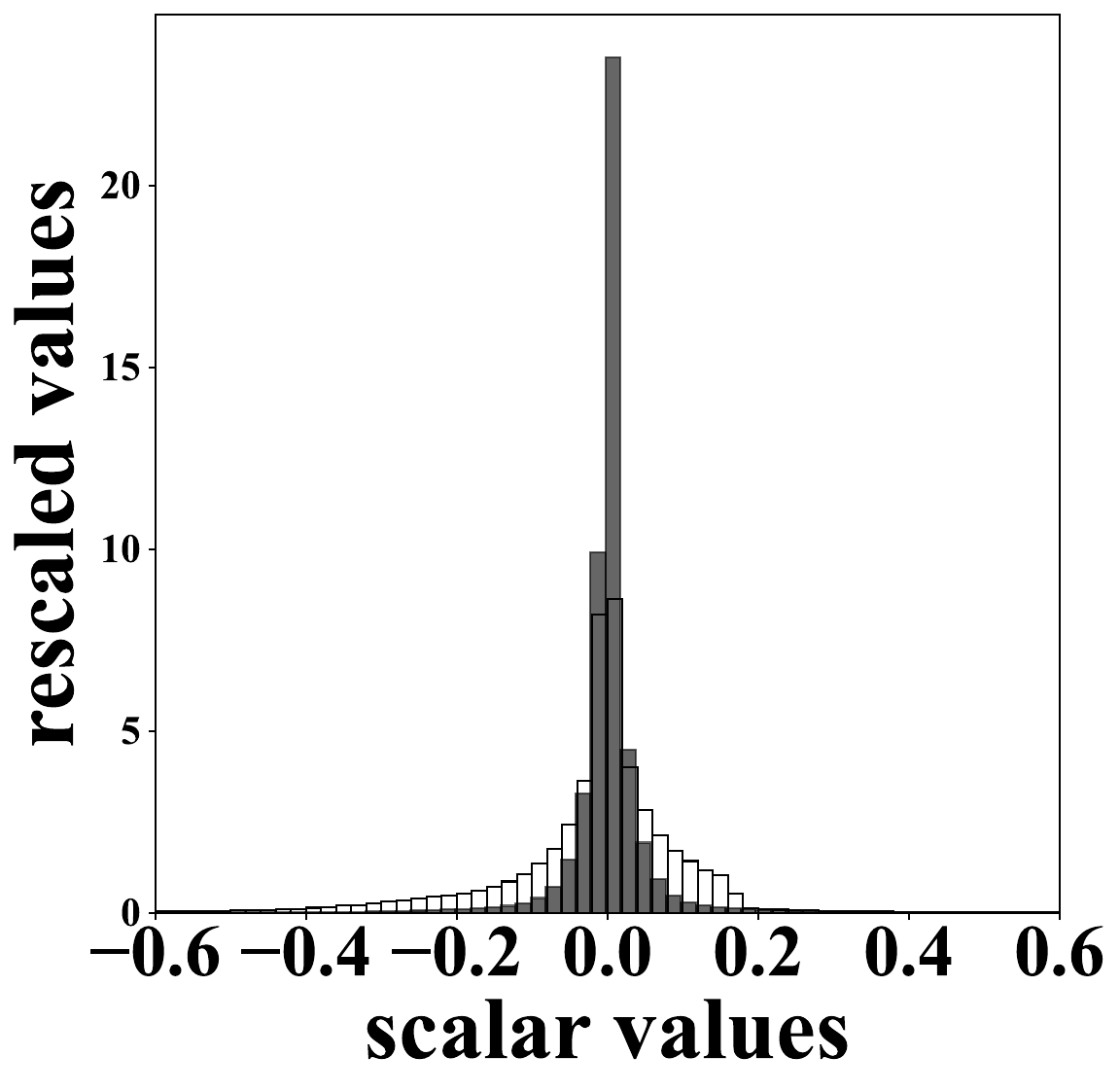}\\
    \centering
    \subfigure
        \centering
        \includegraphics[scale=0.125]{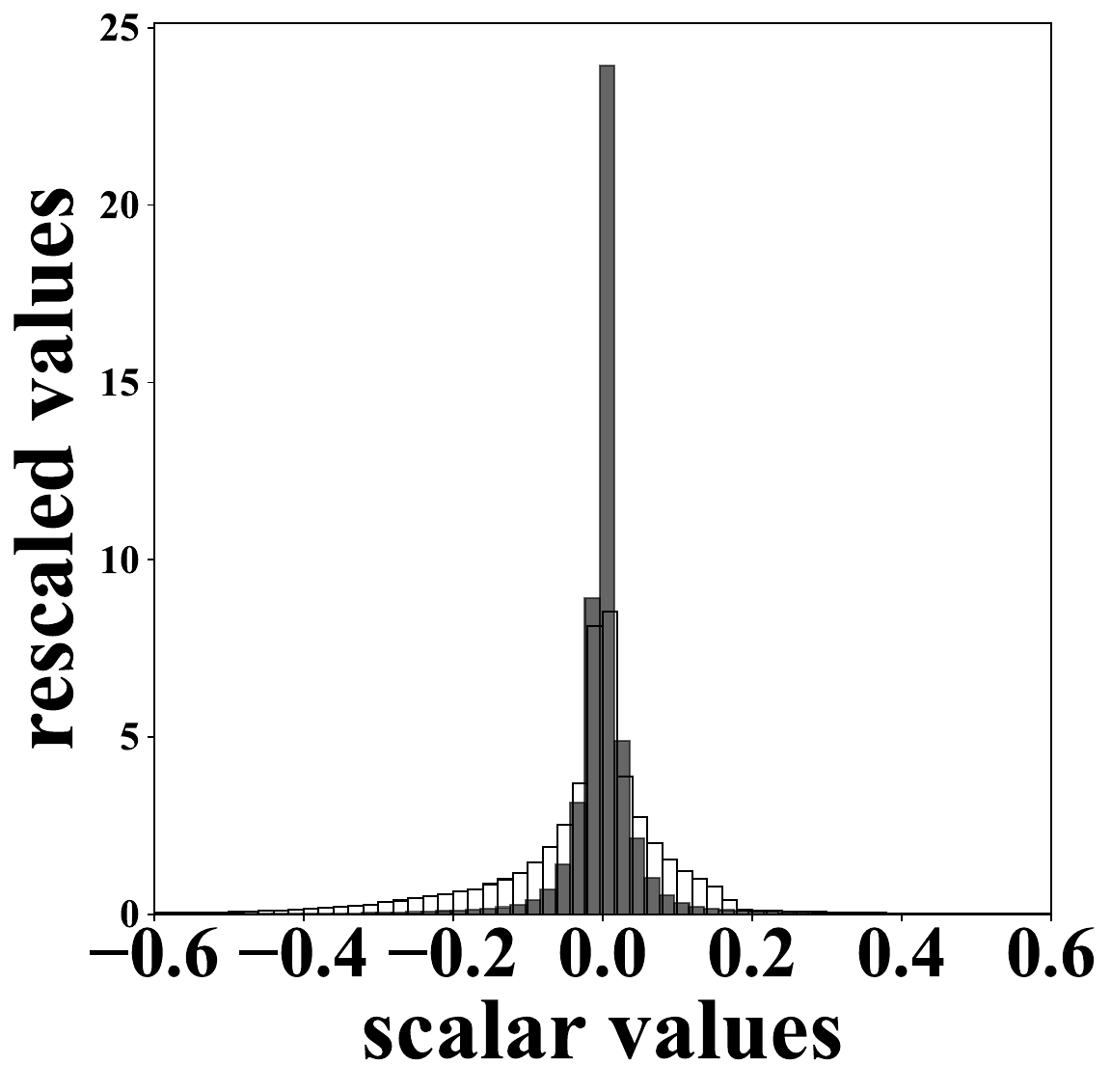}
    \subfigure
        \centering
        \includegraphics[scale=0.125]{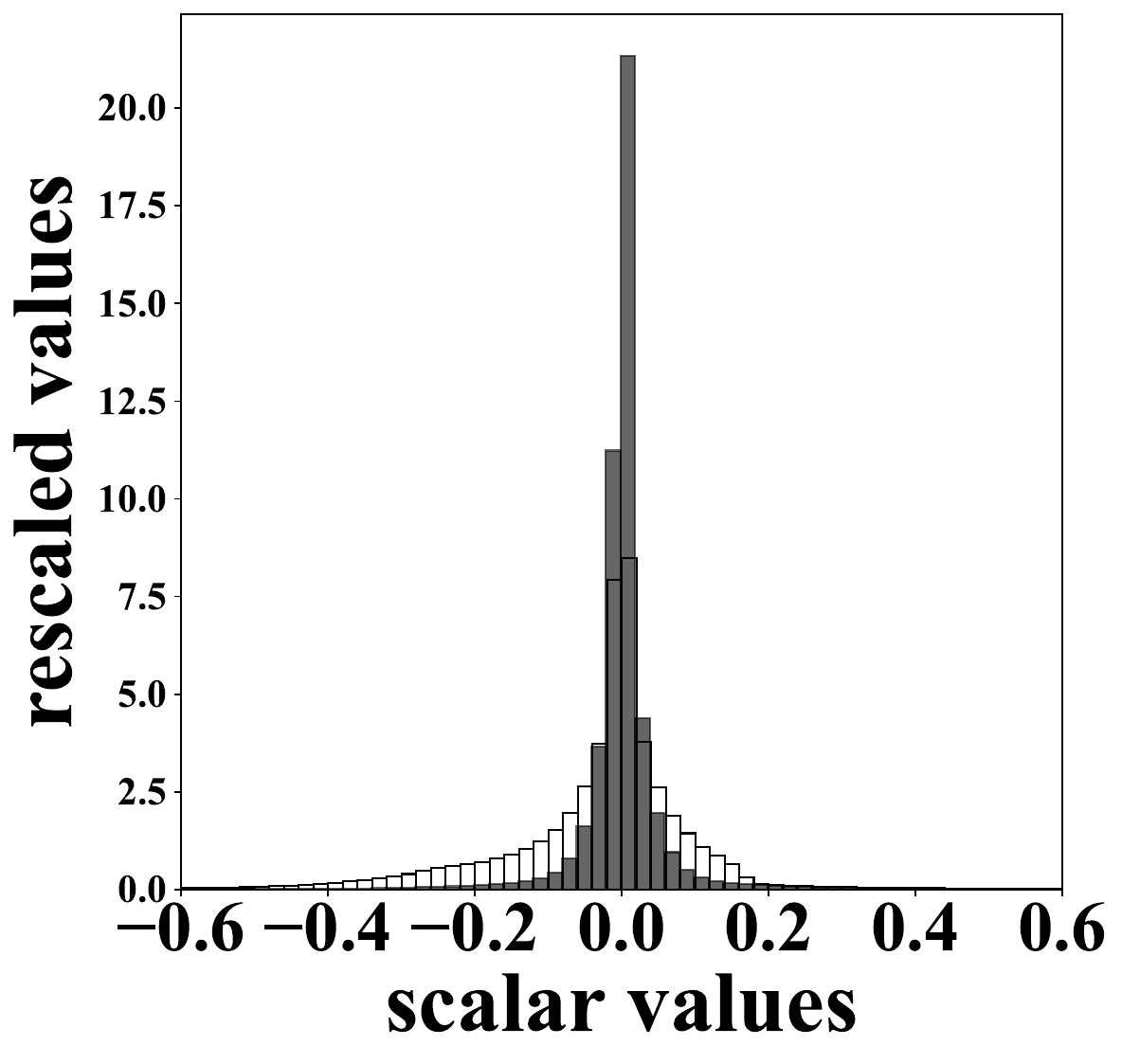}
    \subfigure
        \centering
        \includegraphics[scale=0.125]{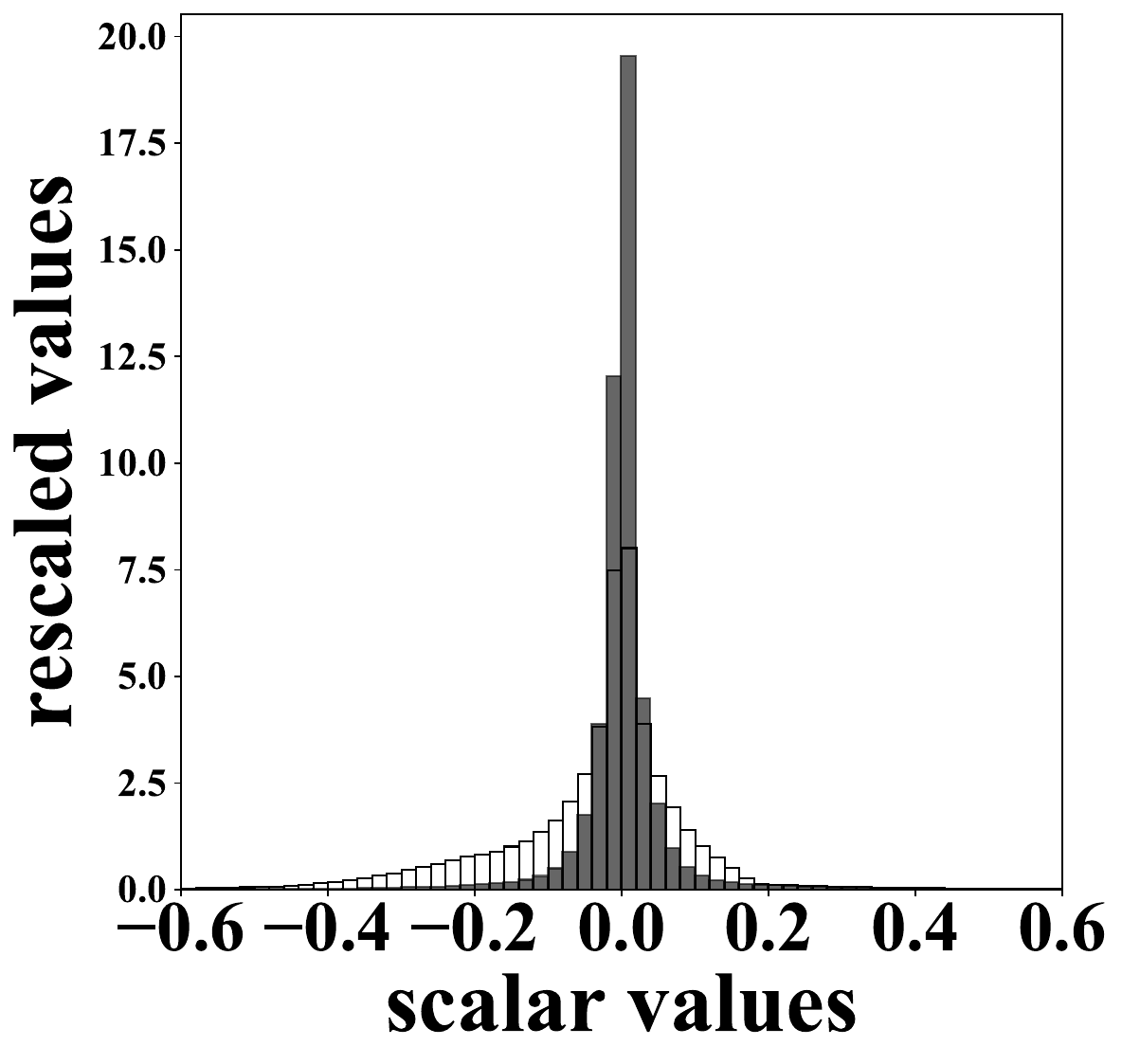}\\
    \centering
    \subfigure
        \centering
        \includegraphics[scale=0.125]{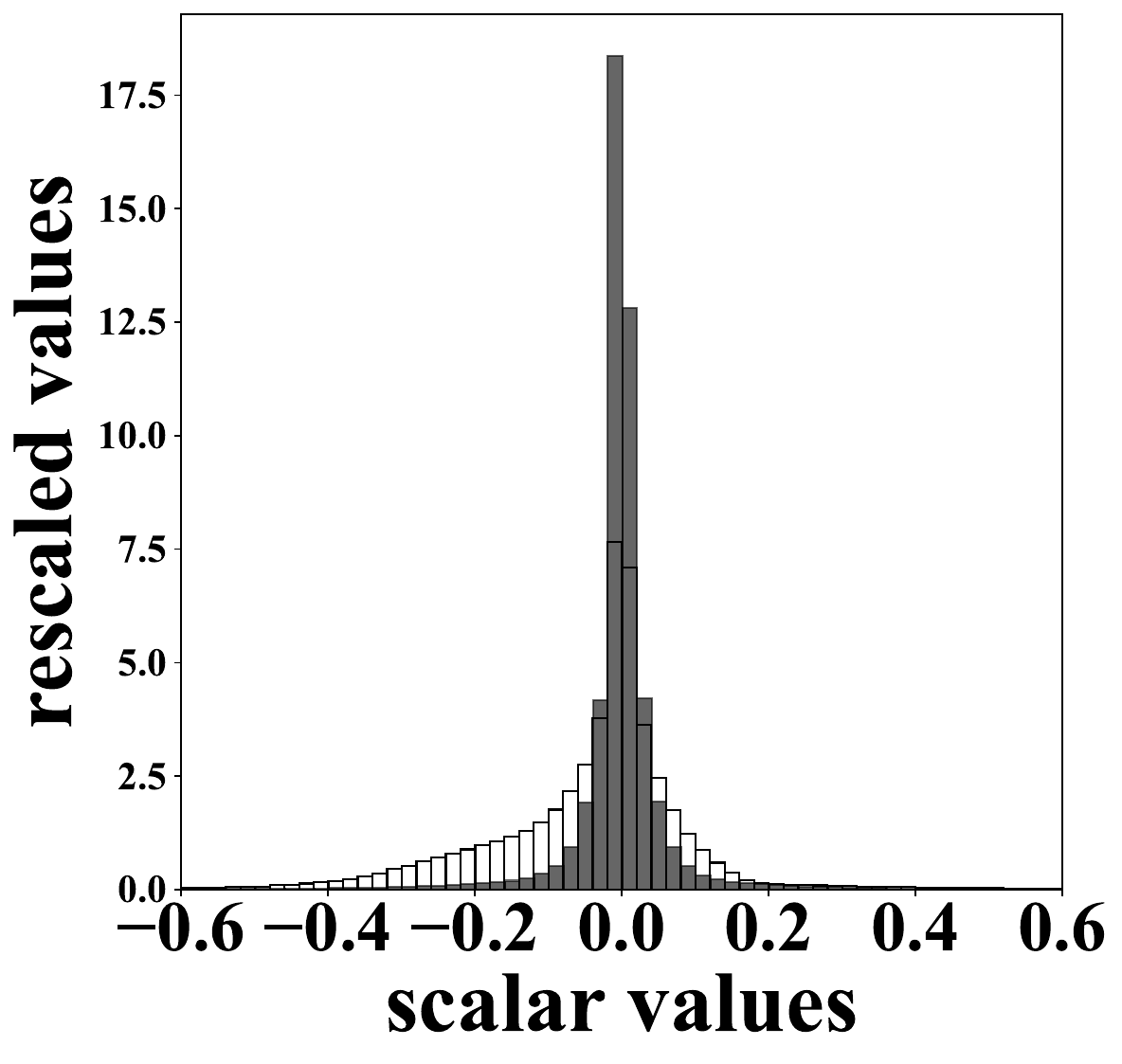}
    \subfigure
        \centering
        \includegraphics[scale=0.125]{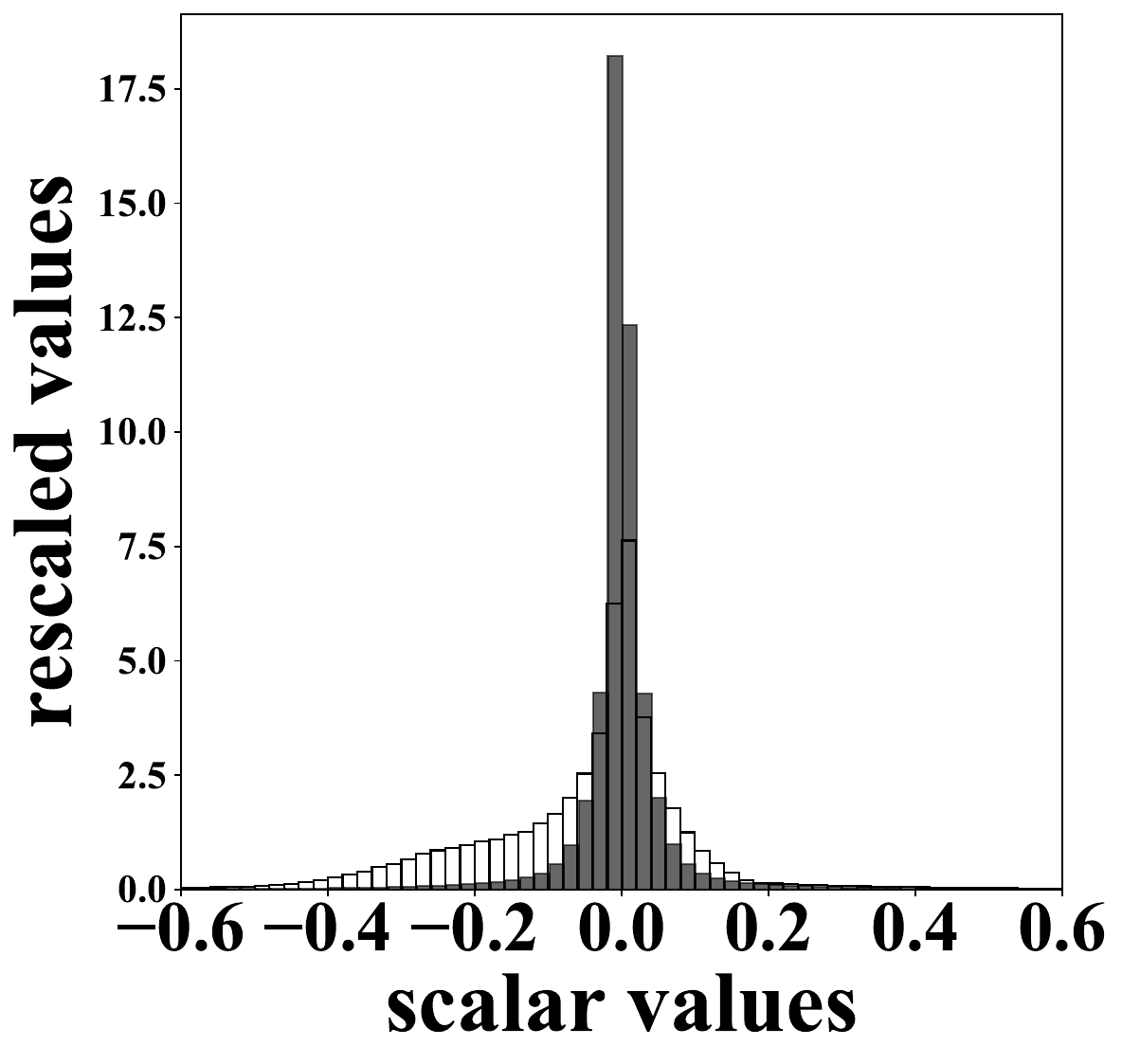}
    \subfigure
        \centering
        \includegraphics[scale=0.125]{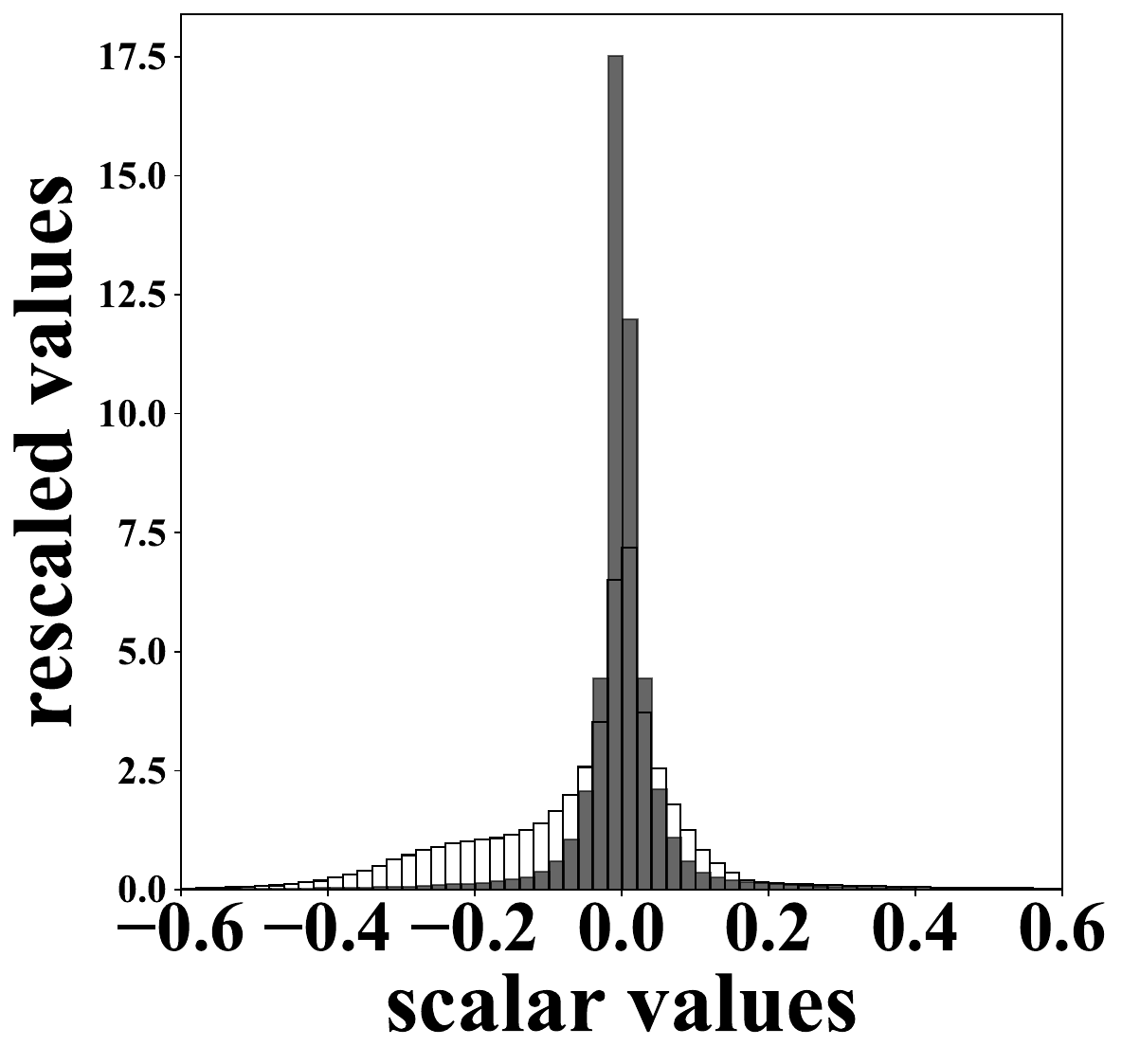}
    \caption{$\sD_s, \sD_c$ of GPT2-xl's layers from $\text{1}^{\text{st}}$ to $\text{18}^{\text{th}}$.}
    \label{fig:all_gpt2}
\end{figure}

Figures \ref{fig:all_gpt2} and \ref{fig:all_gptj} display the histograms for all the layers of the two LLMs we select to analyze, where the black rectangles plot the experimental sets and the white ones plot the control sets and the layer index goes up horizontally and then vertically.
We can see that, the black rectangles (activations of the knowledge-related tokens) remain to be concentrated abound the zeros and descend symmetrically and evenly to the both sides while the white rectangles (activations of normal-strings tokens), in all layers, show much more significant skewness.

\section{Additional Experimental Results}
\label{sec:aer}
\begin{table*}[ht]
\small
    \begin{tabular}{c|c|c|c|c|c|c|c|c}
    \toprule
    \multirow{2}{*}{\textbf{$\left|\sM\right|$}} & \multirow{2}{*}{\textbf{Editor}} 
    & \textbf{Score} & \textbf{Efficacy} & \multicolumn{2}{c|}{\textbf{Generalization}} & \textbf{Specificity} & \textbf{Fluency} & \textbf{Consistency} \\
    \cmidrule[0.005em](lr){3-3}\cmidrule[0.005em](lr){4-4}\cmidrule[0.005em](lr){5-6}\cmidrule[0.005em](lr){7-7}\cmidrule[0.005em](lr){8-8}\cmidrule[0.005em](lr){9-9}
    & & \makecell[r]{S $\uparrow$} & \makecell[r]{ES $\uparrow$} & \makecell[r]{PS $\uparrow$} & \makecell[r]{PA $\uparrow$} & \makecell[r]{NS $\uparrow$} & \makecell[r]{GE $\uparrow$} & \makecell[r]{RS $\uparrow$} \\
    \midrule
    \multirow{5}{*}{1e$^{\text{0}}$}
    & ROME & 89.80  & \textbf{99.94 (0.0)} & 97.08 (0.2) & 74.04 (0.5) & \textbf{76.34 (0.4)} & \textbf{622.18 (0.3)} & 42.00 (0.2) \\
    & \CC{ROME$_\text{DNE}$} & \CC\textbf{89.81}  & \CC99.93 (0.0) & \CC\textbf{98.30 (0.1)} & \CC\textbf{77.93 (0.5)} & \CC75.63 (0.4) & \CC620.16 (0.4) & \CC\textbf{42.40 (0.2)} \\
    \cmidrule[0.005em]{2-9}
    & MEMIT & 83.40  & 94.22 (0.3) & 79.95 (0.5) & 41.02 (0.6) & 77.82 (0.4) & \textbf{627.33 (0.2)} & 39.44 (0.2) \\
    & PMET & 58.59  & 61.78 (0.7) & 44.88 (0.6) & 8.41 (0.3) & \textbf{78.50 (0.4)} & 627.18 (0.2) & 34.16 (0.2) \\
    & \CC{MEMIT$_\text{DNE}$} & \CC\textbf{88.10}  & \CC\textbf{98.16 (0.2)} & \CC\textbf{92.23 (0.3)} & \CC\textbf{58.78 (0.6)} & \CC76.80 (0.4) & \CC617.02 (0.4) & \CC\textbf{41.41 (0.2)} \\
    \cmidrule[0.005em]{1-9}
    \multirow{3}{*}{1e$^{\text{1}}$}
    & MEMIT & 83.54  & 94.33 (0.3) & 80.25 (0.5) & 41.53 (0.6) & 77.84 (0.4) & \textbf{627.34 (0.2)} & 39.47 (0.2) \\
    & PMET & 58.78  & 62.06 (0.7) & 45.07 (0.6) & 8.55 (0.3) & \textbf{78.52 (0.4)} & 627.21 (0.2) & 34.17 (0.2) \\
    & \CC{MEMIT$_\text{DNE}$} & \CC\textbf{87.79}  & \CC\textbf{97.91 (0.2)} & \CC\textbf{90.94 (0.3)} & \CC5\textbf{7.43 (0.6)} & \CC77.15 (0.4) & \CC622.58 (0.3) & \CC\textbf{41.62 (0.2)} \\
    \cmidrule[0.005em]{1-9}
    \multirow{3}{*}{1e$^{\text{2}}$}
    & MEMIT & 84.26  & 94.72 (0.3) & 81.77 (0.5) & 44.41 (0.6) & 78.02 (0.4) & 627.22 (0.2) & 39.73 (0.2) \\
    & PMET & 60.20  & 63.86 (0.7) & 46.56 (0.6) & 9.66 (0.3) & \textbf{78.75 (0.4)} & \textbf{627.41 (0.2)} & 34.37 (0.2) \\
    & \CC{MEMIT$_\text{DNE}$} & \CC\textbf{86.67}  & \CC\textbf{96.79 (0.2)} & \CC\textbf{88.90 (0.4)} & \CC\textbf{56.08 (0.6)} & \CC76.72 (0.4) & \CC625.26 (0.2) & \CC\textbf{41.29 (0.2)} \\
    \cmidrule[0.005em]{1-9}
    \multirow{3}{*}{1e$^{\text{3}}$}
    & MEMIT & 82.30  & 93.03 (0.4) & 80.27 (0.5) & 43.23 (0.6) & 75.49 (0.4) & 626.50 (0.2) & 39.25 (0.2) \\
    & PMET & 61.73  & 65.32 (0.7) & 48.60 (0.6) & 11.20 (0.4) & \textbf{78.65 (0.4)} & \textbf{627.82 (0.2)} & 34.65 (0.2) \\
    & \CC{MEMIT$_\text{DNE}$} & \CC\textbf{82.61}  & \CC\textbf{93.62 (0.3)} & \CC\textbf{82.90 (0.5)} & \CC\textbf{47.49 (0.6)} & \CC73.68 (0.4) & \CC626.34 (0.2) & \CC\textbf{39.76 (0.2)} \\
    \cmidrule[0.005em]{1-9}
    \multirow{3}{*}{1e$^{\text{4}}$}
    & MEMIT & 71.73  & 80.09 (0.5) & 66.19 (0.6) & 25.32 (0.5) & 70.28 (0.4) & 625.91 (0.2) & 36.43 (0.2) \\
    & PMET & 50.82  & 49.77 (0.7) & 38.58 (0.6) & 5.61 (0.3) & \textbf{76.82 (0.4)} & \textbf{627.87 (0.2)} & 33.38 (0.1) \\
    & \CC{MEMIT$_\text{DNE}$} & \CC\textbf{72.28}  & \CC\textbf{80.91 (0.5)} & \CC\textbf{67.66 (0.6)} & \CC\textbf{26.32 (0.5)} & \CC69.60 (0.4) & \CC626.14 (0.2) & \CC3\textbf{6.67 (0.2)} \\
    \bottomrule
    \end{tabular}
    \caption{Additional results of editing GPT2-xl with Counterfacts from 1e$^{\text{0}}$ to 1e$^{\text{4}}$. Within parentheses is the 95\% confidence interval.}
    \label{tab:gpt2_mcf}
\end{table*}
\begin{table}[ht]
\scriptsize
    \centering
    \begin{tabular}{c|c|c|c|c|c}
     \toprule
     \textbf{$\left|\sM\right|$} & \textbf{Editor} & \textbf{S. $\uparrow$} & \textbf{Effi. $\uparrow$} & \textbf{Para.$\uparrow$} & \textbf{Spec. $\uparrow$} \\
     \midrule
     \multirow{5}{*}{1e$^{\text{0}}$}
     & ROME & 52.44 & \textbf{99.88 (0.0)} & 95.27 (0.2) & \textbf{27.25 (0.4)} \\
     & \CC{ROME$_{\text{DNE}}$} & \CC{\textbf{52.58}} & \CC{99.87 (0.0)} & \CC{\textbf{96.66 (0.2)}} & \CC{\textbf{27.25 (0.4)}} \\
     \cmidrule[0.005em]{2-6}
     & MEMIT & 51.60 & \textbf{99.84 (0.0)} & 87.65 (0.4) & \textbf{27.24 (0.4)} \\
     & PMET & 51.57 & 98.43 (0.1) & 88.45 (0.4) & \textbf{27.24 (0.4)} \\
     & \CC{MEMIT$_{\text{DNE}}$} & \CC{\textbf{52.59}} & \CC{99.10 (0.1)} & \CC{\textbf{97.85 (0.2)}} & \CC{27.22 (0.4)} \\
     \cmidrule[0.005em]{1-6}
     \multirow{3}{*}{1e$^{\text{1}}$}
     & MEMIT & 52.20 & \textbf{99.70 (0.1)} & 92.99 (0.3) & \textbf{27.26 (0.4)} \\
     & PMET & 51.86 & 97.22 (0.2) & 92.17 (0.3) & 27.24 (0.4) \\
     & \CC{MEMIT$_{\text{DNE}}$} & \CC{\textbf{52.69}} & \CC{99.26 (0.1)} & \CC{\textbf{98.37 (0.1)}} & \CC{27.24 (0.4)} \\
     \cmidrule[0.005em]{1-6}
     \multirow{3}{*}{1e$^{\text{2}}$}
     & MEMIT & 52.63 & \textbf{99.56 (0.1)} & 93.45 (0.3) & \textbf{27.58 (0.4)} \\
     & PMET & 52.22 & 96.82 (0.2) & 92.21 (0.3) & 27.57 (0.4) \\
     & \CC{MEMIT$_{\text{DNE}}$} & \CC{\textbf{52.66}} & \CC{98.97 (0.1)} & \CC{\textbf{96.94 (0.2)}} & \CC{27.36 (0.4)} \\
     \cmidrule[0.005em]{1-6}
     \multirow{3}{*}{1e$^{\text{3}}$}
     & MEMIT & \textbf{53.37} & \textbf{98.81 (0.1)} & 93.38 (0.3) & \textbf{28.26 (0.4)} \\
     & PMET & 52.72 & 95.46 (0.2) & 90.73 (0.3) & 28.24 (0.4) \\
     & \CC{MEMIT$_{\text{DNE}}$} & \CC52.23 & \CC98.49 (0.1) & \CC \textbf{94.19 (0.3)} & \CC 27.27 (0.4) \\
     \cmidrule[0.005em]{1-6}
     \multirow{3}{*}{1e$^{\text{4}}$}
     & MEMIT & \textbf{51.01} & 96.35 (0.2) & 89.95 (0.4) & \textbf{26.80 (0.4)} \\
     & PMET & 49.42 & 90.47 (0.3) & 84.36 (0.4) & 26.46 (0.4) \\
     & \CC{MEMIT$_{\text{DNE}}$} & \CC 50.52 & \CC \textbf{96.45 (0.2)} & \CC \textbf{90.01 (0.4)} & \CC26.38 (0.4) \\
     \bottomrule
    \end{tabular}
    \caption{Additional results of editing GPT-J on zsRE from 1e$^{\text{0}}$ to 1e$^{\text{4}}$ edits. Within the parentheses is the 95\% confidence interval.}
    \label{tab:gptj_zsre}
\end{table}
Table \ref{tab:gptj_zsre} and \ref{tab:gpt2_mcf} report the results about editing GPT-J (6B) on zsRE and editing GPT2-xl on Counterfacts.
These are the additional results to our main experiments in Section \ref{sec:zsre_main} and \ref{sec:mcf_main}.
In all cases, editing with DNE returns higher generalization and in the most cases return the highest Score.
As we have pointed out in our main experiments' discussions, although DNE makes the 'Fluency' lower, this metrics only considers the diversity of the generation texts and can not well reflect the classical text fluency in our common sense.
And as the RS remains rather high, DNE will not cause the generation degeneration to stupidly repeat nonsense words.

\section{Detailed Experimental Results}
\label{sec:d_ab}
\begin{table}[ht]
\scriptsize
    \centering
    \begin{tabular}{c|c|c|c|c|c}
     \toprule
     \textbf{$\left|\sM\right|$} & \textbf{Editor} & \textbf{S. $\uparrow$} & \textbf{Effi. $\uparrow$} & \textbf{Para.$\uparrow$} & \textbf{Spec. $\uparrow$} \\
     \midrule
     \multirow{6}{*}{1e$^{\text{0}}$}
     & \CC{MEMIT$_{\text{DNE}}$} & \CC \textbf{44.48}  & \CC \textbf{80.31 (0.4)} & \CC \textbf{72.17 (0.5)} & \CC 24.31 (0.4) \\
     & MEMIT$_{\text{NT}}$ & 39.30  & 65.83 (0.5) & 49.91 (0.5) & \textbf{24.33 (0.4)} \\
     & MEMIT$_{\text{NE}}$ & 37.07  & 58.15 (0.5) & 44.19 (0.5) & \textbf{24.33 (0.4)} \\
     & MEMIT$_{\text{SNE}}$ & 41.88  & 74.43 (0.5) & 58.48 (0.5) & \textbf{24.33 (0.4)} \\
     & MEMIT$_{\text{UN}}$ & 39.34  & 62.80 (0.5) & 51.98 (0.5) & \textbf{24.33 (0.4)} \\
     & MEMIT$_{\text{RNP}}$ & 37.88  & 57.28 (0.5) & 48.44 (0.5) & \textbf{24.33 (0.4)} \\
     \cmidrule[0.005em]{1-6}
     \multirow{6}{*}{1e$^{\text{2}}$}
     & \CC{MEMIT$_{\text{DNE}}$} & \CC\textbf{47.31}  & \CC\textbf{89.31 (0.3)} & \CC\textbf{84.30 (0.4)} & \CC\textbf{24.78 (0.4)} \\
     & MEMIT$_{\text{NT}}$ & 45.34  & 82.88 (0.4) & 74.02 (0.5) & 24.64 (0.4) \\
     & MEMIT$_{\text{NE}}$ & 44.52  & 79.69 (0.5) & 70.65 (0.5) & 24.58 (0.4) \\
     & MEMIT$_{\text{SNE}}$ & 46.09  & 86.02 (0.4) & 77.63 (0.5) & 24.64 (0.4) \\
     & MEMIT$_{\text{UN}}$ & 43.52  & 74.60 (0.5) & 68.10 (0.5) & 24.48 (0.4) \\
     & MEMIT$_{\text{RNP}}$ & 45.88  & 84.39 (0.3) & 77.59 (0.5) & 24.60 (0.4) \\
     \cmidrule[0.005em]{1-6}
     \multirow{6}{*}{1e$^{\text{4}}$}
     & \CC{MEMIT$_{\text{DNE}}$} & \CC41.57  & \CC63.47 (0.5) & \CC58.59 (0.6) & \CC25.42 (0.4) \\
     & MEMIT$_{\text{NT}}$ & 41.58  & 62.62 (0.5) & 57.95 (0.6) & 25.69 (0.5) \\
     & MEMIT$_{\text{NE}}$ & 41.74  & 63.02 (0.5) & 57.94 (0.6) & \textbf{25.81 (0.4)} \\
     & MEMIT$_{\text{SNE}}$ & 41.89  & 63.26 (0.5) & \textbf{58.68 (0.6)} & 25.80 (0.4) \\
     & MEMIT$_{\text{UN}}$ & 29.55  & 35.92 (0.5) & 32.36 (0.5) & 23.38 (0.3) \\
     & MEMIT$_{\text{RNP}}$ & \textbf{41.93}  & 62.99 (0.5) & 58.51 (0.6) & 25.92 (0.4) \\
     \bottomrule
    \end{tabular}
    \caption{Detailed experimental results of editing GPT2-xl on zsRE. Within the parentheses is the 95\% confidence interval.}
    \label{tab:d_gpt2}
\end{table}
\begin{table*}[ht]
\small
    \begin{tabular}{c|c|c|c|c|c|c|c|c}
    \toprule
    \multirow{2}{*}{\textbf{$\left|\sM\right|$}} & \multirow{2}{*}{\textbf{Editor}} 
    & \textbf{Score} & \textbf{Efficacy} & \multicolumn{2}{c|}{\textbf{Generalization}} & \textbf{Specificity} & \textbf{Fluency} & \textbf{Consistency} \\
    \cmidrule[0.005em](lr){3-3}\cmidrule[0.005em](lr){4-4}\cmidrule[0.005em](lr){5-6}\cmidrule[0.005em](lr){7-7}\cmidrule[0.005em](lr){8-8}\cmidrule[0.005em](lr){9-9}
    & & \makecell[r]{S $\uparrow$} & \makecell[r]{ES $\uparrow$} & \makecell[r]{PS $\uparrow$} & \makecell[r]{PA $\uparrow$} & \makecell[r]{NS $\uparrow$} & \makecell[r]{GE $\uparrow$} & \makecell[r]{RS $\uparrow$} \\
    \midrule
    \multirow{6}{*}{1e$^{\text{0}}$}
     & \CC{MEMIT$_{\text{DNE}}$} & \CC92.47  & \CC99.75 (0.1) & \CC\textbf{99.08 (0.1)} & \CC\textbf{87.40 (0.4)} & \CC81.14 (0.4) & \CC614.80 (0.4) & \CC\textbf{42.40 (0.2)} \\
     & MEMIT$_{\text{NT}}$ & 91.70  & \textbf{99.86 (0.1)} & 95.19 (0.3) & 67.28 (0.6) & 82.00 (0.4) & 621.95 (0.2) & 41.63 (0.2) \\
     & MEMIT$_{\text{NE}}$ & 91.25  & 99.76 (0.1) & 93.14 (0.3) & 63.53 (0.5) & \textbf{82.53 (0.3)} & \textbf{622.09 (0.2)} & 41.69 (0.2) \\
     & MEMIT$_{\text{SNE}}$ & 92.38  & \textbf{99.86 (0.1)} & 97.83 (0.2) & 77.24 (0.5) & 81.71 (0.4) & 620.84 (0.2) & 42.73 (0.2) \\
     & MEMIT$_{\text{UN}}$ & \textbf{92.60}  & 99.77 (0.1) & 98.37 (0.2) & 81.85 (0.4) & 81.90 (0.4) & 621.11 (0.2) & 42.71 (0.2) \\
     & MEMIT$_{\text{RNP}}$ & 92.15  & 99.71 (0.1) & 96.24 (0.2) & 80.75 (0.4) & 82.41 (0.3) & 616.84 (0.4) & 41.37 (0.2) \\
    \cmidrule[0.005em]{2-9}
    \multirow{6}{*}{1e$^{\text{2}}$}
     & \CC{MEMIT$_{\text{DNE}}$} & \CC\textbf{92.64}  & \CC99.79 (0.1) & \CC\textbf{97.96 (0.2)} & \CC\textbf{81.21 (0.4)} & \CC82.27 (0.4) & \CC620.35 (0.2) & \CC\textbf{42.70 (0.2)} \\
     & MEMIT$_{\text{NT}}$ & 91.70  & 99.85 (0.1) & 94.89 (0.3) & 66.61 (0.5) & 82.23 (0.3) & 622.02 (0.2) & 41.65 (0.2) \\
     & MEMIT$_{\text{NE}}$ & 91.31  & 99.76 (0.1) & 92.90 (0.3) & 63.37 (0.5) & \textbf{82.88 (0.3)} & 621.75 (0.2) & 41.55 (0.2) \\
     & MEMIT$_{\text{SNE}}$ & 92.12  & \textbf{99.87 (0.1)} & 96.46 (0.2) & 71.87 (0.5) & 82.05 (0.3) & 621.61 (0.2) & 42.15 (0.2) \\
     & MEMIT$_{\text{UN}}$ & 91.06  & 99.12 (0.1) & 94.00 (0.3) & 70.44 (0.5) & 81.84 (0.3) & \textbf{622.67 (0.2)} & 41.31 (0.2) \\
     & MEMIT$_{\text{RNP}}$ & 91.97  & 99.81 (0.1) & 95.48 (0.2) & 72.95 (0.5) & 82.46 (0.3) & 621.45 (0.2) & 41.83 (0.2) \\
    \cmidrule[0.005em]{2-9}
    \multirow{6}{*}{1e$^{\text{4}}$}
     & \CC{MEMIT$_{\text{DNE}}$} & \CC\textbf{85.87}  & \CC\textbf{99.26 (0.1)} & \CC\textbf{89.82 (0.4)} & \CC\textbf{58.43 (0.6)} & \CC72.83 (0.4) & \CC618.10 (0.2) & \CC\textbf{40.33 (0.2)} \\
     & MEMIT$_{\text{NT}}$ & 85.83  & 99.09 (0.1) & 88.42 (0.4) & 55.68 (0.6) & \textbf{73.79 (0.4)} & 619.47 (0.2) & 40.13 (0.2) \\
     & MEMIT$_{\text{NE}}$ & 82.29  & 96.54 (0.3) & 80.70 (0.4) & 36.46 (0.5) & 72.97 (0.4) & 572.75 (0.3) & 36.97 (0.2) \\
     & MEMIT$_{\text{SNE}}$ & 85.84  & 99.14 (0.1) & 88.82 (0.4) & 56.49 (0.6) & 73.51 (0.4) & \textbf{619.59 (0.2)} & 40.32 (0.2) \\
     & MEMIT$_{\text{UN}}$ & 70.94  & 79.53 (0.6) & 68.52 (0.5) & 20.30 (0.5) & 66.13 (0.4) & 527.55 (0.5) & 24.85 (0.2) \\
     & MEMIT$_{\text{RNP}}$ & 85.85  & 99.12 (0.1) & 88.56 (0.4) & 56.32 (0.6) & 73.73 (0.4) & 618.90 (0.2) & 40.05 (0.2) \\
     \bottomrule
    \end{tabular}
    \caption{Detailed experimental results on GPT-J (6B). Within parentheses is the 95\% confidence interval.}
    \label{tab:d_gptj}
\end{table*}

Table \ref{tab:d_gpt2} and \ref{tab:d_gptj}, in integrate, report the detailed results of: Section \ref{sec:noisy} comparing with other methods of adding noises (NT and NE), and Section \ref{sec:m_ab}: the three ablation studies (SNE, UN, and RNP).
In the most cases, DNE achieves the highest scores and the best generalization.
While in some cases, DNE can be beaten by other methods, DNE still achieve the most robust performance gains on two models in all the cases.
For example, in Table \ref{tab:d_gptj}, MEMIT$_{\text{UN}}$ gets higher Score in editing 1e$^{\text{0}}$ case but its performance becomes dramatically degenerated when editing 1e$^{\text{4}}$ cases.
And also in Table \ref{tab:d_gpt2}, the results of NT and NE largely falls behind on editing 1e$^{\text{0}}$ case.
And NE shows significant lower generation qualities when applied on GPT-J in editing 1e$^{\text{4}}$ cases in Table \ref{tab:d_gptj}, because its generation fluency GE and consistency RS both get lower.

\end{document}